\documentclass[10pt,twocolumn,letterpaper,compsoc]{IEEEtran}
\usepackage{myothers}
\usepackage{times}
\usepackage{epsfig}
\usepackage{graphicx}
\usepackage{epstopdf}
\usepackage{amsmath}
\usepackage{amssymb}
\usepackage{amsfonts}
\usepackage{subfigure}
\usepackage{algorithm}
\usepackage{algorithmicx}

\usepackage{array}
\usepackage{multirow}
\usepackage{booktabs}

\usepackage{mathdots}
\usepackage{bm}
\usepackage{subfigure}

\usepackage{ulem}

\usepackage[pagebackref=true,breaklinks=true,letterpaper=true,colorlinks,bookmarks=false]{}

\usepackage{mdwmath}
\usepackage{mdwtab}

\newtheorem{corollary}[theorem]{Corollary}

\begin{document}
\normalem
%
\title{On the Relations of Correlation Filter Based Trackers and Struck
}
%
%
%
%

\author{Jinqiao WANG,~\IEEEmembership{Member,~IEEE,} and Ming~TANG,~\IEEEmembership{Member,~IEEE,} and Linyu ZHENG, and Jiayi Feng
\IEEEcompsocitemizethanks{\IEEEcompsocthanksitem J. Wang, M. Tang, L. Zheng and Jiayi Feng are with the National Lab of Pattern Recognition, Institute of Automation, Chinese Academy of Sciences, Beijing 100190, China (e-mail: jqwang@nlpr.ia.ac.cn).
This work was supported by Natural Science Foundation of China. Grant No. 61375035. \protect\\ Color versions of the figures in this paper are available online at http://ieeexplore.ieee.org.}}

%
%

\markboth{Journal of \LaTeX\ Class Files,~Vol.~6, No.~1, January~2016}%
{Shell \MakeLowercase{\textit{et al.}}: Bare Demo of IEEEtran.cls for Computer Society Journals}
%



\IEEEcompsoctitleabstractindextext{%
\begin{abstract}
In recent years, two types of trackers, namely correlation filter based tracker (CF tracker) and structured output tracker (Struck), have exhibited the state-of-the-art performance. However, there seems to be lack of analytic work on their relations in the computer vision community. In this paper, we investigate two state-of-the-art CF trackers, \ie, spatial regularization discriminative correlation filter (SRDCF) and correlation filter with limited boundaries (CFLB), and Struck, and reveal their relations. Specifically, after extending the CFLB to its multiple channel version we prove the relation between SRDCF and CFLB on the condition that the spatial regularization factor of SRDCF is replaced by the masking matrix of CFLB. We also prove the asymptotical approximate relation between SRDCF and Struck on the conditions that the spatial regularization factor of SRDCF is replaced by an indicator function of object bounding box, the weights of SRDCF in its loss item are replaced by those of Struck, the linear kernel is employed by Struck, and the search region tends to infinity. Extensive experiments on public benchmarks OTB50 and OTB100 are conducted to verify our theoretical results. Moreover, we explain how detailed differences among SRDCF, CFLB, and Struck would give rise to slightly different performances on visual sequences.

\end{abstract}

\begin{IEEEkeywords}
Visual tracking, correlation filters, structured output SVM tracker, Struck, ranking SVM tracker.
\end{IEEEkeywords}}

\maketitle

\IEEEdisplaynotcompsoctitleabstractindextext

%
\IEEEpeerreviewmaketitle

\section{Introduction}
\label{sec:introduction}
In recent years, two types of regression based trackers have exhibited the state-of-the-art performances. They are correlation filter based tracker (CF tracker)~\cite{bolme10,zhong2012robust,kalal2012tracking,dane14b,henriques15,dane15a,galoo15,tangm15} and structured output tracker (Struck)~\cite{hare2011,zhang2013preserving,yao2013part}. These trackers are of high accuracy and robustness for model-free tracking tasks in which no prior knowledge about the target object is known except for the initial frame.

The correlation filter has been used to solve various computer vision problems, such as object detection and recognition~\cite{kumar2005correlation, henriques13} and pose detection~\cite{henriques2014fast}. Since 2010 when the first CF tracker was proposed, the CF trackers have achieved top location performance at high speed. Bolme~\etal\cite{bolme10} proposed the first CF tracker, called minimum output sum of squared error (MOSSE). The expression of MOSSE in the spatial domain turned out to be the ridge regression~\cite{rifkin03} with a linear kernel~\cite{galoo13}. Therefore, Henriques~\etal~\cite{henriques2012,henriques15} utilized the circulant structure produced by the base sample to propose a kernelized correlation filter based tracker (KCF). The KCF used a single kernel and enabled really efficient learning with fast Fourier transform (FFT). Galoogani~\etal~\cite{galoo13} introduced multiple channels into the correlation filter. Danelljan~\etal~\cite{dane14a} extended the KCF~\cite{henriques2012} with low-dimensional adaptive color channels. Danelljan~\etal~\cite{dane14b} used correlation filters to fast estimate the proper object scale. Galoogani~\etal~\cite{galoo15} proposed the correlation filter with limited boundaries (CFLB) by introducing a mask on the samples into the loss item to restrain the boundary effect. By introducing a spatial regularization factor into the regularization item to restrain the boundary effect, Danelljan~\etal~\cite{dane15a} proposed the spatial regularized discriminative correlation filter tracker (SRDCF). Essentially, all CF trackers explore only one real sample, called base sample~\cite{henriques15}, and a set of virtual samples generated with the base sample to discriminatively fit their regressors to the Gaussian function with FFT, achieving high location accuracy at high speed with a low memory requirement. Currently, one of the most representative and publicly published CF trackers is SRDCF.

Before Struck~\cite{hare2011} was proposed, the tracking-by-detection algorithms labeled their training samples as positive or negative ones. Nevertheless, it is difficult for those trackers to decide the boundary between the two classes of samples in an image. The common ways usually rely on heuristic rules, for instance, using the distance between the centers of a sample and the located object to decide the samples' labels. Another drawback of the previous tracking-by-detection algorithms is that their label predictions are not directly coupled to the parameter estimation of the object location. To overcome these drawbacks, Hare~\etal~\cite{hare2011} proposed Struck which employed the support vector machines (SVMs) of structured outputs to explicitly match the objective of the tracker with the output space, and used continuous labels to avoid the heuristic rules in labeling samples. Essentially, Struck forces its model to regress to the exponential-like function in training, and then accurately predicts the object location according to the regression scores of candidates in locating. Up till 2013, Struck possessed the top location performance~\cite{wu13}.

Although CF trackers and Struck have achieved the state-of-the-art performances, there is still lack of deep insight into their relations in the computer vision community. In this paper, we try to reveal the theoretical correlation of these popular trackers, and experimentally verify it on public benchmarks. 

The \textbf{main contributions} of this paper are summarized as follows.
\begin{itemize}
  \item[1.] Extend CFLB to its novel version of multiple channels, CFLBMC.\footnote{After this work has been finished almost a year, we found a similar work~\cite{galoo17a}}
  \item[2.] Prove the relation between SRDCF and CFLBMC on the condition that the spatial regularization factor of SRDCF is replaced by the masking matrix of CFLBMC.
  \item[3.] Prove the asymptotical approximate relation between SRDCF and Struck on the conditions that the spatial regularization factor of SRDCF is replaced by an indicator function of object bounding box, the weights of SRDCF in its loss item are replaced by those of Struck, the linear kernel is employed by Struck, and the search region tends to infinity.
  \item[4.] Prove the asymptotical approximate equivalence between CFLBMC and Struck on the conditions that the weights of CFLBMC in its loss item are identical to those of Struck, the linear kernel is employed by Struck, and the search region tends to infinity.
  \item[5.] Experimentally explain how detailed differences of SRDCF, CFLBMC, and Struck with linear kernel would give rise to slightly different performances on some visual sequences even if they are really similar.
\end{itemize}

The remainder of this paper is organized as follows. In Sec.~\ref{sec:srdcf}, we first extend CFLB to its multi-channel version, CFLBMC, and then prove the relation of SRDCF and CFLBMC. 
Sec.~\ref{sec:asyEquiva} proves the asymptotical relation between SRDCF and Struck. 
The relations of CF trackers, Struck, and the ranking SVM based trackers~\cite{bai2011robust,bai2012robust} are analyzed in Sec.~\ref{sec:rankingCompare}. Extensive experiments on the public benchmarks OTB50~\cite{wu13} and OTB100~\cite{wu15} and their analysis are presented in Sec.~\ref{sec:experiments}. Sec.~\ref{sec:conclusion} summarizes our work.

\section{CFLB vs SRDCF}
\label{sec:srdcf}
In this section, we will first extend CFLB to its multiple channel version, CFLBMC, and then prove the relation of CFLBMC and SRDCF on the condition that the spatial regularization factor of SRDCF equals the masking matrix of CFLBMC. For simplicity of expressions, we describe our inference in one dimensional image. The inference in two dimensional image can be derived in a similar way.

The normal optimization function of correlation filter (CF) with linear kernel in spatial domain is~\cite{henriques2012,galoo13}
\begin{equation}
\label{eq:cfnormal}
E_n(\bm{\omega})=\frac{1}{2}\sum_{i=0}^{T-1}\left(y_i-\bm{\omega}^{\top}\mathbf{x}_i\right)^2 + \frac{\lambda}{2}\|\bm{\omega}\|^2_2,
\end{equation}
where $T$ is the number of circulant samples, $\mathbf{x}_i\in\mathbb{R}^T$ is a circulant sample, $\mathbf{x}_0$ is the base sample, $\mathbf{x}_i$, $i\neq 0$, are the virtual samples generated by periodic extension of $\mathbf{x}_0$, $\bm{\omega}\in\mathbb{R}^T$ is the filter, $\mathbf{y}\equiv(y_0,\ldots,y_{T-1})$ is the desired Gaussian label, and $\lambda$ is a tradeoff parameter.

\subsection{Multi-channel CFLB (CFLBMC)}
\label{sec:cflbmc}
A drawback of the CF comes from the periodic extension of the base sample. Such extension enables efficient training and detection with FFT, but it also leads to the unwanted boundary effect, which limits the spatial ranges of reliable negative samples in the training stage and reliable scores in the detection stage, resulting in degradation of location performance in cases of fast object motion, deformation, and occlusion. In CF trackers, a commonly used approach to resist the boundary effect is to extend the bounding box to include some local background of the target object. The proper extension of bounding box can also improve the robustness of CF trackers. Nevertheless, including local background in the bounding box may degrade the reliability of location in some visual sequences.

\begin{figure}[t]
  \centering
  \includegraphics[width=3.4in]{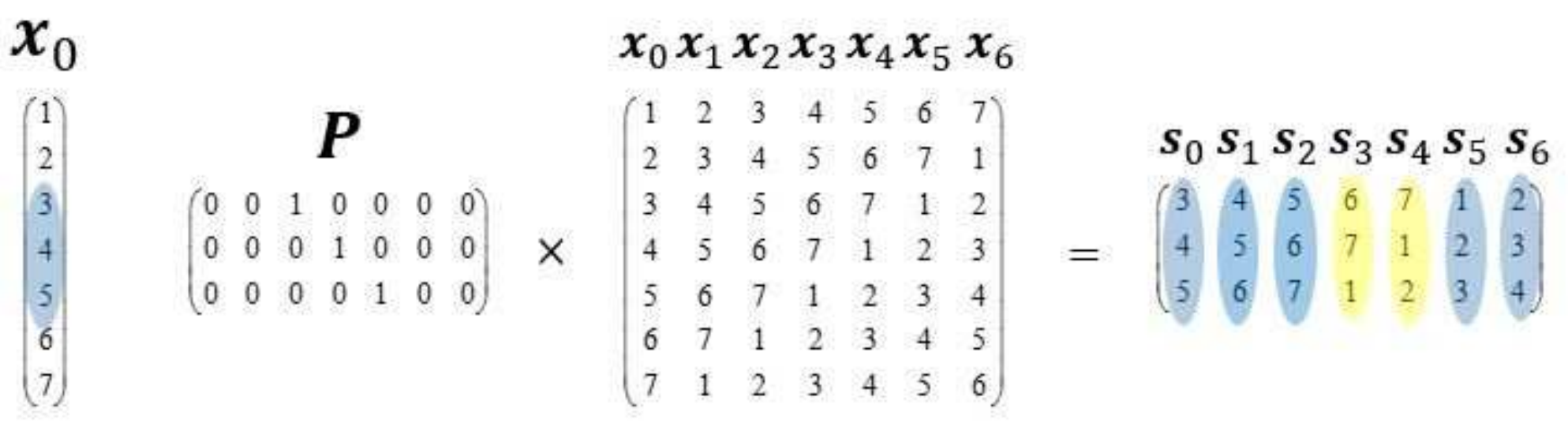}
  \caption{$\mathbf{P}$ decides which part of the circulant sample $\mathbf{x}_i$ is acting. $\mathbf{x}_0$ is the base sample, and $\mathbf{x}_i$, $i=1,\ldots,6$, are the virtual ones. The blue columns and yellow ones, which are acting regions of $\mathbf{x}_i$s and utilized to train the model of CFLB, indicate the real and virtual samples, respectively.}
  \label{fig:maskingmatrix}
\end{figure}

In order to alleviate the negative influence of boundary effect on location performance in a principled way, Galoogahi~\etal~\cite{galoo15} proposed the correlation filter with limited boundaries (CFLB) for single channel. 
The optimization objective of CFLB is as follows.
\begin{equation}
\label{eq:cflb-obj}
E_o(\bm{\omega}_o)=
\frac{1}{2}\sum_{i=0}^{T-1}\left(y_i-\bm{\omega}_o^{\top}\mathbf{P}\mathbf{x}_{i}\right)^2
+\frac{\lambda}{2}\left\|\bm{\omega}_o\right\|^2_2,
\end{equation}
where $\bm{\omega}_o\in\mathbb{R}^D$ is the filter, the masking matrix $\mathbf{P}=\left(\mathbf{0}_{D\times\frac{T-D}{2}},\mathbf{I}_D,\mathbf{0}_{D\times\frac{T-D}{2}}\right)$, where $\mathbf{I}_D$ is a $D\times D$ identity matrix, and $\mathbf{0}_{D\times\frac{T-D}{2}}$ is the $D\times\frac{T-D}{2}$ matrix of all 0s. $\mathbf{P}$ decides which part of the image patch is acting. Fig.~\ref{fig:maskingmatrix} illustrates how $\mathbf{P}$ works. Through introducing the masking matrix $\mathbf{P}$, the proportion of samples affected by the boundary effect are dramatically reduced. Specifically, the proportion will be decreased from the normal $\frac{1}{D}$ to $\frac{T-D+1}{T}$, and the boundary effects can be ignored if $T\gg D$.

In order to employ multiple channel features in CFLB, in this paper, we extend it to its multiple channel version, CFLBMC, as follows.
\begin{equation}
\label{eq:cflbmc-obj}
E_g(\bm{\omega}_g)=
\frac{1}{2}\sum_{i=0}^{T-1}\left(y_i-\sum_{l=0}^{L-1}\bm{\omega}_{g,l}^{\top}\mathbf{P}\mathbf{x}_{i,l}\right)^2
+\frac{\lambda}{2}\sum_{l=0}^{L-1}\left\|\bm{\omega}_{g,l}\right\|^2_2,
\end{equation}
where $L$ is the number of channels, $\bm{\omega}_g\equiv(\bm{\omega}_{g,0},...,\bm{\omega}_{g,L-1})$, $\bm{\omega}_{g,l}\in\mathbb{R}^D$ is the filter for channel $l$, and $\mathbf{x}_{i,l}\in\mathbb{R}^T$ is the channel $l$ of sample $\mathbf{x}_{i}$. In the following, we will describe how to minimize the objective function of CFLBMC by means of extending the solution procedure of CFLB to the multiple channel case.

Because the samples are circulant, Eq.~(\ref{eq:cflbmc-obj}) can be expressed in the Fourier domain by means of Parseval's equation~\cite{brigham74} as follows.
\begin{equation}
\label{eq:cflbmcfrequency}
\begin{aligned}
E_g(\bm{\omega}_g)=
&\frac{1}{2}\left\|\hat{\mathbf{y}}-\sum_{l=0}^{L-1}\mathrm{diag}(\hat{\mathbf{x}}_l)\sqrt{T}\mathbf{F}\mathbf{P}^{\top}\bm{\omega}_{g,l}\right\|^2_2\\
& +\frac{\lambda}{2}\sum_{l=0}^{L-1}\left\|\bm{\omega}_{g,l}\right\|^2_2,
\end{aligned}
\end{equation}
where $\hat{\mathbf{y}}$ is the discrete Fourier transform (DFT) of $\mathbf{y}$, $\hat{\mathbf{x}}_l$ is the DFT of $\mathbf{x}_{0,l}$, and $\mathbf{F}$ is the matrix of DFT. By using the fast Fourier transform, evaluation of Eq.~(\ref{eq:cflbmcfrequency}) can be accelerated greatly.

To simplify the expression, we cancel the subscript $g$ of $\bm{\omega}$ in the rest of this section.

\subsubsection{Augmented Lagrangian}
\label{sec:augmentedlagrangian}
Similar to Galoogahi~\etal~\cite{galoo15}, we introduce a set of auxiliary variables $\hat{\mathbf{g}}=(\hat{\mathbf{g}}_0,\ldots,\hat{\mathbf{g}}_{L-1})$ to minimize $E(\bm{\omega})$ by means of the augmented Lagrangian method (ALM)~\cite{boyd10}. Specifically,
\begin{equation}
\label{eq:cflbmc-aug}
\begin{array}{cl}
  \min\limits_{\bm{\omega}} & E(\bm{\omega},\hat{\mathbf{g}}),\\
  \text{s.t.} & \hat{\mathbf{g}}_l=\sqrt{T}\mathbf{F}\mathbf{P}^{\top}\bm{\omega}_l,\,l=0,\ldots,L-1,
\end{array}
\end{equation}
where
\begin{equation}
\label{eq:cflbmc-aug-obj}
\begin{aligned}
E(\bm{\omega},\hat{\mathbf{g}})=
& \frac{1}{2}\left\|\hat{\mathbf{y}}-\sum_{l=0}^{L-1}\mathrm{diag}(\hat{\mathbf{x}}_l)\hat{\mathbf{g}}_l\right\|^2_2\\
& +\frac{\lambda}{2}\sum_{l=0}^{L-1}\left\|\bm{\omega}_l\right\|^2_2.
\end{aligned}
\end{equation}
Then the augmented Lagrangian of Problem~(\ref{eq:cflbmc-aug}) is formulated as
\begin{equation}
\label{eq:cflbalm2}
\begin{aligned}
\mathcal{L}\left(\bm{\omega},\hat{\mathbf{g}},\hat{\bm{\zeta}}\right)=
&\frac{1}{2}\left\|\hat{\mathbf{y}}-\sum_{l=0}^{L-1}\mathrm{diag}(\hat{\mathbf{x}}_l)\hat{\mathbf{g}}_l\right\|^2_2
+\frac{\lambda}{2}\sum_{l=0}^{L-1}\left\|\bm{\omega}_l\right\|^2_2\\
&+\sum_{l=0}^{L-1}\hat{\bm{\zeta}}^{\top}_l\left(\hat{\mathbf{g}}_l-\sqrt{T}\mathbf{F}\mathbf{P}^{\top}\bm{\omega}_l\right)\\
&+\sum_{l=0}^{L-1}\frac{\mu_l}{2}\left\|\hat{\mathbf{g}}_l-\sqrt{T}\mathbf{F}\mathbf{P}^{\top}\bm{\omega}_l\right\|^2_2,
\end{aligned}
\end{equation}
where $\hat{\bm{\zeta}}=\left(\hat{\bm{\zeta}}_0,\ldots,\hat{\bm{\zeta}}_{L-1}\right)$, $\hat{\bm{\zeta}}_l$ is the Fourier transform of the
Lagrangian vector, and $\mu_l$ is the penalty factor that controls the rate of convergence of ALM. The alternate iteration process for optimization is as follows.

\vspace{2mm}
\noindent\textbf{Subproblem} $\hat{\mathbf{g}}$
\begin{equation}
\label{eq:cflbalmg}
\begin{aligned}
\hat{\mathbf{g}}_l&=\arg\min\mathcal{L}\left(\hat{\mathbf{g}}_l;\hat{\bm{\omega}}_l,\hat{\mathbf{g}}_{\setminus l},\hat{\bm{\zeta}}\right)\\
&=\left(\hat{\mathbf{y}}\circ\hat{\mathbf{x}}_l+\mu_l\hat{\bm{\omega}}_l-\hat{\bm{\zeta}}_l-\sum_{i=0,i\neq l}^{L-1}\hat{\mathbf{x}}_l\circ\hat{\mathbf{x}}_i^*\circ\hat{\mathbf{g}}_i\right)\\
&\circ^{-1}\left(\hat{\mathbf{x}}_l\circ\hat{\mathbf{x}}_l^*+\mu_l\mathbf{1}\right),
\end{aligned}
\end{equation}
where $\hat{\mathbf{g}}_{\setminus l}=\left(\hat{\mathbf{g}}_0,\ldots,\hat{\mathbf{g}}_{l-1},\hat{\mathbf{g}}_{l+1},\ldots,\hat{\mathbf{g}}_{L-1}\right)$, $\circ$ and $\circ^{-1}$ are point-wise multiplication and division, respectively, $\hat{\bm{\omega}}_l=\sqrt{T}\mathbf{F}\mathbf{P}^{\top}\bm{\omega}_l$, and $\mathbf{1}$ is a $T$ dimensional vector with 1 as its elements.

\vspace{2mm}
\noindent\textbf{Subproblem} $\bm{\omega}$
\begin{equation}
\label{eq:cflbalmh}
\begin{aligned}
\bm{\omega}_l&=\arg\min\mathcal{L}\left(\bm{\omega}_l;\bm{\omega}_{\setminus l},\mathbf{g}_l,\bm{\zeta}\right)\\
&=\left(\mu_l+\frac{\lambda}{\sqrt{T}}\right)^{-1}(\mu_l \mathbf{g}_l+\bm{\zeta}_l),
\end{aligned}
\end{equation}
where $\bm{\omega}_{\setminus l}=(\bm{\omega}_0,\ldots,\bm{\omega}_{l-1},\bm{\omega}_{l+1},\ldots,\bm{\omega}_{L-1})$,
$\mathbf{g}_l=\frac{1}{\sqrt{T}}\mathbf{P}\mathbf{F}^{\top}\hat{\mathbf{g}}_l$, and
$\bm{\zeta}_l=\frac{1}{\sqrt{T}}\mathbf{P}\mathbf{F}^{\top}\hat{\bm{\zeta}}_l$.

\vspace{2mm}
\noindent\textbf{Lagrange Multipliers Update}
\begin{equation}
\label{eq:cflbalmze}
\begin{aligned}
\hat{\bm{\zeta}}^{(i+1)}_l\leftarrow\hat{\bm{\zeta}}^{(i)}_l+\mu_l\left(\hat{\mathbf{g}}^{(i+1)}_l-\hat{\bm{\omega}}^{(i+1)}_l\right).
\end{aligned}
\end{equation}

\vspace{2mm}
\noindent\textbf{Choice of} $\bm{\mu}$
\begin{equation}
\label{eq:cflbalmmiu}
\begin{aligned}
\bm{\mu}^{(i+1)}=\min\left(\bm{\mu}_{\max},\beta\bm{\mu}^{(i)}\right),
\end{aligned}
\end{equation}
where $\bm{\mu}=(\mu_0,\ldots,\mu_{L-1})$, and $\bm{\mu}_{\max}=(\mu_{\max},\ldots,\mu_{\max})$. In practice, $\mu^{(0)}_l=0.01$, $l=0,\ldots,L-1$, $\beta=1.1$, $\mu_{\max}=20$, and $\lambda=10$. 6 alternate iterations may often produce accurate enough solutions.

Note that both the optimization problem and solution procedure of CFLBMC are not involved in any samples in historical frames for brief expression. And Problem~(\ref{eq:cflbmc-obj}) can be extended to involve historical samples simply based on what was done in SRDCF~\cite{dane15a}.

\subsection{Relation of CFLBMC and SRDCF}
\label{sec:cflbvssrdcf}
\begin{figure}[t]
  \centering
  \includegraphics[width=3.4in]{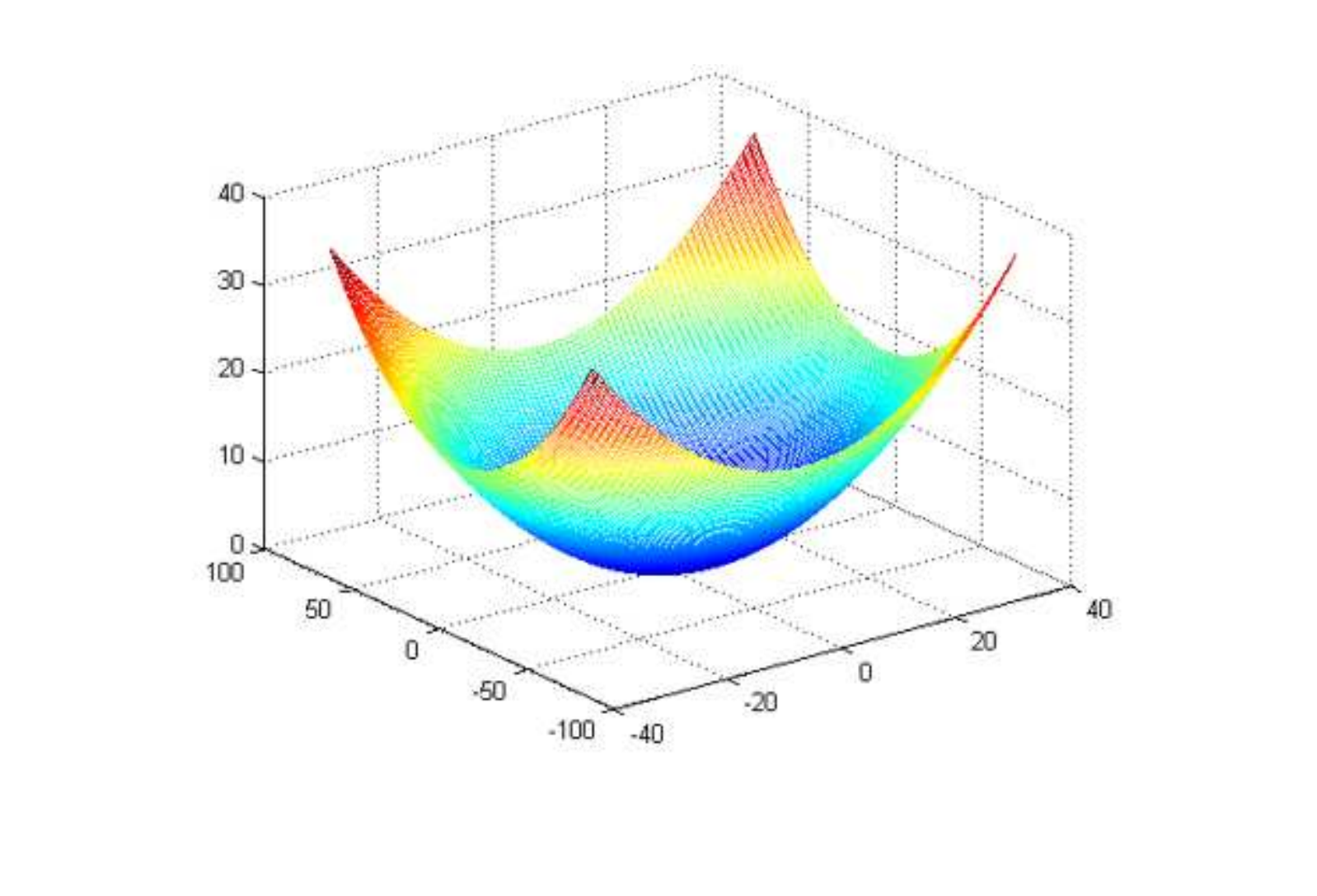}
  \caption{An example of the spatial regularization factor used in SRDCF~\cite{dane15a}.}
  \label{fig:srdcf-srf}
\end{figure}

While CFLB introduces a mask matrix on the spatially extended samples into its loss item, SRDCF~\cite{dane15a} introduces a spatial factor into its regularization item to restrain the boundary effect and extend the spatial ranges of reliable negative samples. Specifically, if only the current frame is involved,
%
the optimization function of SRDCF is
\begin{equation}
\label{eq:cfdane}
E_d(\bm{\omega}_d)=\frac{1}{2}\sum_{i=0}^{T-1}\left(y_i-\sum_{l=0}^{L-1}\bm{\omega}^{\top}_{d,l}\mathbf{x}_{i,l}\right)^2 + \frac{\lambda}{2}\sum_{l=0}^{L-1}\|\bm{\varsigma}\circ\bm{\omega}_{d,l}\|^2_2,
\end{equation}
where 
$\bm{\omega}_d\equiv(\bm{\omega}_{d,0},...,\bm{\omega}_{d,L-1})$, $\bm{\omega}_{d,l}\in\mathbb{R}^T$, 
and $\bm{\varsigma}\in\mathbb{R}^T$ is the regularization factor which is a upside down bell-shaped function, and decides how $\bm{\omega}_{d,l}$ acts in location. Fig.~\ref{fig:srdcf-srf} illustrates what it looks like.

\subsubsection{Proof of Relation}
\label{sec:proofofequivalence}
$\mathbf{P}\mathbf{v}=\left(0,\ldots,0,v_{\frac{T-D}{2}+1},\ldots,v_{\frac{T-D}{2}+D},0,\ldots,0\right)^{\top}$ for $\mathbf{v}\in\mathbb{R}^T$. Let 
$\bm{\omega}_{e,l}^{\top}=\bm{\omega}_{g,l}^{\top}\mathbf{P}$, then 
$(\mathbf{P}\bm{\omega}_{e,l})^{\top}(\mathbf{P}\bm{\omega}_{e,l})=\bm{\omega}_{g,l}^{\top}\bm{\omega}_{g,l}$, where $l=0,\ldots,L-1$.
We have
\begin{equation}
\label{galoodane1}
E_g(\bm{\omega}_g)=\frac{1}{2}\sum_{i=0}^{T-1}\left(y_i-\sum_{l=0}^{L-1}\bm{\omega}^{\top}_{e,l}\mathbf{x}_{i,l}\right)^2 + \frac{\lambda}{2}\sum_{l=0}^{L-1}\left\|\mathbf{P}\bm{\omega}_{e,l}\right\|^2_2,
\end{equation}
and
\begin{equation}\nonumber
\begin{array}{l}
\displaystyle\min_{\bm{\omega}_g}E_g(\bm{\omega}_g)=\\
\displaystyle\min_{\bm{\omega}_e}\frac{1}{2}\sum_{i=0}^{T-1}\left(y_i-\sum_{l=0}^{L-1}\bm{\omega}_{e,l}^{\top}\mathbf{x}_{i,l}\right)^2 + \frac{\lambda}{2}\sum_{l=0}^{L-1}\|\bm{\varsigma}_0\circ\bm{\omega}_{e,l}\|^2_2,
\end{array}
\end{equation}
where $\bm{\varsigma}_0=\left(\bm{\infty}_{1\times\frac{T-D}{2}},\mathbf{1}_{1\times D},\bm{\infty}_{1\times\frac{T-D}{2}}\right)_{1\times T}^{\top}$, $\bm{\infty}_{1\times\frac{T-D}{2}}$ and $\mathbf{1}_{1\times D}$ are two vectors with a large enough number and 1 as their elements. Therefore
\begin{equation}
\min_{\bm{\omega}_g}E_g(\bm{\omega}_g)=\min_{\bm{\omega}_d}E_d(\bm{\omega}_d),
\end{equation}
and their optimal solutions hold $\bm{\omega}_{g,l}^*=\mathbf{P}\bm{\omega}_{d,l}^*$, $l=0,\ldots,L-1$, on the condition of $\bm{\varsigma}_0=\bm{\varsigma}$.

\section{Asymptotical Approximate Relations of SRDCF, CFLBMC and Struck}
\label{sec:asyEquiva}
\begin{figure}[t]
  \centering
  \includegraphics[width=3.4in]{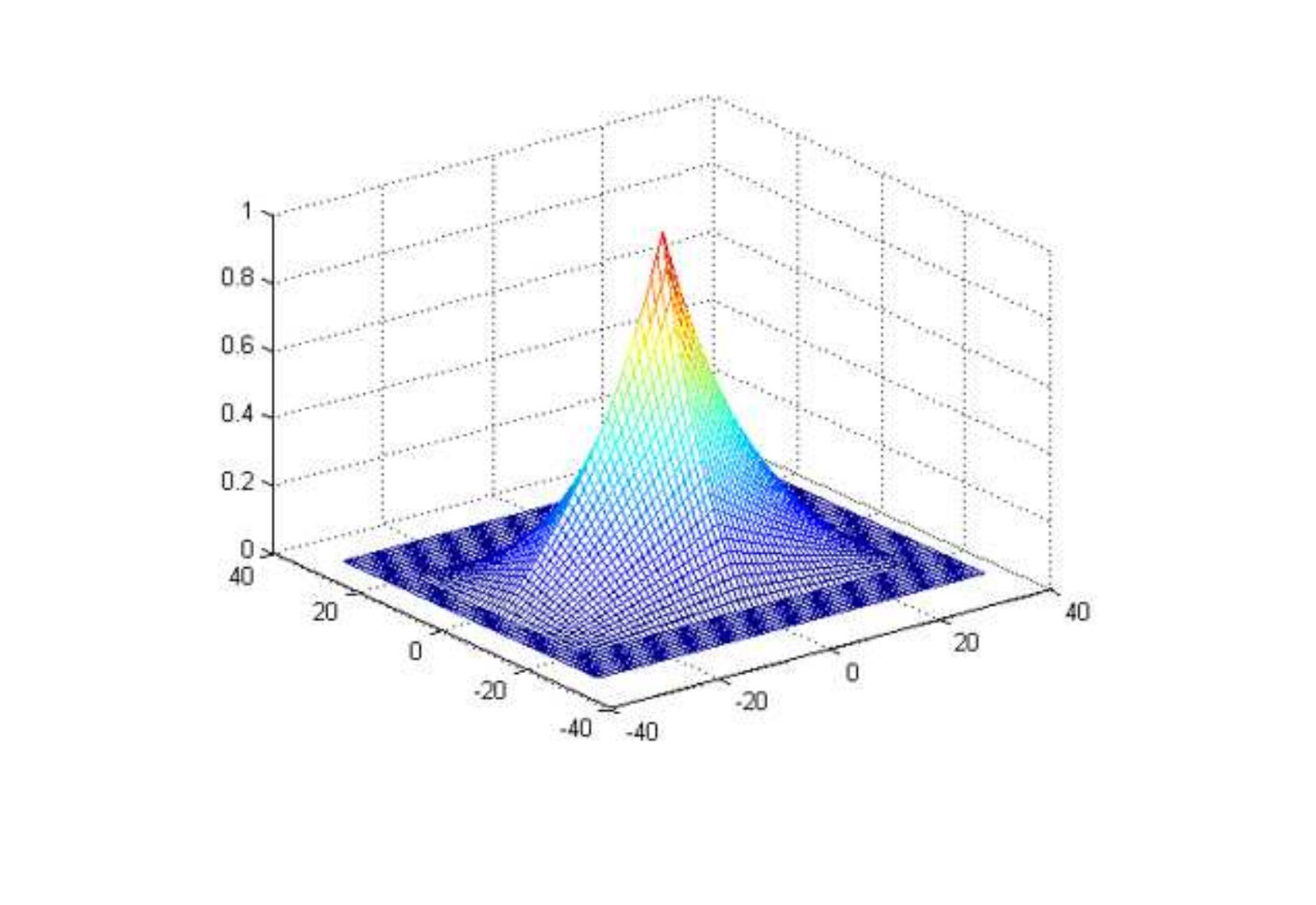}
  \caption{The overlap function used as labels in Struck~\cite{hare15}. See text for details.}
  \label{fig:iou}
\end{figure}

Struck~\cite{hare15} learns the appearance models of target object through solving the following optimization problem.
\begin{equation}
\label{eq:struckobj}
\begin{array}{rl}
\min\limits_{\bm{\omega},\bm{\xi}} & \displaystyle\frac{1}{2}\left\|\bm{\omega}\right\|^2_2+\lambda\sum_{i=1}^{N}\xi_i\\
\text{s.t.} & \forall i,\xi_i \geq 0,\\
            & \forall i,\forall\mathbf{y}\neq\mathbf{y}_i,\langle\bm{\omega},\delta\bm{\Phi}_i(\mathbf{y})
            \rangle\geq\Delta(\mathbf{y}_i,\mathbf{y})-\xi_i,
\end{array}
\end{equation}
where $\bm{\xi}=(\xi_1,\ldots\xi_N)$, $\delta\bm{\Phi}_i(\mathbf{y})=\bm{\Phi}(\mathbf{x}_i,\mathbf{y}_i)-\bm{\Phi}(\mathbf{x}_i,\mathbf{y})$, $\{(\mathbf{x}_i,\mathbf{y}_i)|i=1,\ldots,N\}$ is the set of training samples, $\bm{\Phi}(\mathbf{x},\mathbf{y})$ is a joint kernel map, $\mathbf{y}_i$ and $\mathbf{y}$ are bounding boxes of $w\times h$ with left-top corner coordinates $(l_i,t_i)$ and $(l,t)$, respectively. Note that $\mathbf{x}_i$ is generally larger than $w\times h$. $\Delta(\mathbf{y}_i,\mathbf{y})$ is defined such that $\Delta(\mathbf{y}_i,\mathbf{y})=0$ if $\mathbf{y}=\mathbf{y}_i$, the larger the distance between $\mathbf{y}_i$ and $\mathbf{y}$, the larger $\Delta(\mathbf{y}_i,\mathbf{y})$, and $\max_{\mathbf{y}}\Delta(\mathbf{y}_i,\mathbf{y})=1$. Specifically,
\begin{equation}
\label{eq:delta}
\begin{aligned}
\Delta(\mathbf{y}_i,\mathbf{y})=1-s_o(l,t;l_i,t_i,w,h),
\end{aligned}
\end{equation}
where $s_o(l,t;l_i,t_i,w,h)$ is the overlap function which expresses the ratio of intersection over union of bounding boxes $\mathbf{y}_i$ and $\mathbf{y}$, and
\begin{equation}
\label{eq:iou}
s_o(l,t;l_i,t_i,w,h)=\frac{p(l;l_i,w)p(t;t_i,h)}{2wh - p(l;l_i,w)p(t;t_i,h)},
\end{equation}
where $p(a;a_i,c)=2c-\max(a+c,a_i+c)+\min(a,a_i)$. Fig.~\ref{fig:iou} shows the shape of $s_o(l,t;l_i,t_i,w,h)$. It is seen that the larger the distance between $(l_i,t_i)$ and $(l,t)$, the smaller $s_o(l,t;l_i,t_i,w,h)$ is, and its maximum and minimum are 1 and 0, respectively. Note that the number of samples associated with $(\mathbf{x}_i,\mathbf{y}_i)$ is $(w_l-w+1)\times(h_l-h+1)$ if sampling is dense in the region of $w_l\times h_l$, where $w_l>w$ and $h_l>h$.

Problem~(\ref{eq:struckobj}) can be reformulated as
\begin{equation}
\label{eq:struckobj-2}
\begin{array}{rl}
\min\limits_{\bm{\omega},\bm{\xi}} & \displaystyle\frac{1}{2}\left\|\bm{\omega}\right\|^2_2+\frac{\lambda}{2}\sum_{i=1}^{N}\xi_i\\
\text{s.t.} & \forall i,\forall \mathbf{y}\neq\mathbf{y}_i,\\
&\xi_i \geq \max \{0,f(\mathbf{y};\mathbf{x}_i,\mathbf{y}_i,\bm{\omega})-s_o(l,t;l_i,t_i,w,h)\},
\end{array}
\end{equation}
where $f(\mathbf{y};\mathbf{x}_i,\mathbf{y}_i,\bm{\omega})=1-\langle \bm{\omega},\delta\bm{\Phi}_i(\mathbf{y})\rangle$. Suppose the number of $\mathbf{y}$ associated with $\mathbf{y}_i$ is $N_i$. Then it is drawn from the constraints of Problem~(\ref{eq:struckobj}) that
\begin{equation}
\label{eq:struckobj-2-constraints}
\forall i,\xi_i\geq\displaystyle\frac{1}{N_i}\sum_{\mathbf{y}}\max\{0,f(\mathbf{y};\mathbf{x}_i,\mathbf{y}_i,
\bm{\omega})-s_o(l,t;l_i,t_i,w,h)\}.
\end{equation}
Now consider the optimization problem
\begin{equation}
\label{eq:struck-lowerb}
\begin{array}{rl}
\min\limits_{\bm{\omega}}&\displaystyle\frac{1}{2}\left\|\bm{\omega}\right\|^2_2+\frac{\lambda}{2}\sum_{i=1}^{N}\frac{1}{N_i}
\sum_{\mathbf{y}}\max \{0,\\
&f(\mathbf{y};
\mathbf{x}_i,\mathbf{y}_i,\bm{\omega})-s_o(l,t;l_i,t_i,w,h)\}.
\end{array}
\end{equation}
It is clear that the loss function of Problem~(\ref{eq:struck-lowerb}) is the lower bound of that in Problem~(\ref{eq:struckobj-2}), and the optimal solution $\bm{\omega}^*$ of Problem~(\ref{eq:struck-lowerb}) is also that of Problem~(\ref{eq:struckobj}).

It is noticed that the essential purpose of $s_o(l,t;l_i,t_i,w,h)$ is to avoid the heuristic rules for labeling samples and improve the location precision through bell-shaped regression labels for $f(\mathbf{y};\mathbf{x}_i,\mathbf{y}_i,\bm{\omega})$. And there is no theoretical obstacle for replacing $s_o(l,t;l_i,t_i,w,h)$ with other bell-shaped labels. It will be shown in Sec.~\ref{sec:iouvsgaussianlabels} that there only exists really slight difference between the performances of Strucks with overlap function and Gaussian as regression goals, respectively, when we substitute a Gaussian for $s_o(l,t;l_i,t_i,w,h)$ in Struck codes~\footnote{The original codes of Struck are available at http://www.samhare.net/research}. Therefore from now on, $g(\mathbf{y})=s_o(l,t;l_i,t_i,w,h)$ express a Gaussian in this paper.

It is well-known that the hinge loss produces sparse solutions, which correspond to the sparse number of support vectors with respect to the total number of training samples, whereas the square loss generally generates dense solutions. In addition, the hinge loss is able to be more robust against some distractive training samples. Nevertheless, in the context of regression, such as in CF trackers and Struck, there will not exist distractive samples. Therefore, the only way that hinge and square losses may influence the localization differently is the sparsity of solutions. Sec.~\ref{sec:hingevsls} will show experimentally that the Strucks with hinge or square losses generate almost the same location. And Rifkin~\etal~\cite{rifkin03} has also shown that the square loss often provides a quite similar performance to the hinge loss does even in the context of classification where distractive training samples often exist. Consequently, we can expect that the replacement of hinge loss of Problem~(\ref{eq:struck-lowerb}) by the square one will generate almost the same performance. Then it follows after this substitution that
\begin{equation}
\label{eq:struck-rr}
\min\limits_{\bm{\omega}}\frac{1}{2}\left\|\bm{\omega}\right\|^2_2+\frac{\lambda}{2}\sum_{i=1}^{N}\frac{1}{N_i}
\sum_{\mathbf{y}}(f(\mathbf{y};\mathbf{x}_i,\mathbf{y}_i,\bm{\omega})-g(\mathbf{y}))^2.
\end{equation}

\begin{figure}[t]
  \centering
  \includegraphics[width=3in]{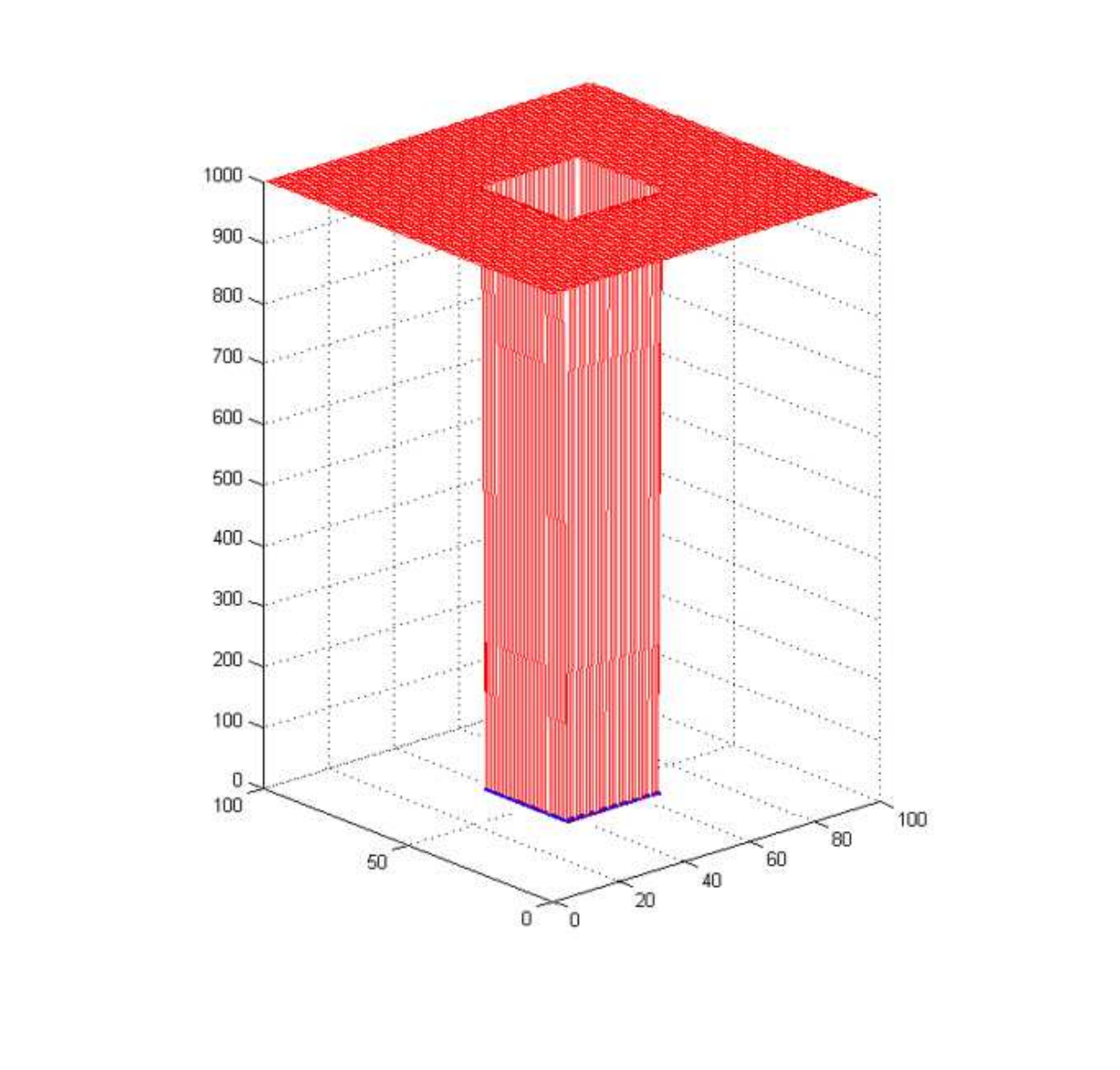}
  \caption{Another type of indicator function $\bm{\varsigma}_t$ used as the spatial regularization factor. See text for details.}
  \label{fig:rectanglewindow}
\end{figure}

\begin{figure}[t]
  \centering
  \includegraphics[width=3.4in]{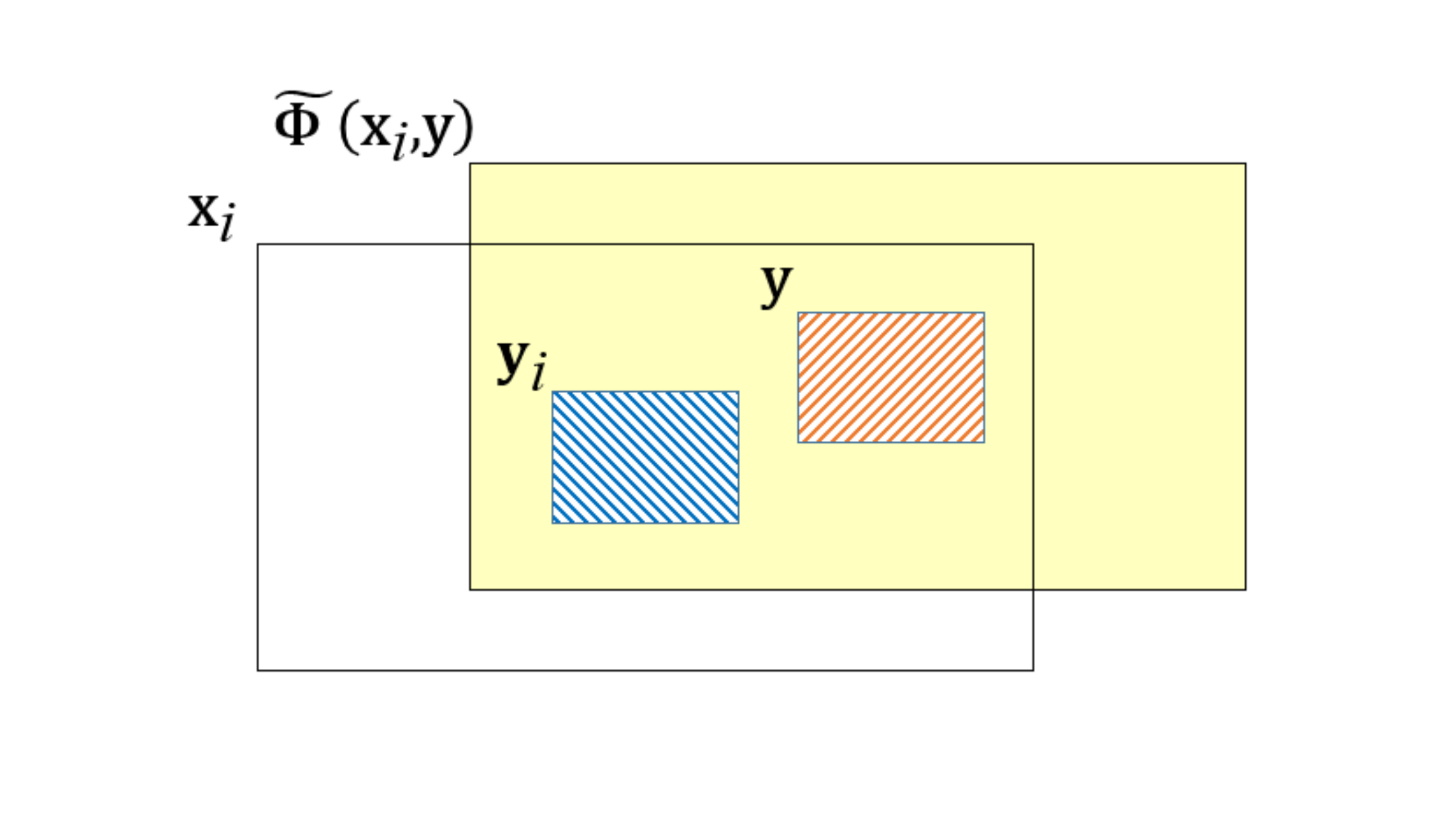}
  \caption{The extended sample $\tilde{\bm{\Phi}}(\mathbf{x}_i,\mathbf{y})$ when the linear kernel is employed in Struck~\cite{hare15}. See text for details.}
  \label{fig:enlargedsample}
\end{figure}

Let $\tilde{\bm{\omega}}$ be an extension of $\bm{\omega}$.
Suppose the sizes of both $\tilde{\bm{\omega}}$ and $\mathbf{x}_i$ are $w_l\times h_l$, $S_w=\{1,\ldots,w_l\}$, $S_h=\{1,\ldots,h_l\}$, $S_{w,c}=\{\frac{w_l-w}{2}+1,\ldots,\frac{w_l-w}{2}+w\}$, $S_{h,c}=\{\frac{h_l-h}{2}+1,\ldots,\frac{h_l-h}{2}+h\}$, and the spatial regularization factor
\begin{equation}
\label{eq:spatialreg1}
\bm{\varsigma}_t=\bm{\varsigma}_t(r,c)=\left\{\begin{array}{rl}
               1, & r\in S_{h,c},\,c\in S_{w,c},\\
               +\infty, & r\in S_h\setminus S_{h,c},\,c\in S_w\setminus S_{w,c},
             \end{array}\right.
\end{equation}
where $+\infty$ means a large enough number. $\bm{\varsigma}_t(r,c)$ takes 1 within the bounding box, and infinity otherwise, therefore can be accepted as another type of indicator function. Fig.~\ref{fig:rectanglewindow} shows the shape of $\bm{\varsigma}_t(r,c)$. Suppose further that the linear kernel is adopted in Problem~(\ref{eq:struckobj}), and $\tilde{\bm{\Phi}}(\mathbf{x}_i,\mathbf{y})$ is a $w_l\times h_l$ patch whose central part is from $\mathbf{x}_i$, and the location of the central part is determined by $\mathbf{y}$. Fig.~\ref{fig:enlargedsample} illustrates what $\tilde{\bm{\Phi}}(\mathbf{x}_i,\mathbf{y})$ looks like. Consider the following optimization problem.

\begin{equation}
\label{eq:struck-rr-enlarged}
\min\limits_{\tilde{\bm{\omega}}}\frac{1}{2}\left\|\bm{\varsigma}_t\circ\tilde{\bm{\omega}}\right\|^2_2+\frac{\lambda}{2}
\sum_{i=1}^{N}\frac{1}{N_i}
\left\|\tilde{\mathbf{f}}(\mathbf{y};\mathbf{x}_i,\mathbf{y}_i,\tilde{\bm{\omega}})-\tilde{\mathbf{g}}(\mathbf{y})\right\|^2_2,
\end{equation}
where $\circ$ means the point-wise multiplication, $\tilde{\mathbf{f}}(\mathbf{y};\mathbf{x}_i,\mathbf{y}_i,\tilde{\bm{\omega}})$ and $\tilde{\mathbf{g}}(\mathbf{y})$ are two vectors of $1\times N_i$ with elements $\tilde{f}(\mathbf{y};\mathbf{x}_i,\mathbf{y}_i,\tilde{\bm{\omega}})$ and $\tilde{g}(\mathbf{y})$, respectively, $\tilde{f}(\mathbf{y};\mathbf{x}_i,\mathbf{y}_i,\tilde{\bm{\omega}})=1-\langle \tilde{\bm{\omega}},\delta\tilde{\bm{\Phi}}_i(\mathbf{y})\rangle$, $\delta\tilde{\bm{\Phi}}_i(\mathbf{y})=\tilde{\bm{\Phi}}(\mathbf{x}_i,\mathbf{y}_i)-\tilde{\bm{\Phi}}(\mathbf{x}_i,\mathbf{y})$, and $\tilde{g}(\mathbf{y})=s_o(l,t;l_i,t_i,w,h)$ is a Gaussian. Note that $\tilde{g}(\mathbf{y})$ and $g(\mathbf{y})$ take the same parameters, with their domains being $w_l\times h_l$ and $w\times h$, respectively. Therefore, $\tilde{g}(\mathbf{y})\approx 0$ if its variables $\mathbf{y}$ do not belong to the $w\times h$ center part of its domain. It is obvious that the squared annular part of minimizer $\tilde{\bm{\omega}}^*$ corresponding to $+\infty$ part of $\bm{\varsigma}_t$ will consist of zero, and $\tilde{f}(\mathbf{y};\mathbf{x}_i,\mathbf{y}_i,\tilde{\bm{\omega}}^*)=f(\mathbf{y};\mathbf{x}_i,\mathbf{y}_i,\bm{\omega}^*)$. Therefore, the $w\times h$ central part of the minimizer $\tilde{\bm{\omega}}^*$ for Problem~(\ref{eq:struck-rr-enlarged}) equals the minimizer $\bm{\omega}^*$ for Problem~(\ref{eq:struckobj}).

Through explicitly expressing Problem~(\ref{eq:struck-rr-enlarged}) in the case of multiple channel features and multiplying the impact of each training sample by its loss, we have
\begin{equation}
\label{eq:struck-rr-enlarged-mc}
\min\limits_{\tilde{\bm{\omega}}}\frac{1}{2}\sum_{l=0}^{L-1}\left\|\bm{\varsigma}_t\circ\tilde{\bm{\omega}}_l\right\|^2_2+
\frac{\lambda}{2}\sum_{i=1}^{N}\alpha_i
\left\|\tilde{\mathbf{f}}(\mathbf{y};\mathbf{x}_i,\mathbf{y}_i,\tilde{\bm{\omega}})-\tilde{\mathbf{g}}(\mathbf{y})\right\|^2_2,
\end{equation}
where $\tilde{\bm{\omega}}=(\tilde{\bm{\omega}}_0,\ldots,\tilde{\bm{\omega}}_{L-1})$, and $\alpha_i=\frac{\beta_i}{N_i}$, $\beta_i$ is the impact of $\mathbf{x}_i$ on the optimal solution $\tilde{\bm{\omega}}^*$. If dense sampling is employed, as done in correlation filter based trackers~\cite{bolme10,henriques15,dane15a}, then $\forall i$, $N_i=N_c\equiv(w_l-w+1)\cdot(h_l-h+1)$.

It is noticed that Problem~(\ref{eq:struck-rr-enlarged-mc}) is exactly the same as the optimization object of SRDCF~\cite{dane15a} if the linear kernel is employed, except for the spatial regularization factor $\bm{\varsigma}$ and training samples. In SRDCF, $\bm{\varsigma}(r,c)=\mu+\eta[(c/w)^2+(r/h)^2]$, as shown in Fig.~\ref{fig:srdcf-srf}, and the training samples are generated cyclically from base samples. If $\bm{\varsigma}(r,c)=\bm{\varsigma}_t(r,c)$, then the ratio of true training samples over virtual ones is $\frac{w_l-w+1}{w_l}\times\frac{h_l-h+1}{h_l}$. Theoretically, the optimal solutions of Struck and SRDCF are of \emph{asymptotically and approximately relation} if both take Eq.~(\ref{eq:spatialreg1}) as their spatial regularization factors, the weights of Struck in its loss item are replaced by those of SRDCF, and Struck employs the linear kernel, because, given $w$ and $h$,
\begin{equation}
\label{eq:sample-ratio}
\begin{array}{cc}
  \displaystyle\lim_{w_l\rightarrow+\infty}\frac{w_l-w+1}{w_l}=1, & \displaystyle\lim_{h_l\rightarrow+\infty}\frac{h_l-h+1}{h_l}=1.
\end{array}
\end{equation}
Therefore, the negative influence of virtual training regions of SRDCF on location performance can be ignored if both $w_l$ and $h_l$ are large enough. According to~\cite{dane15a} and our experiments, while $w_l\approx4.5$ and $h_l\approx4.5$, the negative boundary effect could be ignored.\footnote{Note that an essential difference between the spatial regularization factor (SRC) of SRDCF and the other type of indicator function $\bm{\varsigma}_t$ shown in Fig.~\ref{fig:rectanglewindow} is that the SRC is smooth, resulting in few Fourier coefficients and accelerateing the optimization greatly. Of cause, such difference has the significance of the numerical calculation, but is not able to change our analysis on the relations.}

\begin{corollary}
CFLBMC is asymptotically and approximately equivalent to Struck on the conditions that the weights of CFLBMC in its loss item are identical to those of Struck, Struck employs the linear kernel, and the search region tends to infinity.
\end{corollary}
The corollary is immediately achieved by means of the equivalencies proved in Sec.~\ref{sec:cflbvssrdcf} and this section.


In the following section, we will analyze the relations of CF tracker, Struck, and another type of tracker, the ranking SVMs based tracker.

\section{Relations of Ranking SVM Tracker, CF trackers, and Struck}
\label{sec:rankingCompare}
In recent years, learning to rank has been applied to the visual tracking area. Bai and Tang~\cite{bai2011robust} proposed the tracker, RSVT, which simply employs the ranking SVM~\cite{joachims2002} to learn a ranking function. The function then ranks every candidate patch sampled from the new frame, and the patch with the most preferred ranking score is assumed to be the object. Bai and Tang~\cite{bai2012robust} derived the tracker, LRSVT, which explores both labeled and weakly labeled bipartite samples to train the ranking model, resulting in more robust performance. In this section, we will analyze their relations with CF trackers and Struck.

Primarily, the ranking SVMs based tracker (RSVT) learns its appearance models through solving the following optimization problem.
\begin{equation}
\label{eq:rsvm}
\begin{array}{rl}
\displaystyle \min_{\bm{\omega},\xi} & \displaystyle \frac{1}{2}\|\bm{\omega}\|^2+C\sum_{i,j}\xi_{ij}\\
\textrm{s.t.} & \forall i,j,\hskip 1mm \xi_{ij}\geq 0,\\
& \bm{\omega}^T(\Phi(\mathbf{x}_{i})-\Phi(\mathbf{x}_{j}))\geq 1-\xi_{ij},\\
& \mathbf{x}_{i}\in S_m^{(1)},\mathbf{x}_{j}\in S_m^{(0)},\hskip 1mm m=1,\ldots,M,
\end{array}
\end{equation}
where $M$ is the number of pairs of training sets, and the patches of $S_m^{(1)}$ rank higher than those in $S_m^{(0)}$. Note that the common point between RSVT, Struck, and CF trackers is that they all generate a ranking number for each image patch. The distinctions between them are that RSVT obtains the ranking number through a ranking function, whereas the others through regressed functions. It can be seen through comparing Problems~(\ref{eq:rsvm}),~(\ref{eq:struckobj}),~(\ref{eq:cfnormal}),~(\ref{eq:cflbmc-obj}), and~(\ref{eq:cfdane}) that Problem~(\ref{eq:rsvm}) is solved with patch pairs, whereas others are solved with patch-label pairs.
Therefore, the optimization objective of ranking SVM has to employ much more training samples (patch pairs) than Struck and CF trackers do if each tracker explores all possible samples in training. Specifically, if there are $N$ training patches, then $M=N-1$ and the number of pairs of training samples in ranking SVM is $(N-1)+(N-2)+\ldots+1=\frac{N(N-1)}{2}$, whereas the number of pairs of sample-labels is only $N$ in Struck and CF trackers.

In order to train RSVT more efficiently, less sample pairs often have to be explored. Compared to using all sample pairs, this, however, may introduce unwanted training errors due to inadequate sample pairs, lowering the discriminativity of its model.

Consequently, training the appearance models in Struck and CF trackers may be much more efficient than in RSVT so far as the number of training samples is concerned.

\section{Experimental Verifications}
\label{sec:experiments}
In our experiments, the Struck was implemented in C++, CFLBMC was implemented in MATLAB, and SRDCF used the original implementation in MATLAB. The experiments were performed on a PC with Intel Core i7 3.40GHz CPU and 20GB RAM. We verified the aforementioned relations on popular OTB50~\cite{wu13} and OTB100~\cite{wu15}. All the video results are accessible in the website.\footnote{http://www.nlpr.ia.ac.cn/iva/homepage/jqwang/Relation-video-results.zip}

The performance was quantitatively evaluated with popular criteria used in~\cite{wu13,henriques15,dane15a,tangm15}, \ie, center error criteria (center error, distance precision, precision plot) and overlap ones (overlap ratio, overlap precision, success plot, and area under curve (AUC)). The center error is calculated as the average Euclidean distance between the centers of located objects and their ground truths in a sequence. The distance precision is the percentage of frames where the objects are located within the center errors of 0 to $t_c$ pixels, with $t_c=20$, and the precision plot is simply a curve of the distance precisions with $t_c$ changing from 0 to 50 pixels. The overlap ratio is defined as the average ratio of intersection and union of the estimated bounding box and ground truth in a sequence, overlap precision as the percentage of frames with the overlap ratio exceeding $t_o$ in a sequence, with $t_o=0.5$, the success plot is simply a curve of overlap precisions with the overlap ratio changing from 0 to 1, and AUC is the area under the success plot.

\subsection{Implementation Details}
\label{sec:details}
For fair comparisons, all trackers employ the same feature, \ie, the HOG with $4 \times 4$ cells. According to the theoretical analysis in above sections, the Struck will employ the linear kernel in all experiments except for those in Sec.~\ref{sec:struckwithgaussian}. Gaussian labels with $\sigma=1/16$ are used by SRDCF, CFLBMC, and Struck. We will denote the Struck conducted in our experiments as the modified Struck if necessary in the following.

The search region and update rate of CFLBMC are set the same as SRDCF's. Specifically, in CFLBMC, we also set the squared search region to be 16 times the area of object bounding box, multiply the base sample by a Hann window, and set the update rate 0.025. 

Because of its spatial regularization factor, SRDCF generally learns its models with object and background. Note that the relation of SRDCF and CFLBMC has been proved on the condition that the spatial regularization factor of SRDCF equals the masking matrix of CFLBMC. Therefore, we set the samples of CFLBMC to coincide the target object itself as exactly as possible in order to experimentally show how local backgrounds affect the location performance of a tracker. Recall that the motivation of CFLB is to alleviate the boundary effect of circulant samples in a principled way, and that the equivalency of CFLBMC and Struck has been infered on the three conditions. Therefore, in our experiments, we set the samples of Struck to coincide the target object itself as exactly as possible to verify to what extent the goal of CFLB has been realized\footnote{In general, Struck can well exploit the local background of samples by means of setting the samples larger than the target object.}.

In our modified Struck, the width and height of the search region are 2.5 times of those of the object bounding box, respectively. The reason that 2.5 times is taken is that the fps will decrease about 2.5 times if 4 times of width and height are used. Because Struck can not be accelerated with FFT, dense sampling will result in too low fps in the training stage. Therefore, the sampling stride is set 4 pixels in training. In the location stage, on the other hand, dense sampling scheme is employed to ensure enough location accuracy. In addition, according to the CF trackers, the parts of sampled patches which are out of image borders will be filled by border pixels in the modified Struck. Other parameter values are set the same as the original Struck's.

All parameters of the trackers are fixed in all experiments.

\subsection{IoU Labels vs Gaussian Labels}
\label{sec:iouvsgaussianlabels}
In Sec.~\ref{sec:asyEquiva}, we proved the asymptotical and approximate relation of Struck and SRDCF on the four conditions. In our proof, the Gaussian labels are replaced by the IoU labels in Struck. In this section, we show that such substitution will only result in really slight effect on the location performance, therefore is reasonable.

\begin{figure}[t]
  \centering
  \includegraphics[width=1.7in]{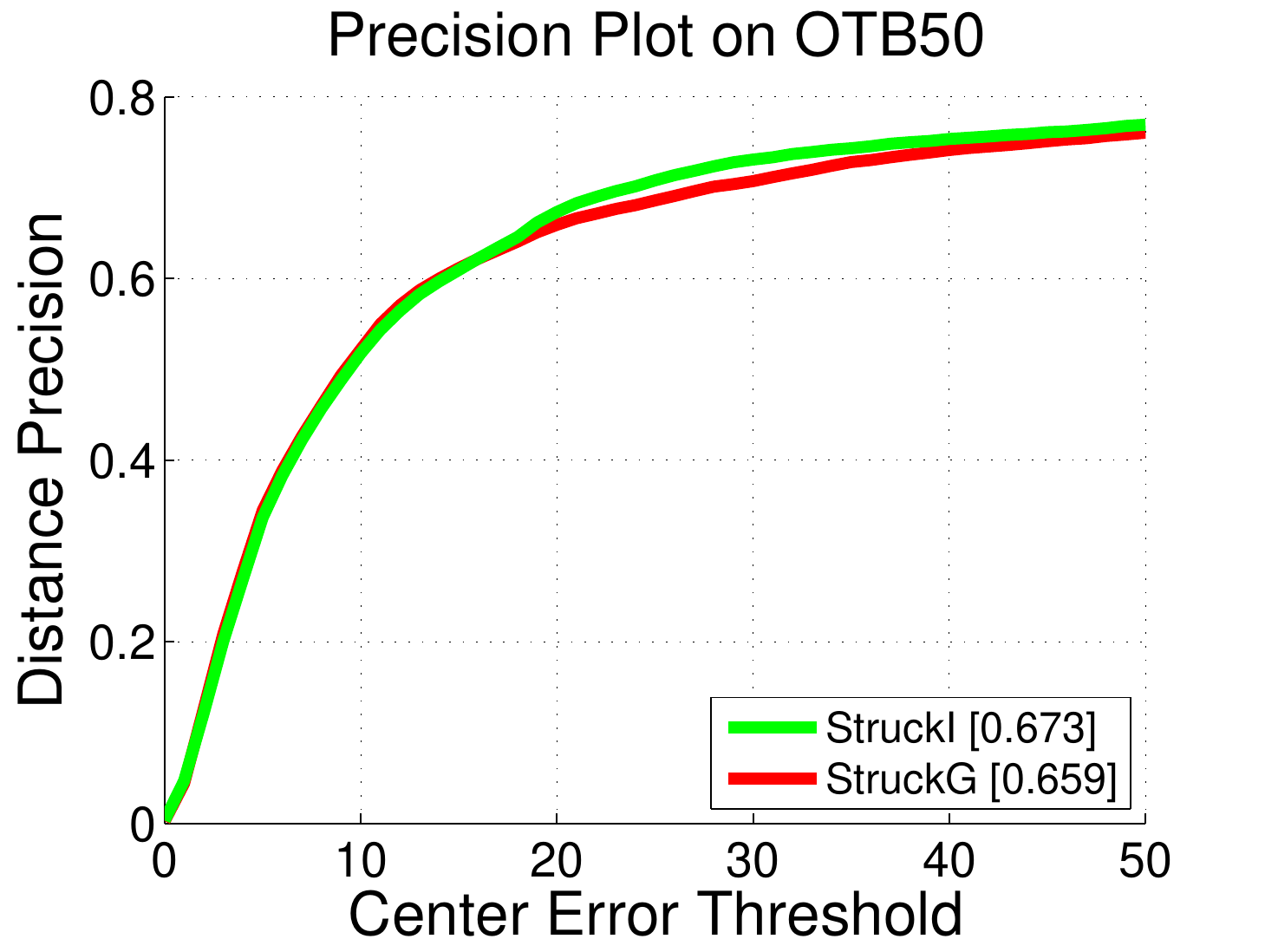}
  \includegraphics[width=1.7in]{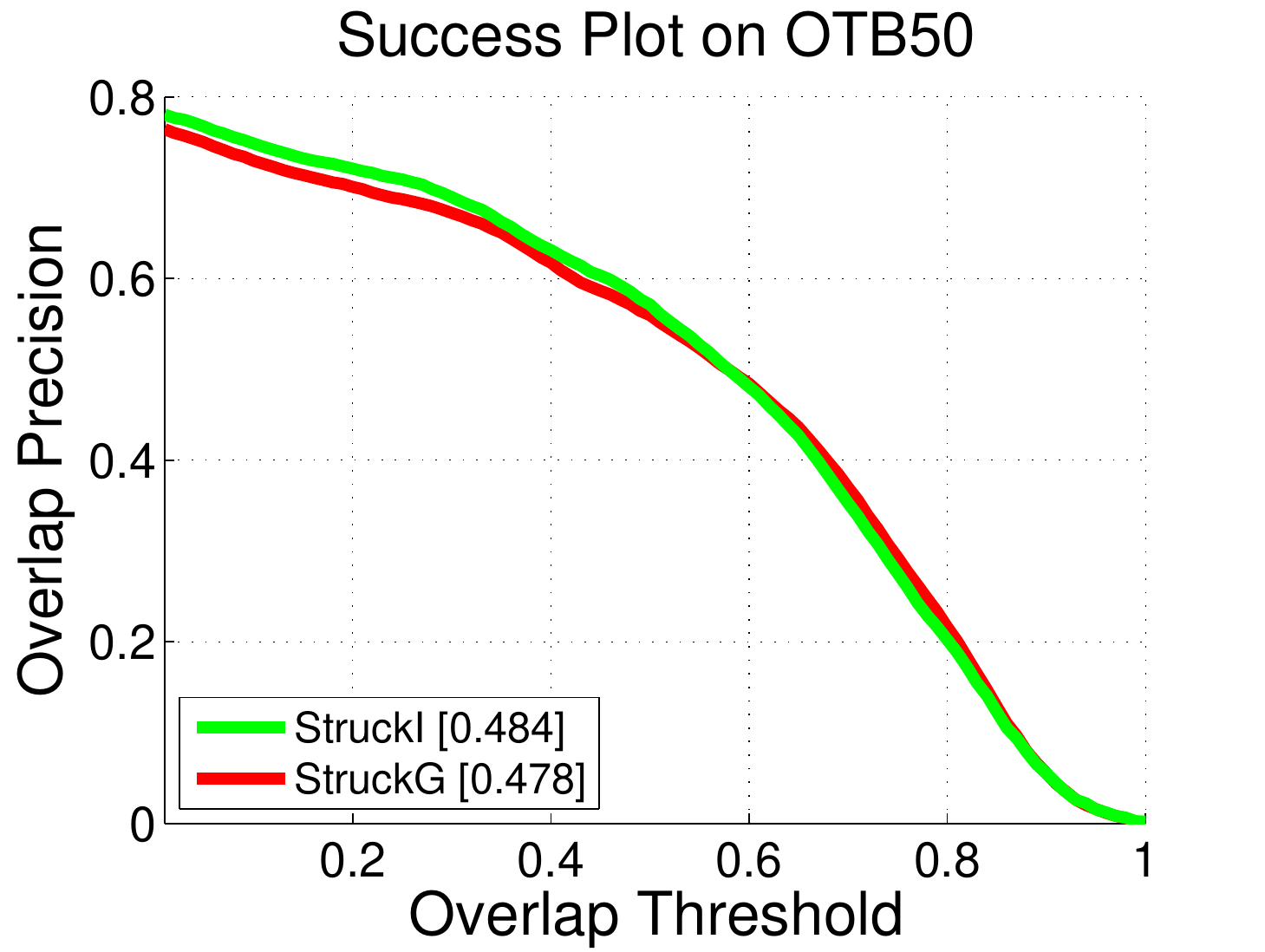}\\
\vskip 3mm
  \includegraphics[width=1.7in]{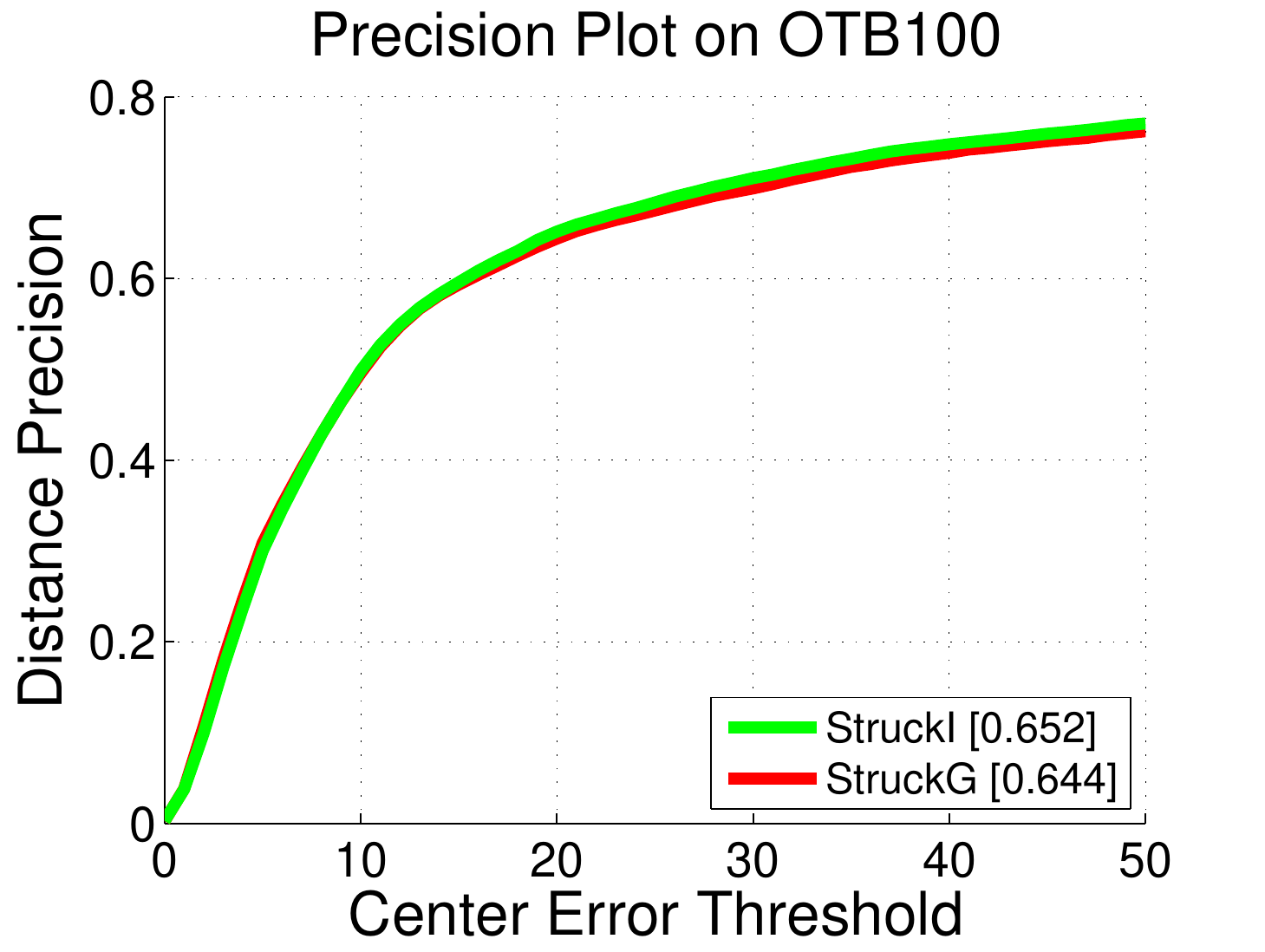}
  \includegraphics[width=1.7in]{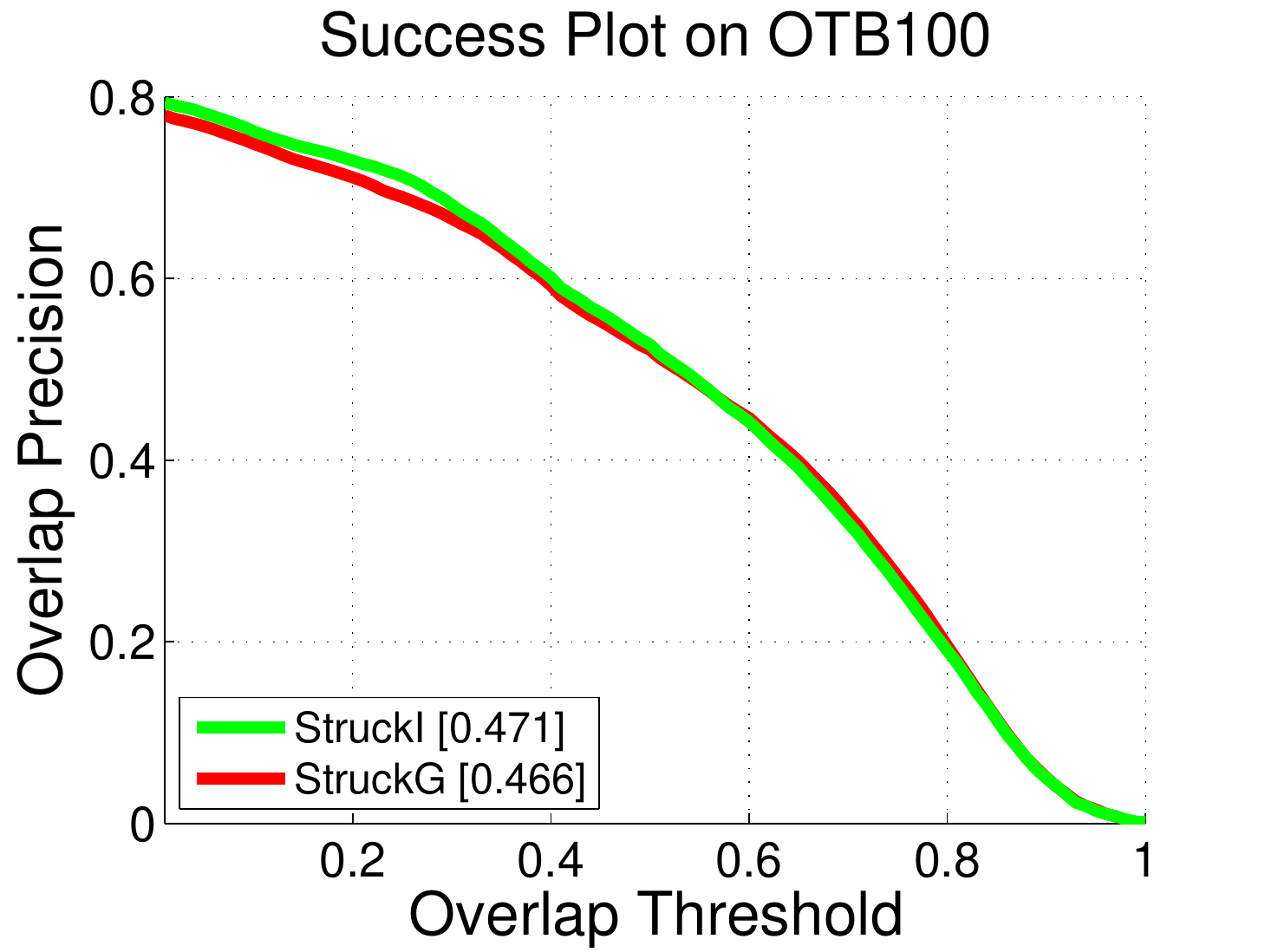}
  \caption{The average precision plots and success plots of Struck with IoU labels and Gaussian ones in OTB50 and OTB100. StruckI and StruckG denote the Struck with IoU labels and Gaussian ones, respectively. The mean distance precision scores and AUCs of the trackers are reported in the legends.}
  \label{fig:struck-gaussianlables}
\end{figure}

Fig.~\ref{fig:struck-gaussianlables} shows the experimental results on OTB50 and OTB100. The Strucks with IoU and Gaussian labels are denoted as StruckI and StruckG, respectively. Note that the only difference between StruckI and StruckG are their label functions. It is seen that the performance curves of StruckI and StruckG are almost overlapped.

\subsection{Hinge Loss vs Square Loss}
\label{sec:hingevsls}
In Sec.~\ref{sec:asyEquiva}, we stated that replacing the hinge loss by the square one will generate almost the same localization performance in the context of regression. In this section, we will validate such conclusion experimentally.

\begin{figure}[t]
  \centering
  \includegraphics[width=1.7in]{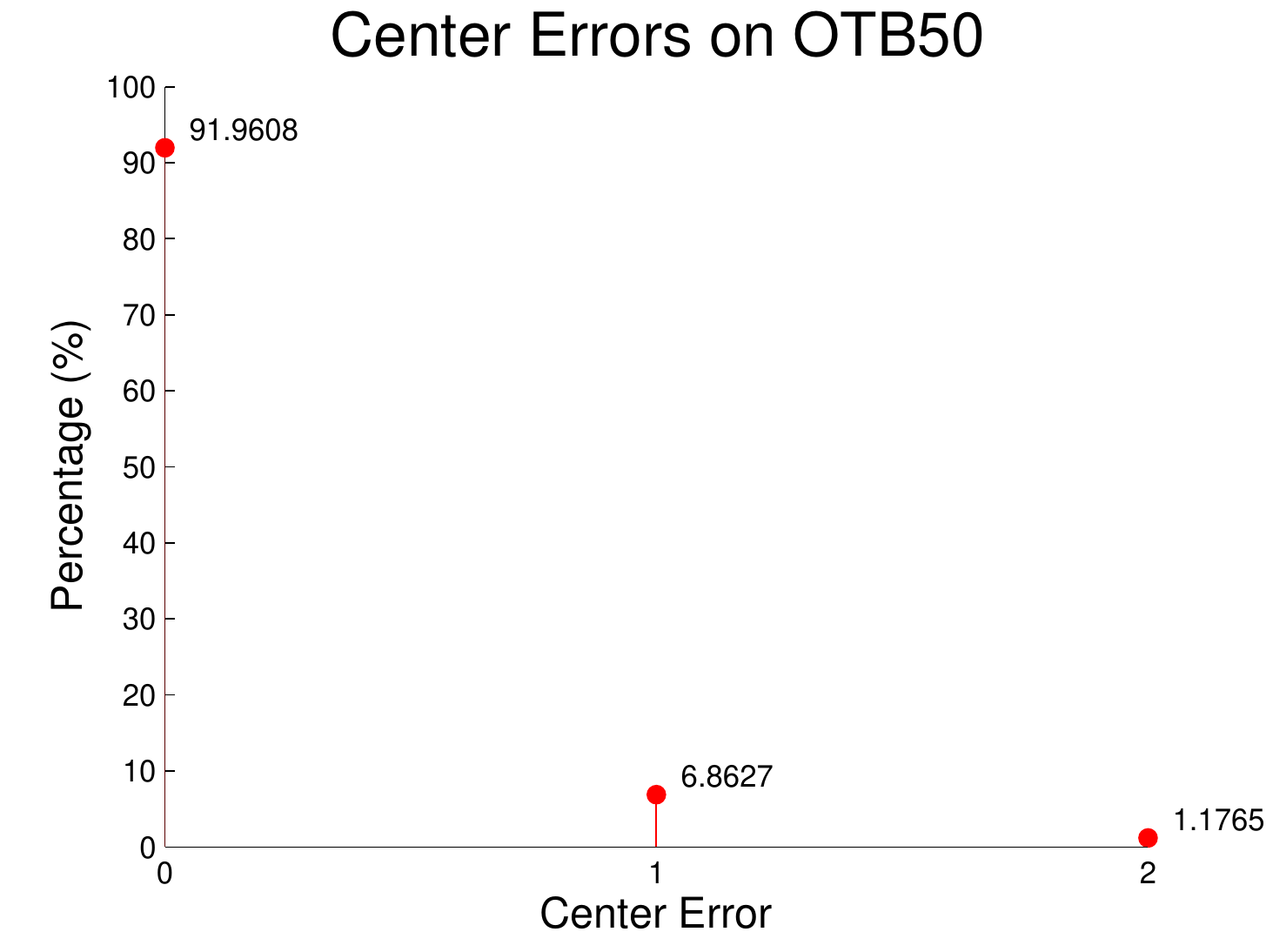}
  \includegraphics[width=1.7in]{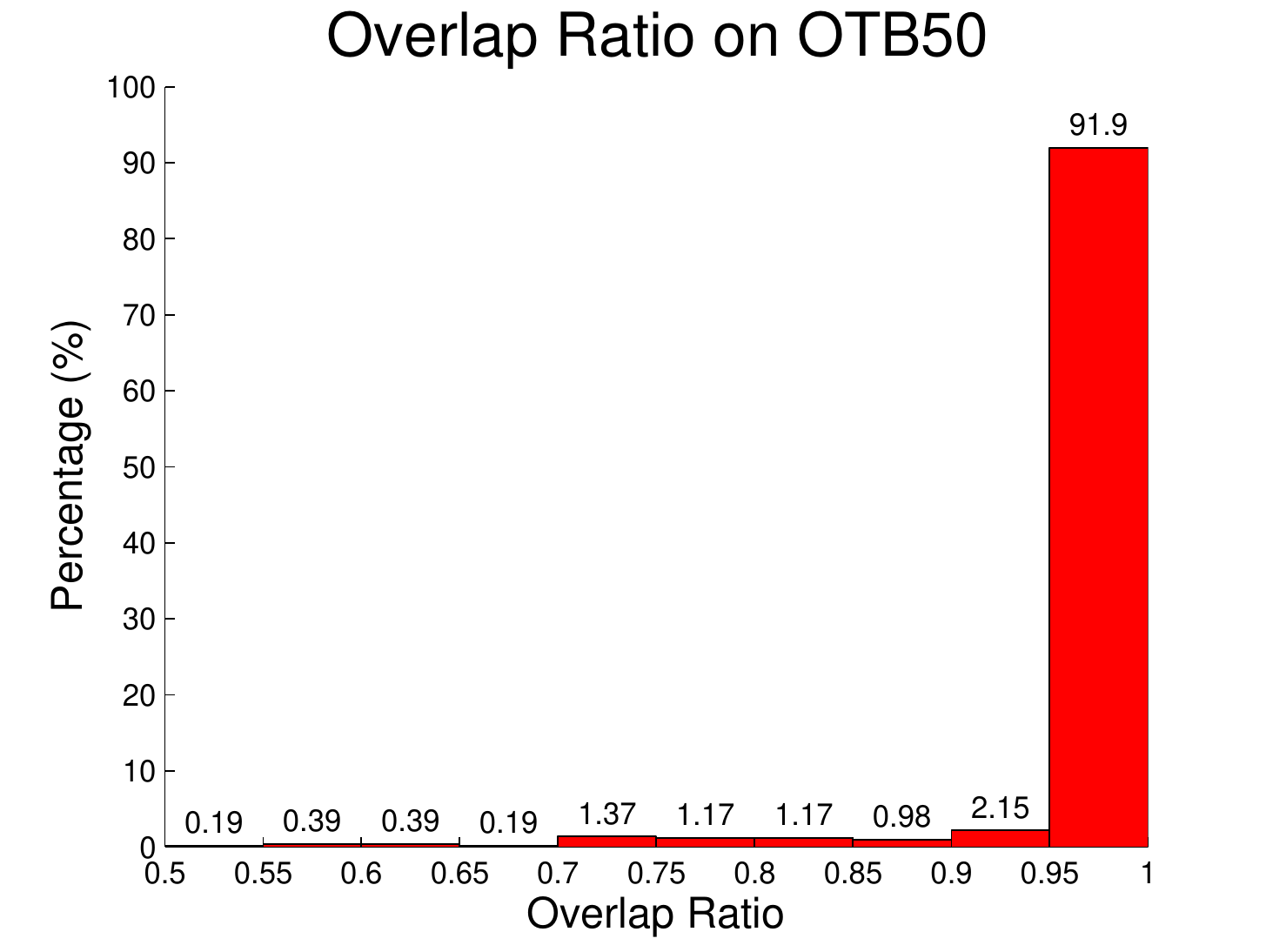}
  \caption{The histograms of center errors and overlap ratio (OR) of two Strucks with hinge and square losses, respectively, on OTB50. Note that the highest bar of OR histogram only consists of the ORs being 1.}
  \label{fig:2strucks-cd-iou}
\end{figure}

\begin{figure}[t]
  \centering
  \subfigure[$(1,0.7419)$]{\includegraphics[width=0.8in]{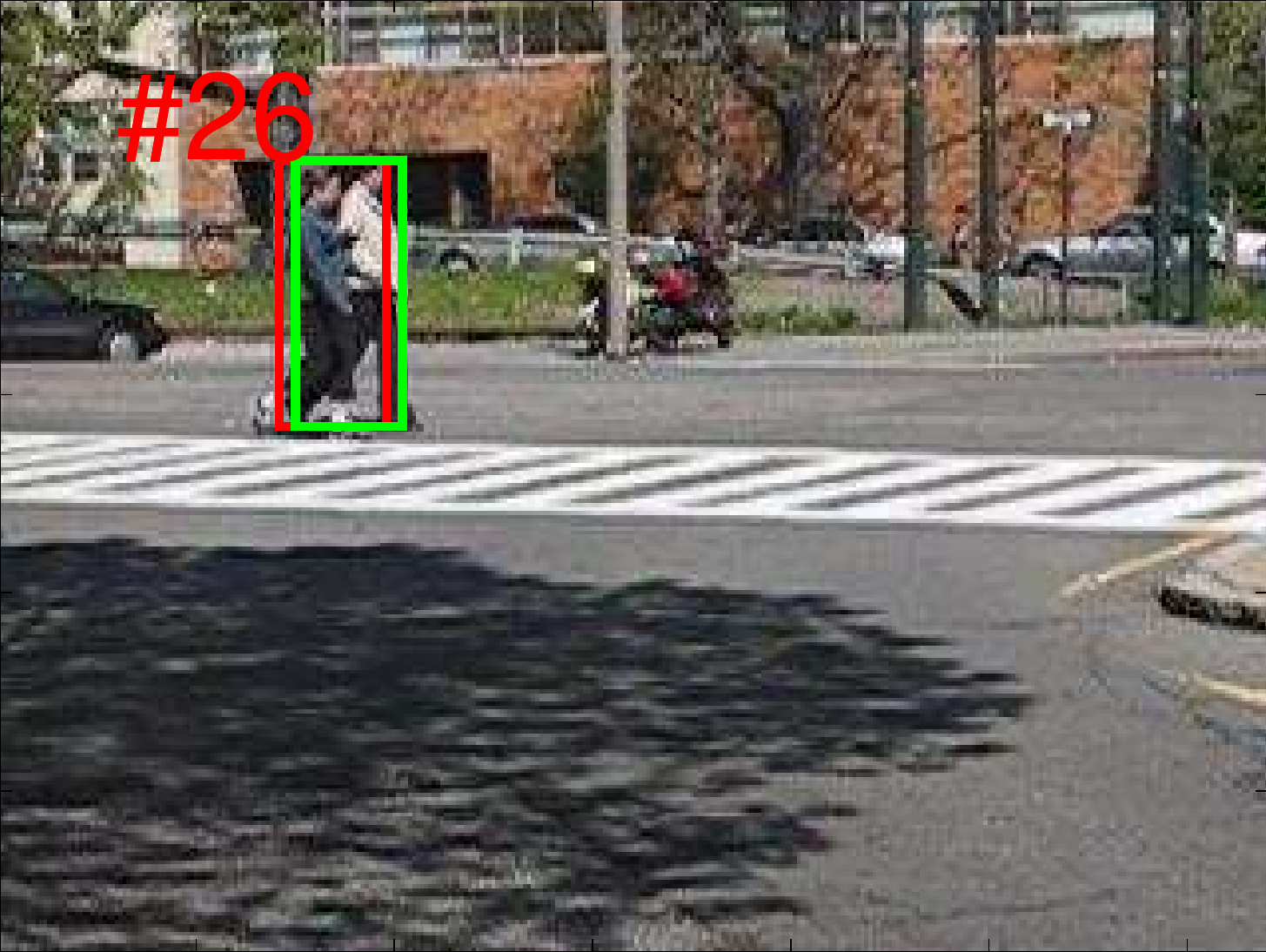}}
  \subfigure[$(1,0.7333)$]{\includegraphics[width=0.8in]{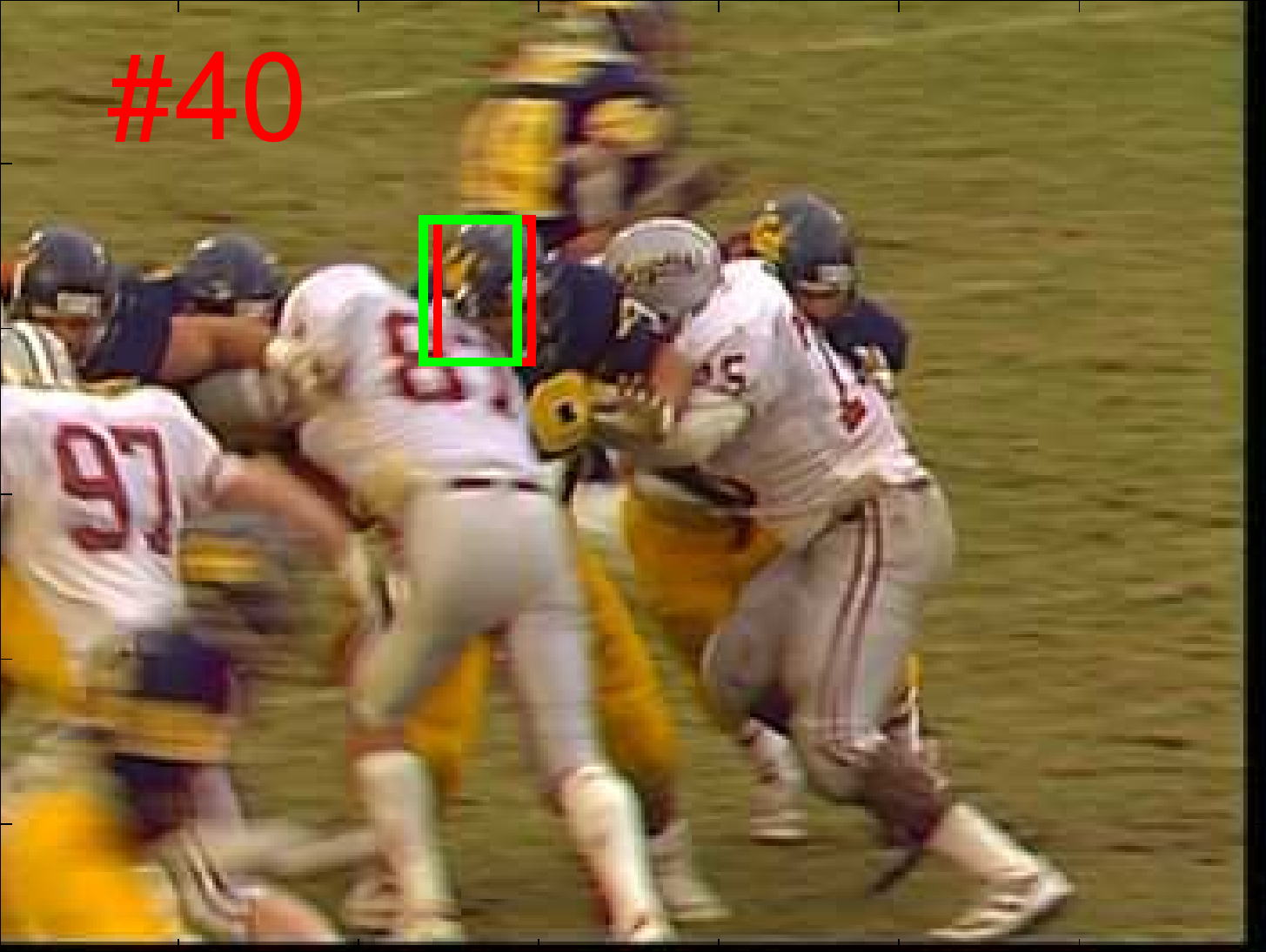}}
  \subfigure[$(1,0.7241)$]{\includegraphics[width=0.8in]{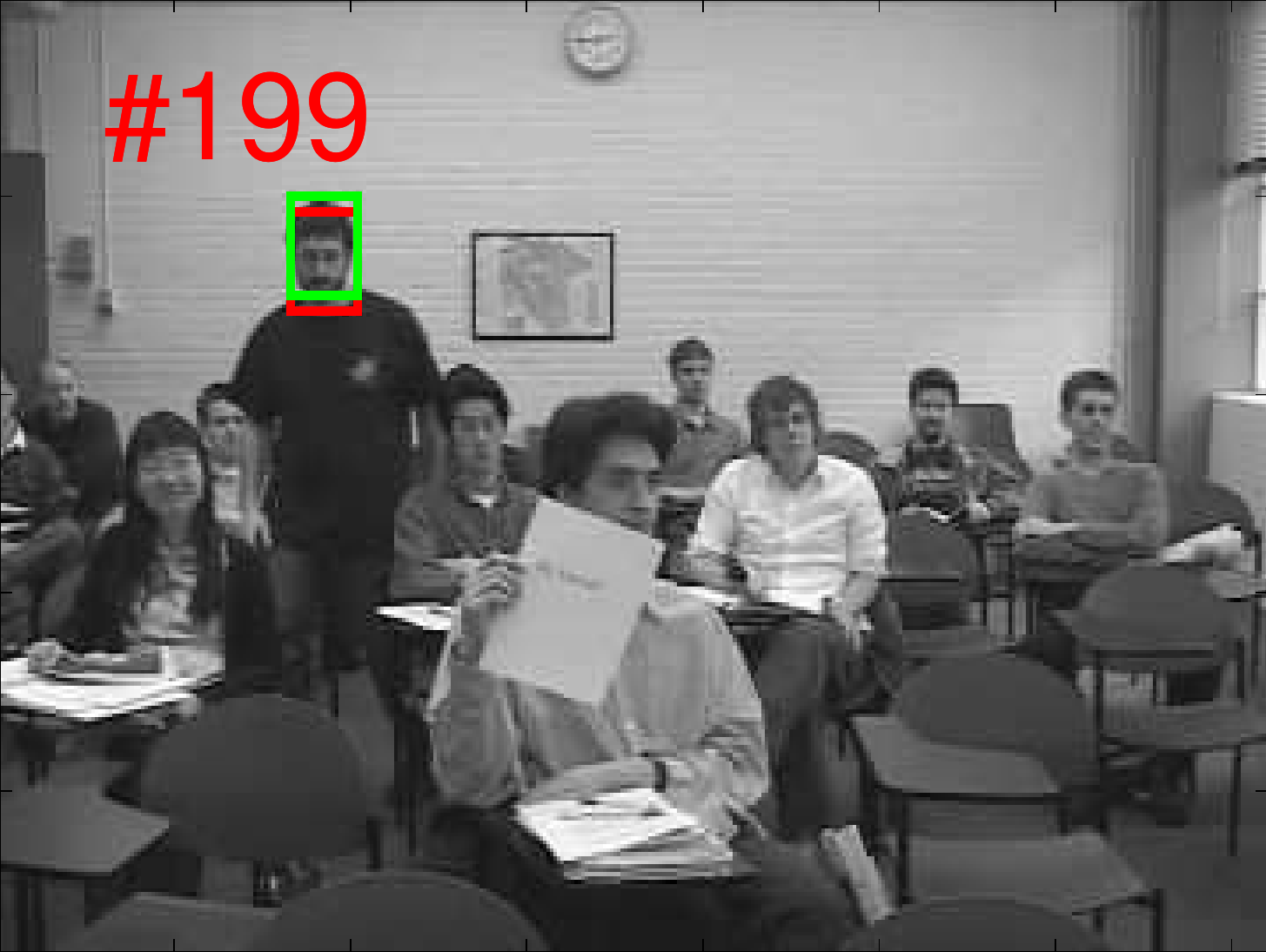}}
  \subfigure[$(1,0.7714)$]{\includegraphics[width=0.8in]{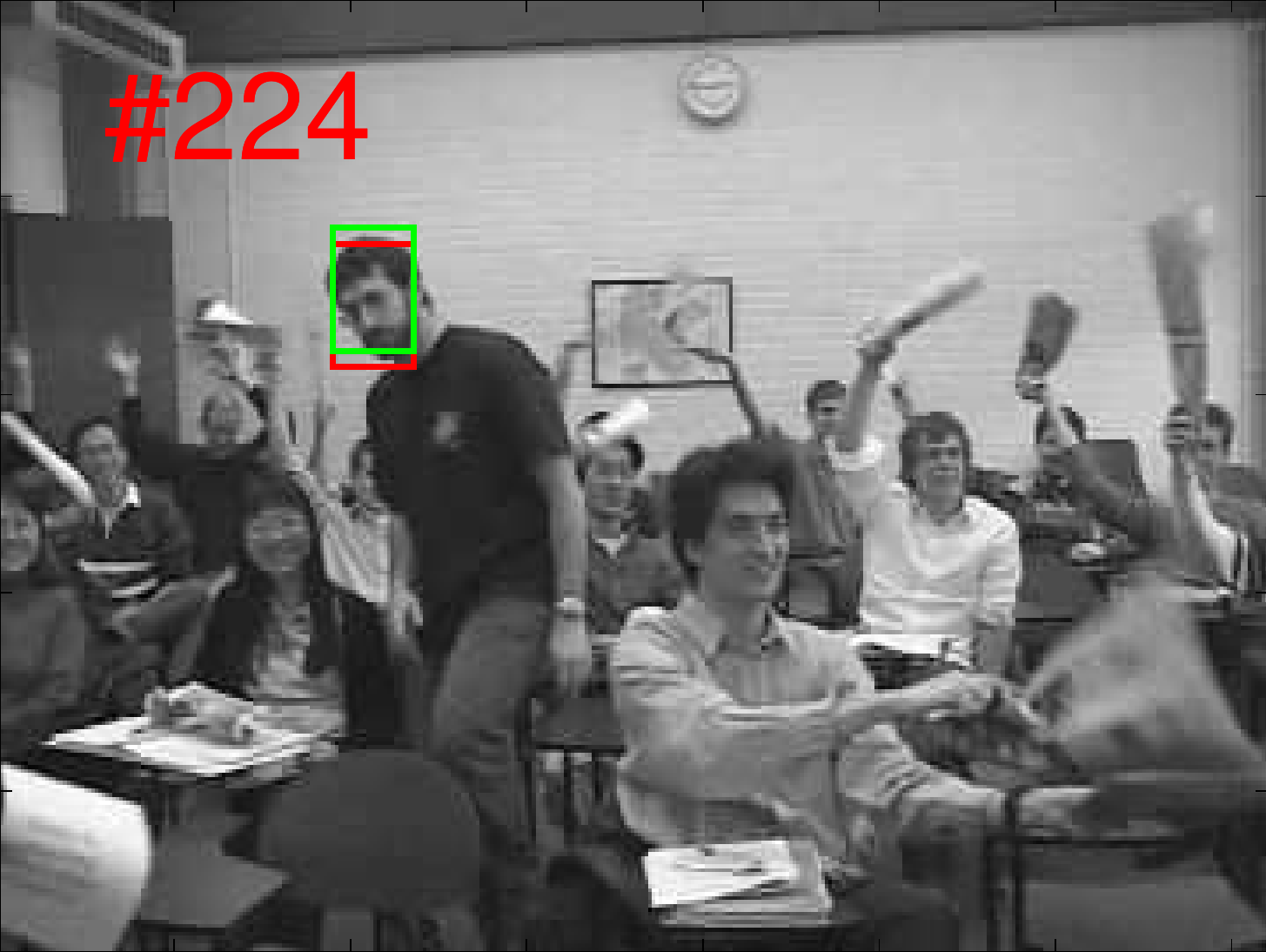}}\\
  \subfigure[$(1,0.6364)$]{\includegraphics[width=0.8in]{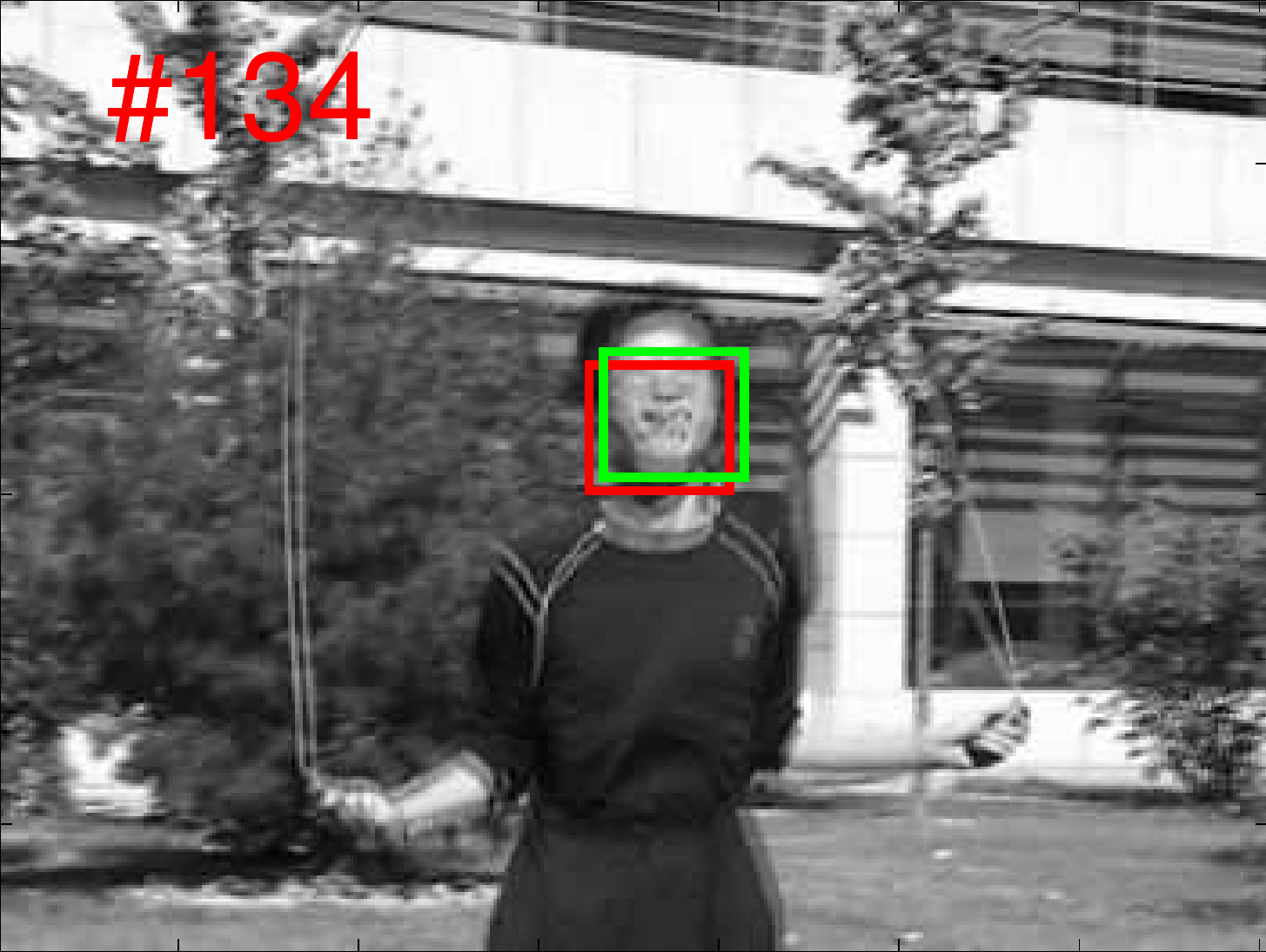}}
  \subfigure[$(2,0.8033)$]{\includegraphics[width=0.8in]{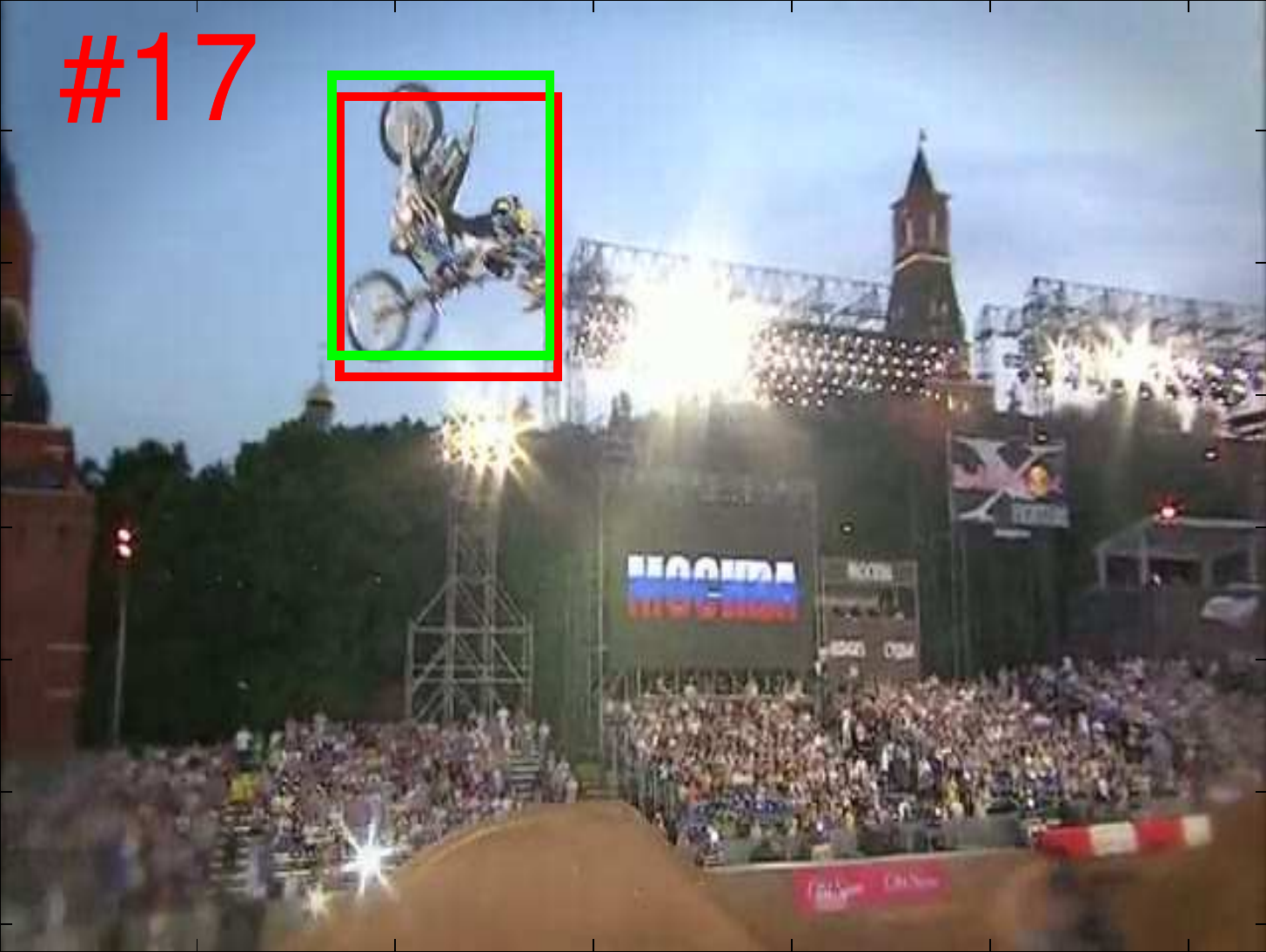}}
  \subfigure[$(2,0.8447)$]{\includegraphics[width=0.8in]{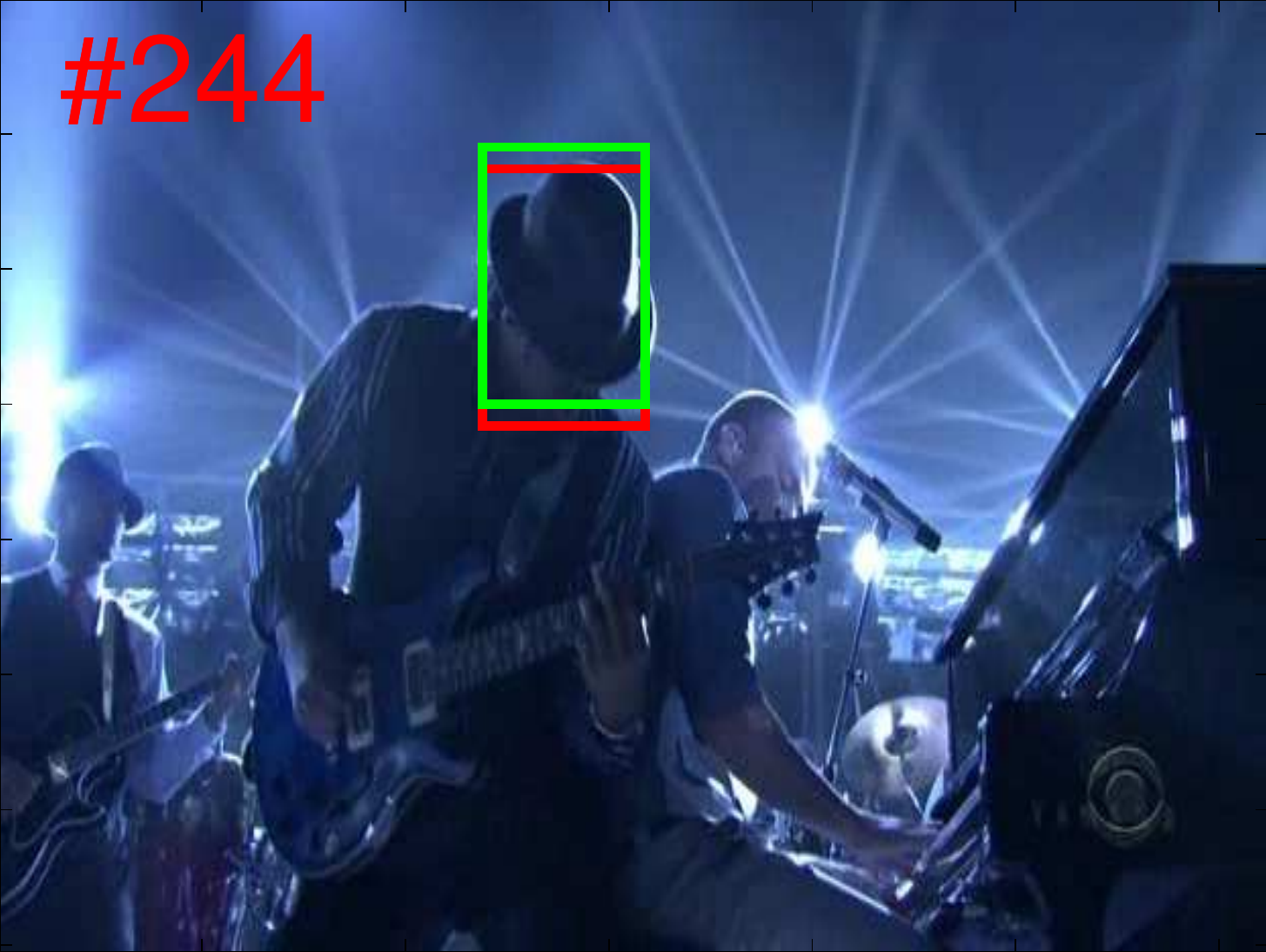}}
  \subfigure[$(2,0.5152)$]{\includegraphics[width=0.8in]{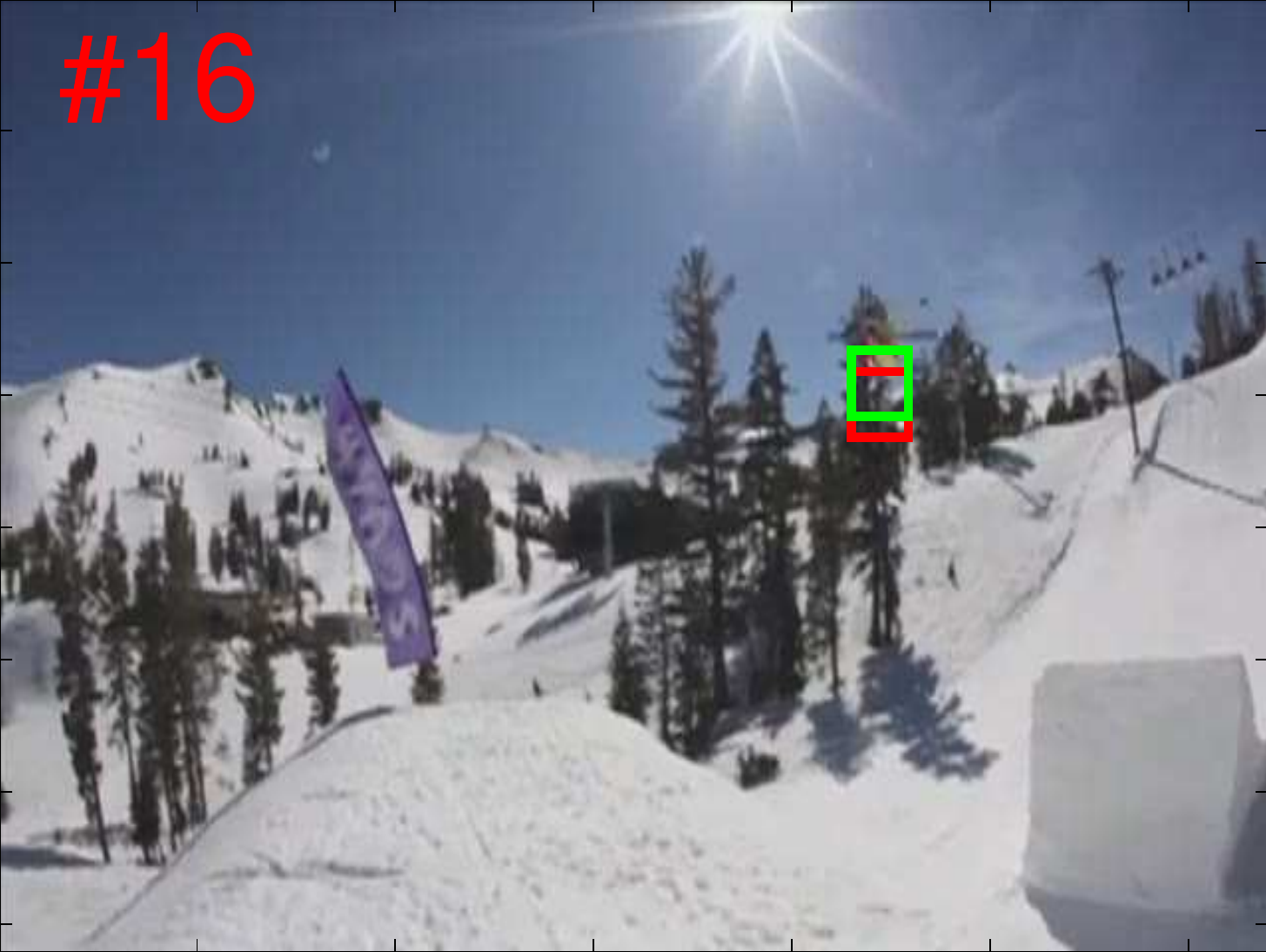}}\\
  \caption{The largest localization differences in our experiments. In each sub-caption, $(a, b)$ means the center error and overlap ratio of two Strucks are $a$ and $b$, respectively, and the red and green bounding boxes are generated by the Strucks with hinge loss and square loss, respectively. See text for details.}
  \label{fig:2strucks-worstcases}
\end{figure}

We compare two Strucks with the hinge and square losses on OTB50. In order to prevent updating schemes from affecting the location performance and keep enough yet limited difference between training and testing samples, we randomly sample 10 ground truthes from all ground truthes of each of 50 sequences, train the two Strucks with each ground truth, and locate the target object in the next third frame. Then, the center error and overlap ratio (OR) of each localization of two Strucks can be evaluated. Consequently, totally 500 center errors and ORs are acquired. Note that only ground truthes which indeed contain the target object are randomly sampled, and the chessboard distance is employed to evaluate the center error for an accurate definition of bin size in this section. Fig.~\ref{fig:2strucks-cd-iou} displays the histograms of center errors and ORs of two Strucks. It is seen that there is no error in almost $92\%$ location in both histograms. To further illustrate the similarity of locations of two Strucks, Fig.~\ref{fig:2strucks-worstcases} gives several frames with largest differences in localization. It is noticed that the locations of two Strucks are really close to each other even in the worst cases. Consequently, it is concluded that the hinge loss and square one are almost identical to each other in the context of regression.

\subsection{Consistency Comparison}
\label{sec:consistencycomparison}
To exactly focus on the performance of localization, there is no scale estimation in our experiments except for the ones in Sec.~\ref{sec:struckvscflbmcwithscale}. Therefore, the scale estimation part of SRDCF is turned off temporally. And the trackers compared in this section are denoted as SRDCF-1, CFLBMC-1, and Struck-L-1, where 1 and L mean without scale estimation and with linear kernel, respectively.

\begin{figure}[t]
  \centering
  \includegraphics[width=1.7in]{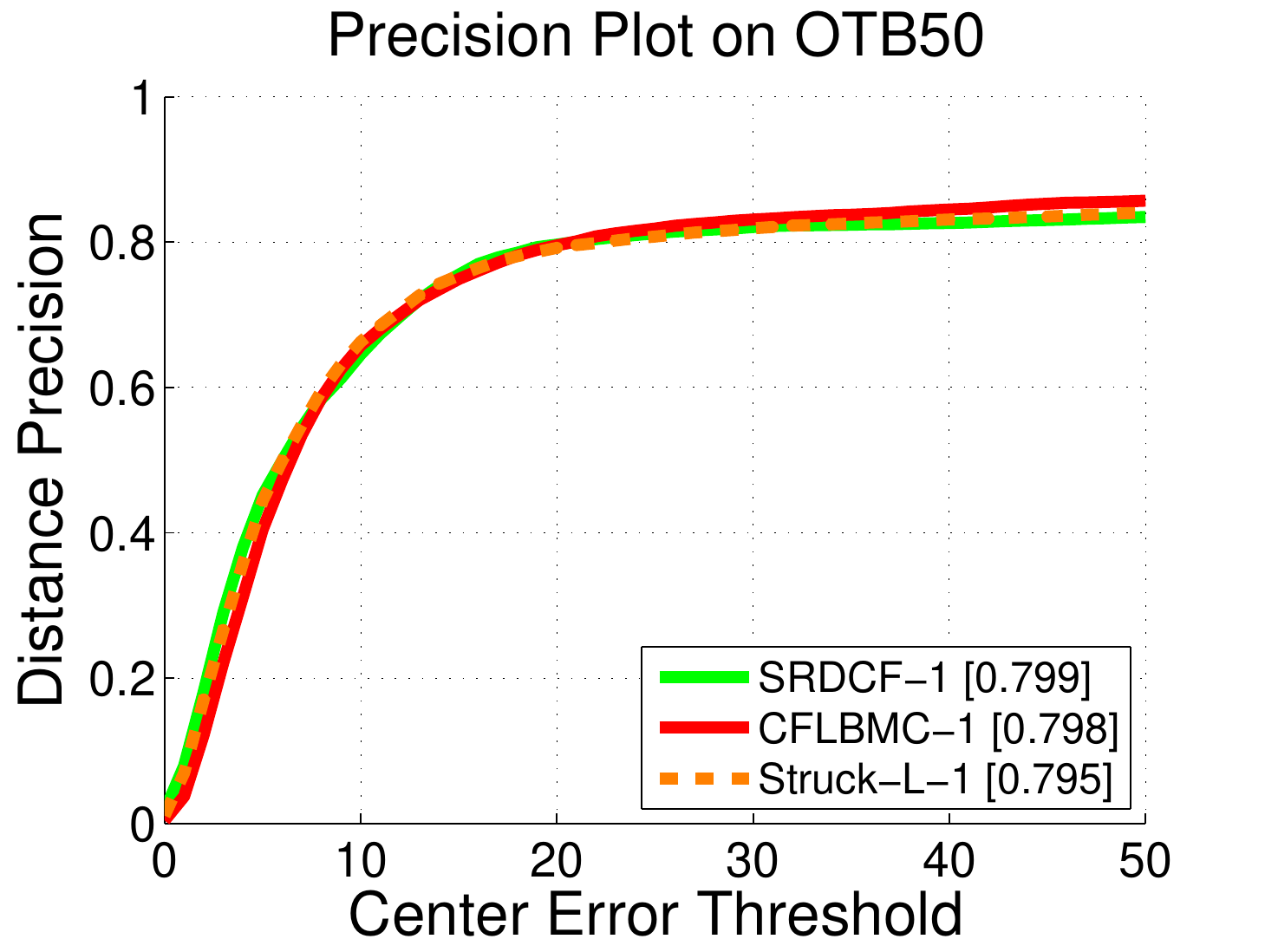}
  \includegraphics[width=1.7in]{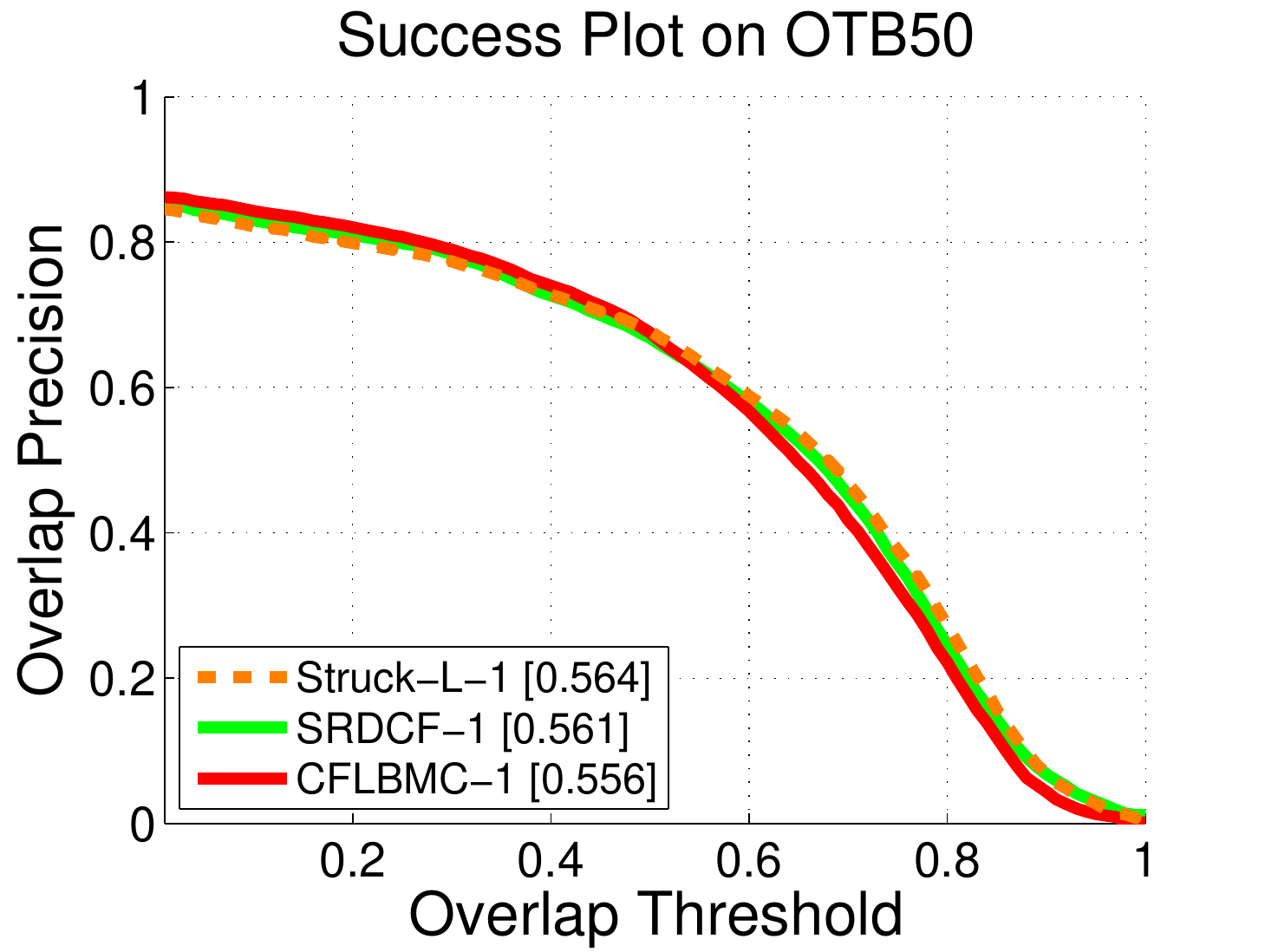}
  \caption{The average precision plots and success plots of the three trackers on OTB50. SRDCF-1, CFLBMC-1, and Struck-L-1 denote new versions of SRDCF, CFLBMC, and the modified Struck, respectively. These new versions are with line kernel and without scale estimation. The mean distance precision scores and AUCs of the trackers are reported in the legend.}
  \label{fig:noscale50}
\end{figure}

\begin{figure*}
  \centering
   \subfigure[]{\includegraphics[width=1.6in]{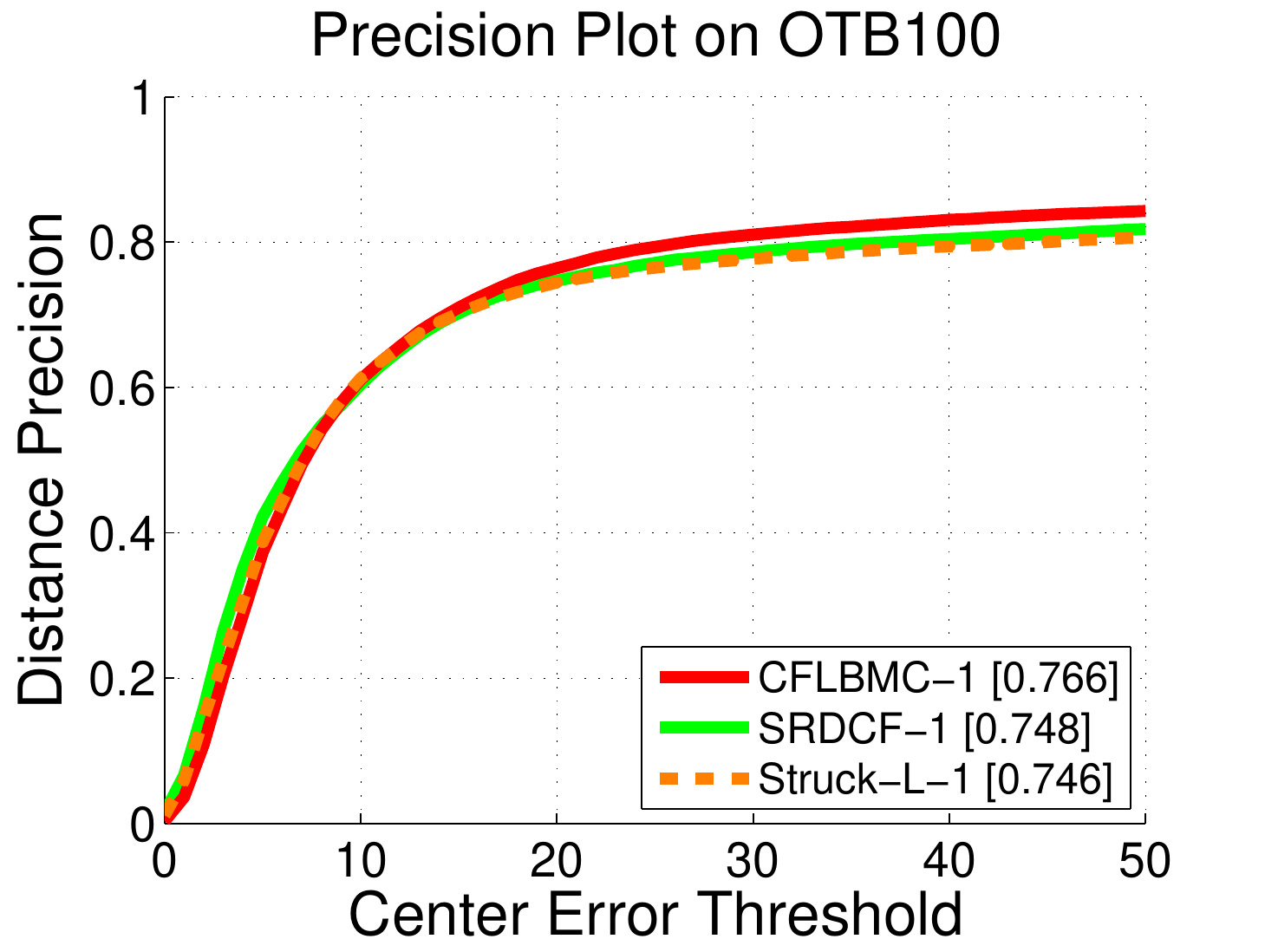}}
    \subfigure[]{\includegraphics[width=1.6in]{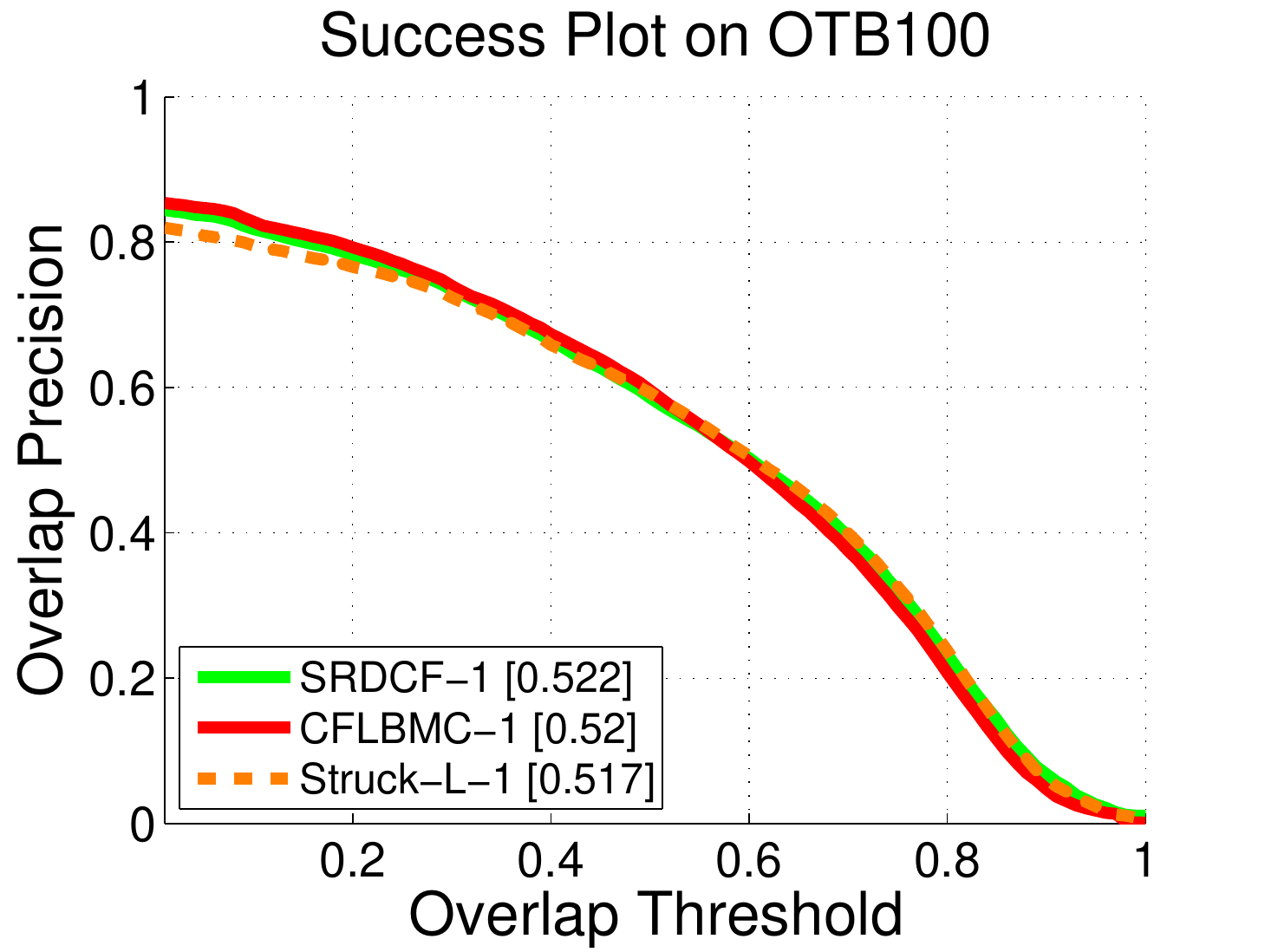}}
    \subfigure[]{\includegraphics[width=1.6in]{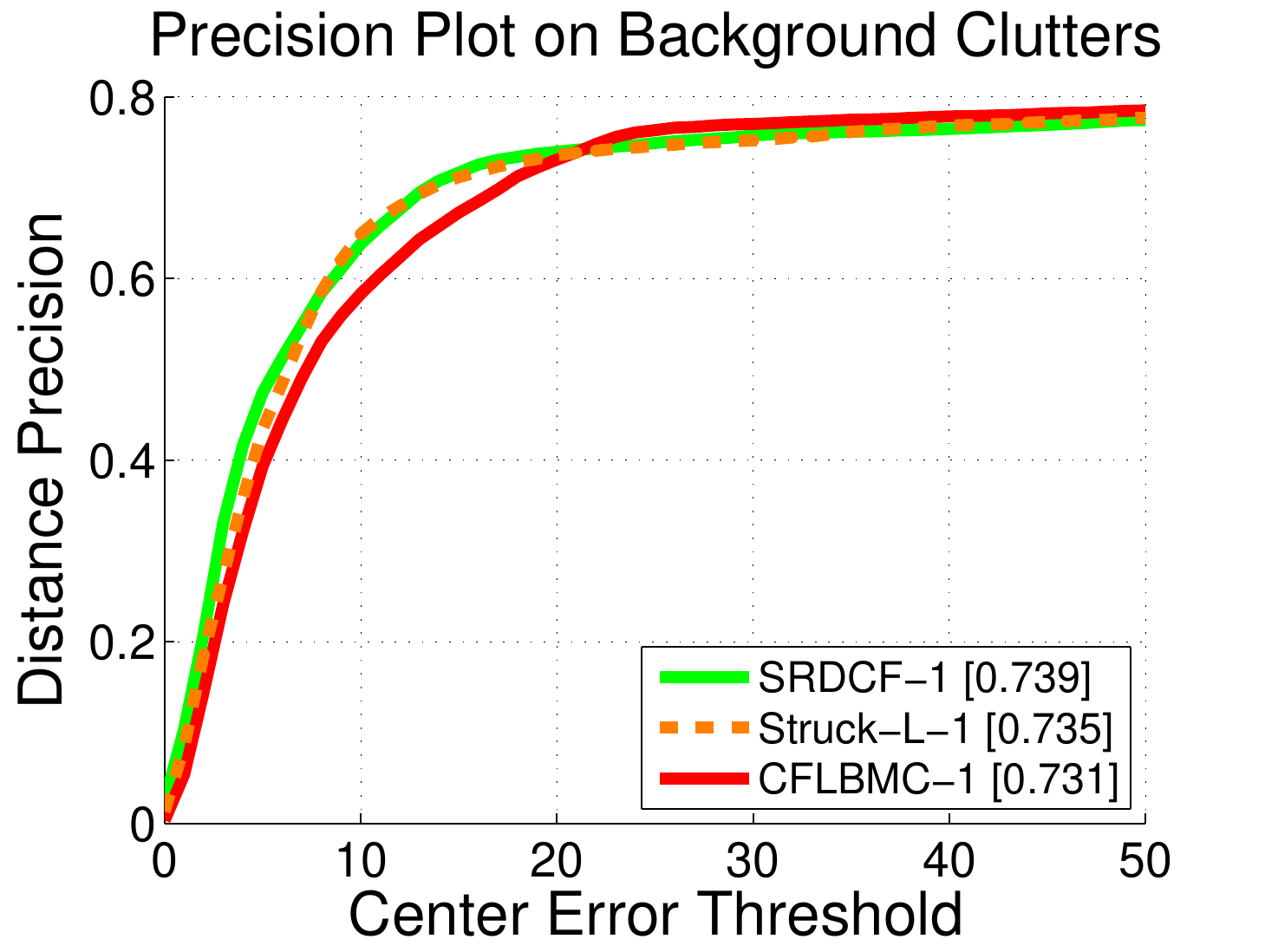}}
    \subfigure[]{\includegraphics[width=1.6in]{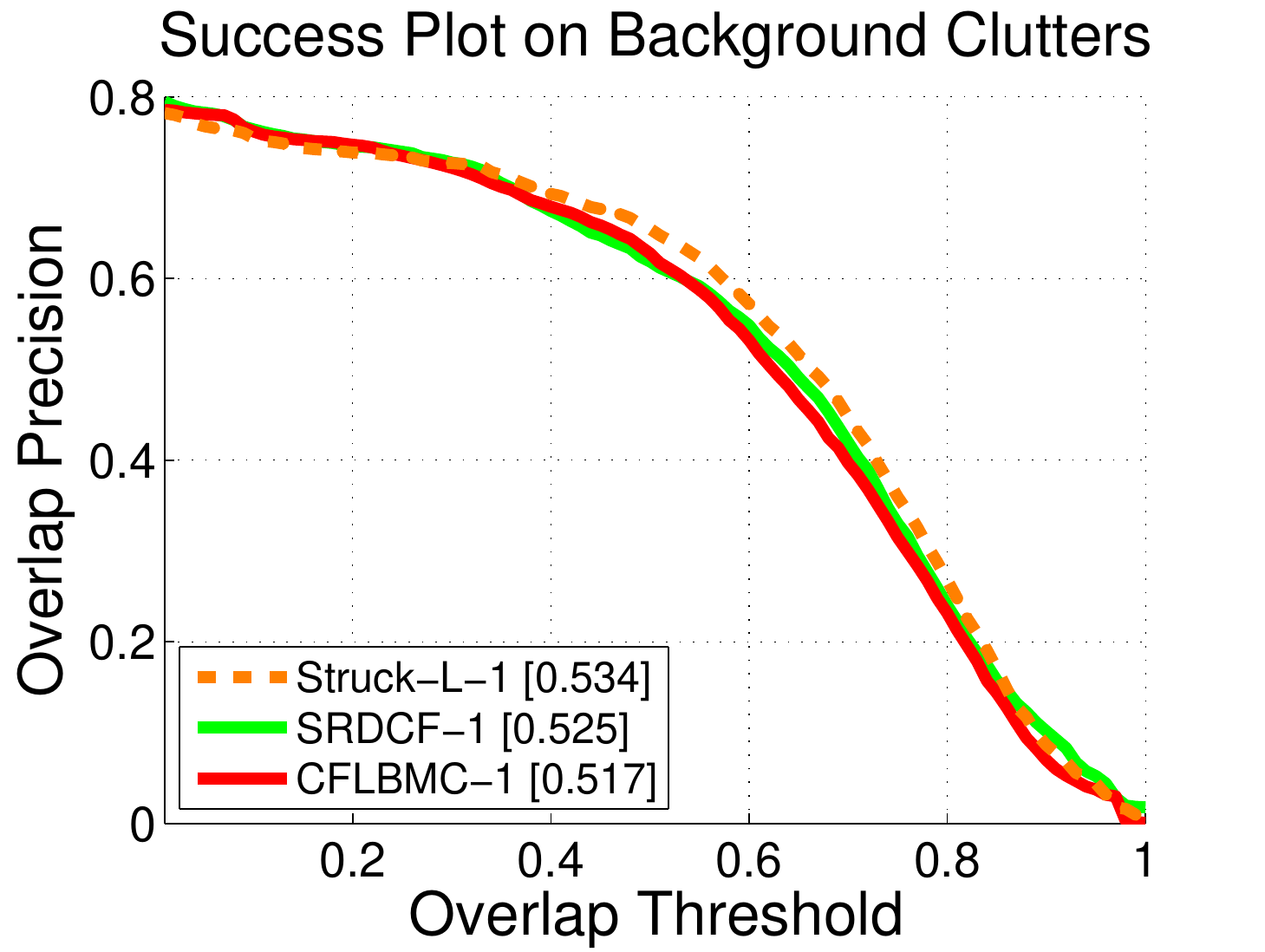}}
    \subfigure[]{\includegraphics[width=1.6in]{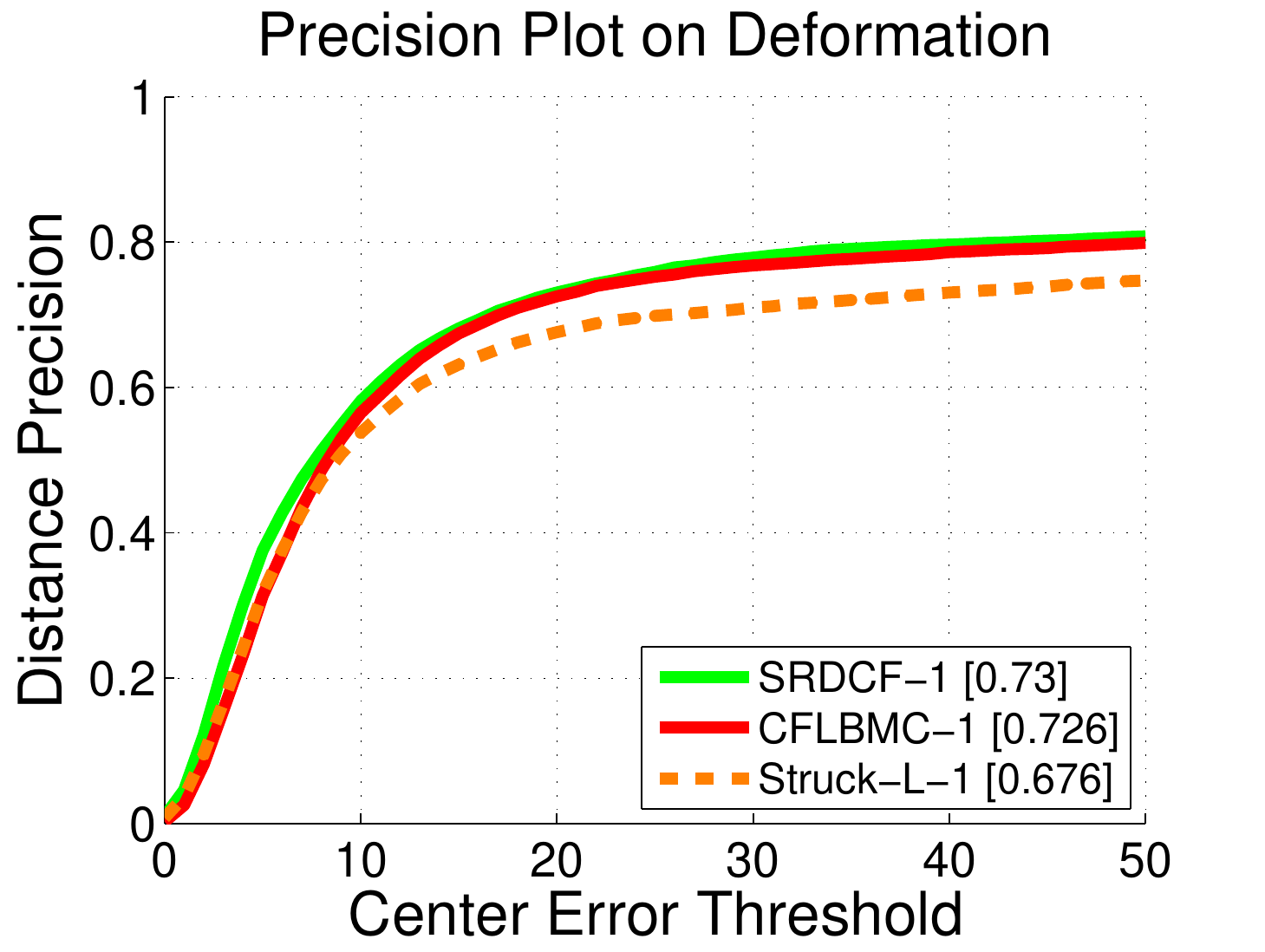}}
  \subfigure[]{\includegraphics[width=1.6in]{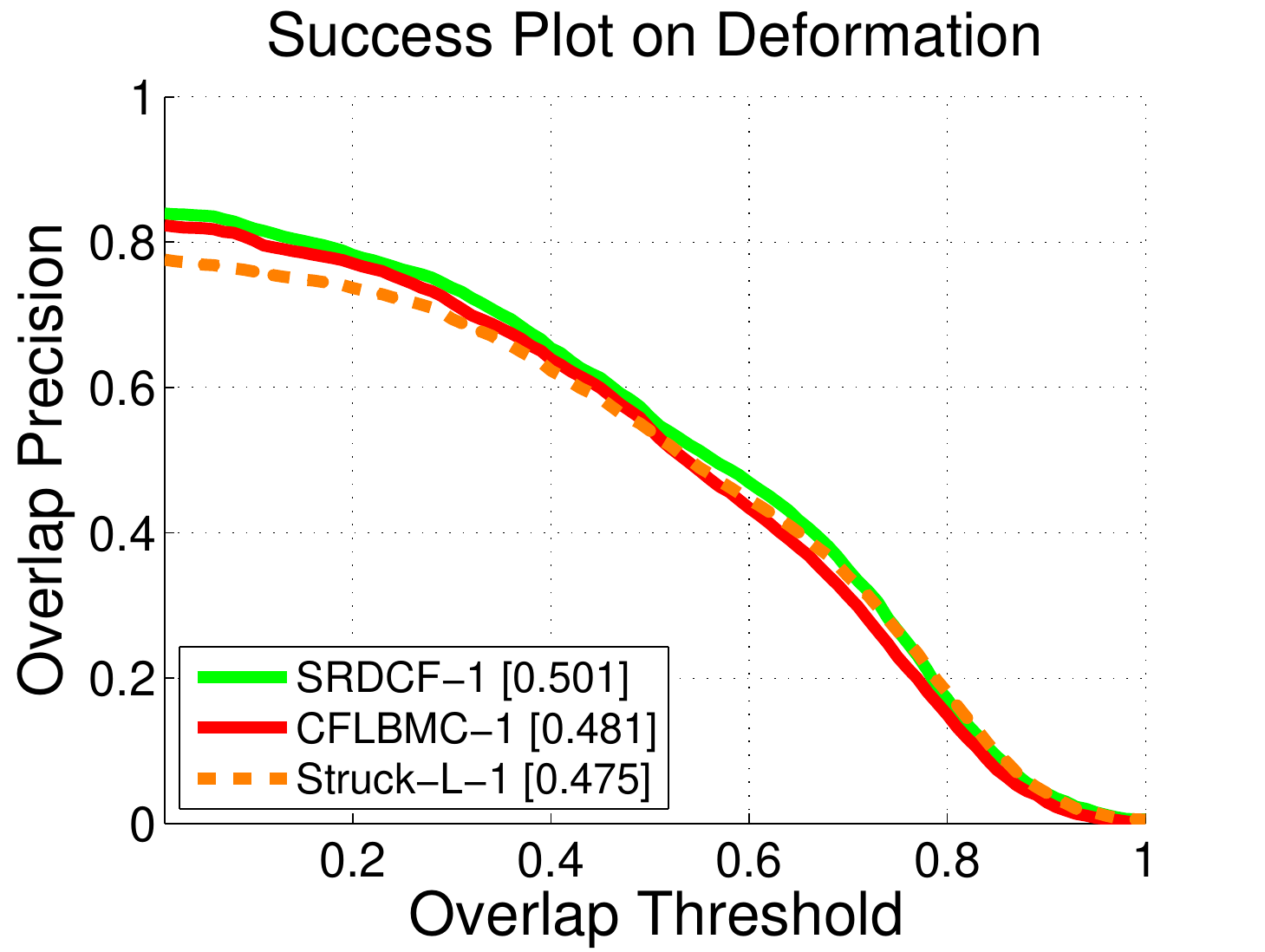}}
  \subfigure[]{\includegraphics[width=1.6in]{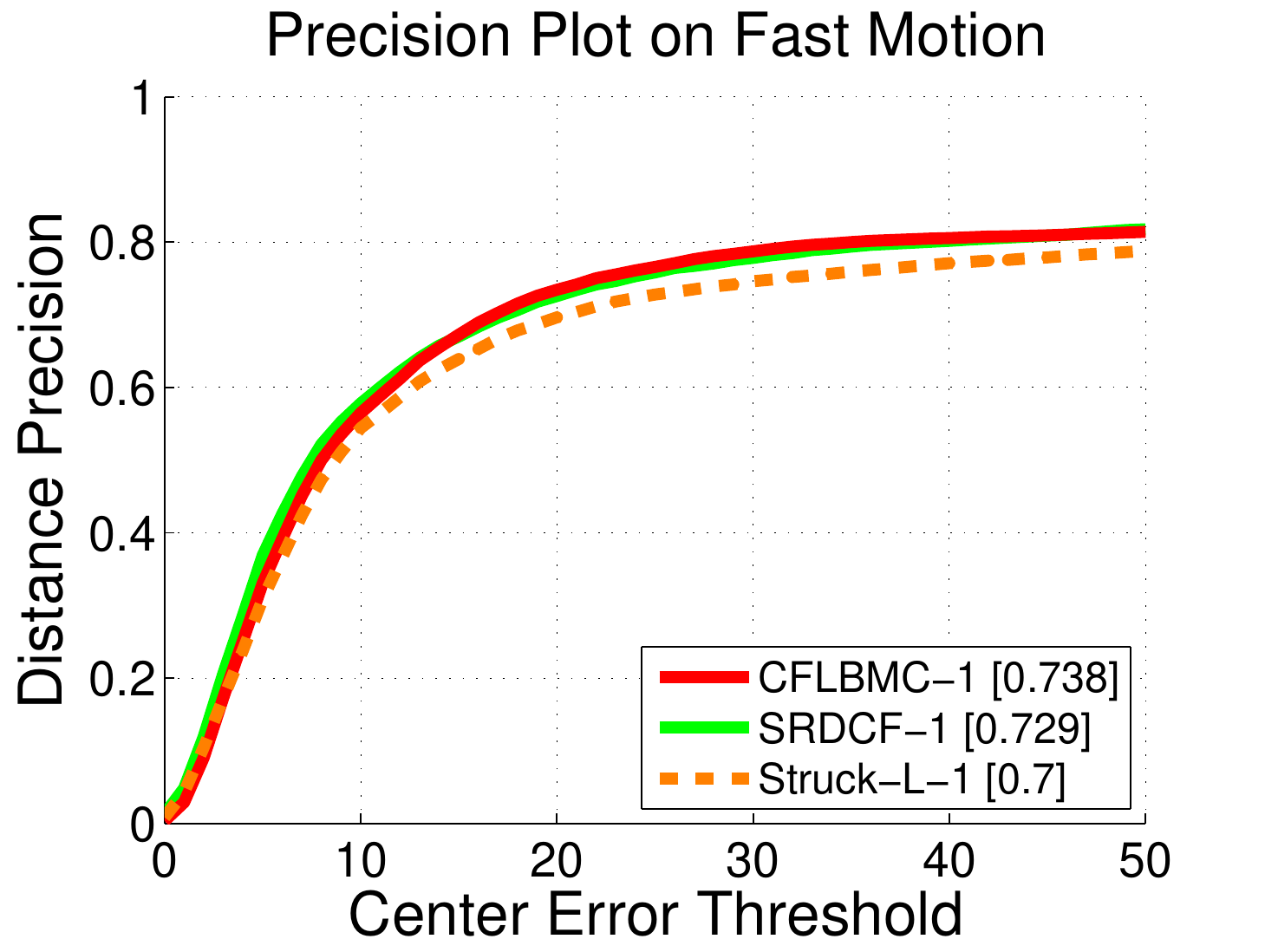}}
  \subfigure[]{\includegraphics[width=1.6in]{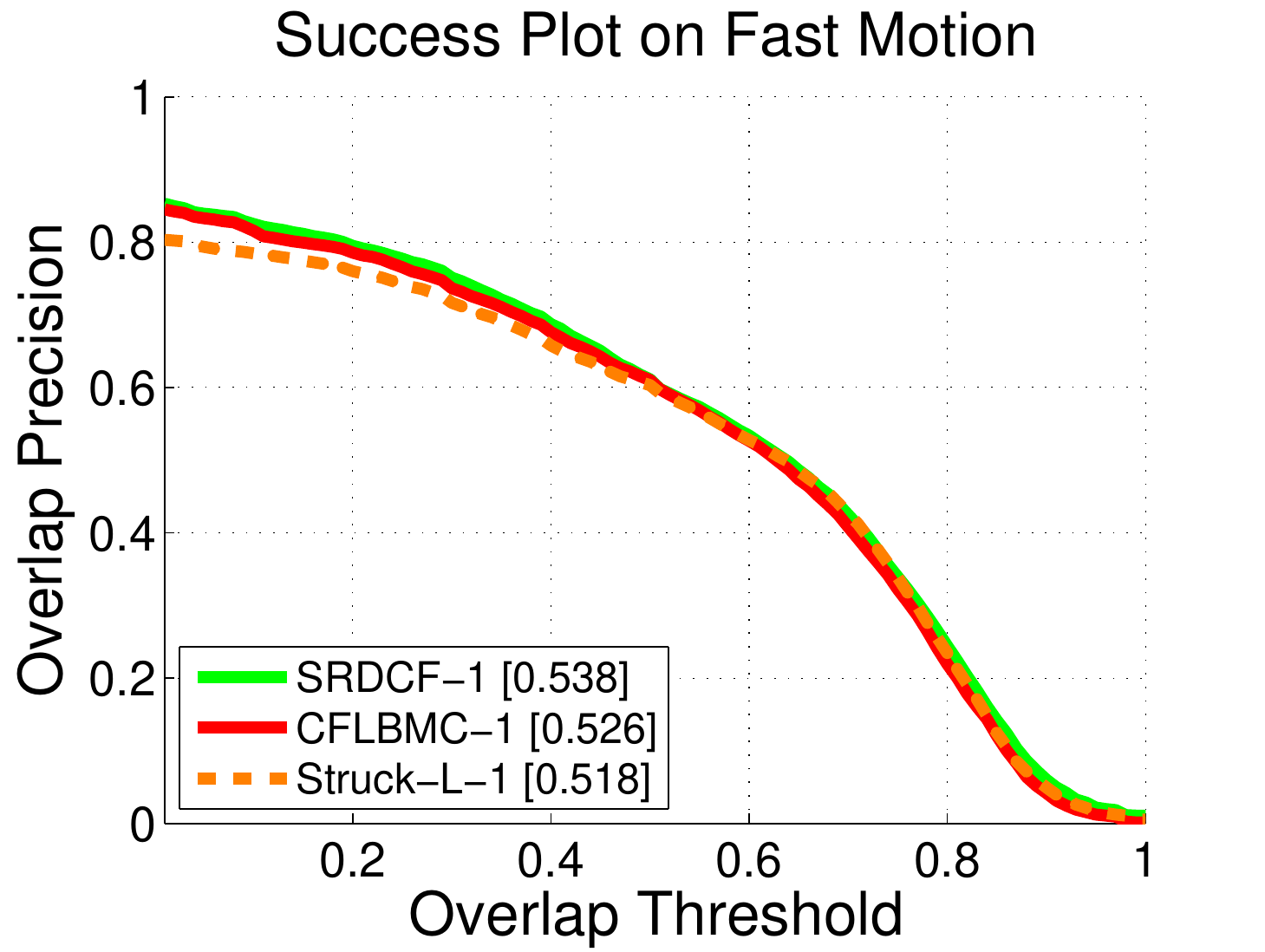}}

  \subfigure[]{\includegraphics[width=1.6in]{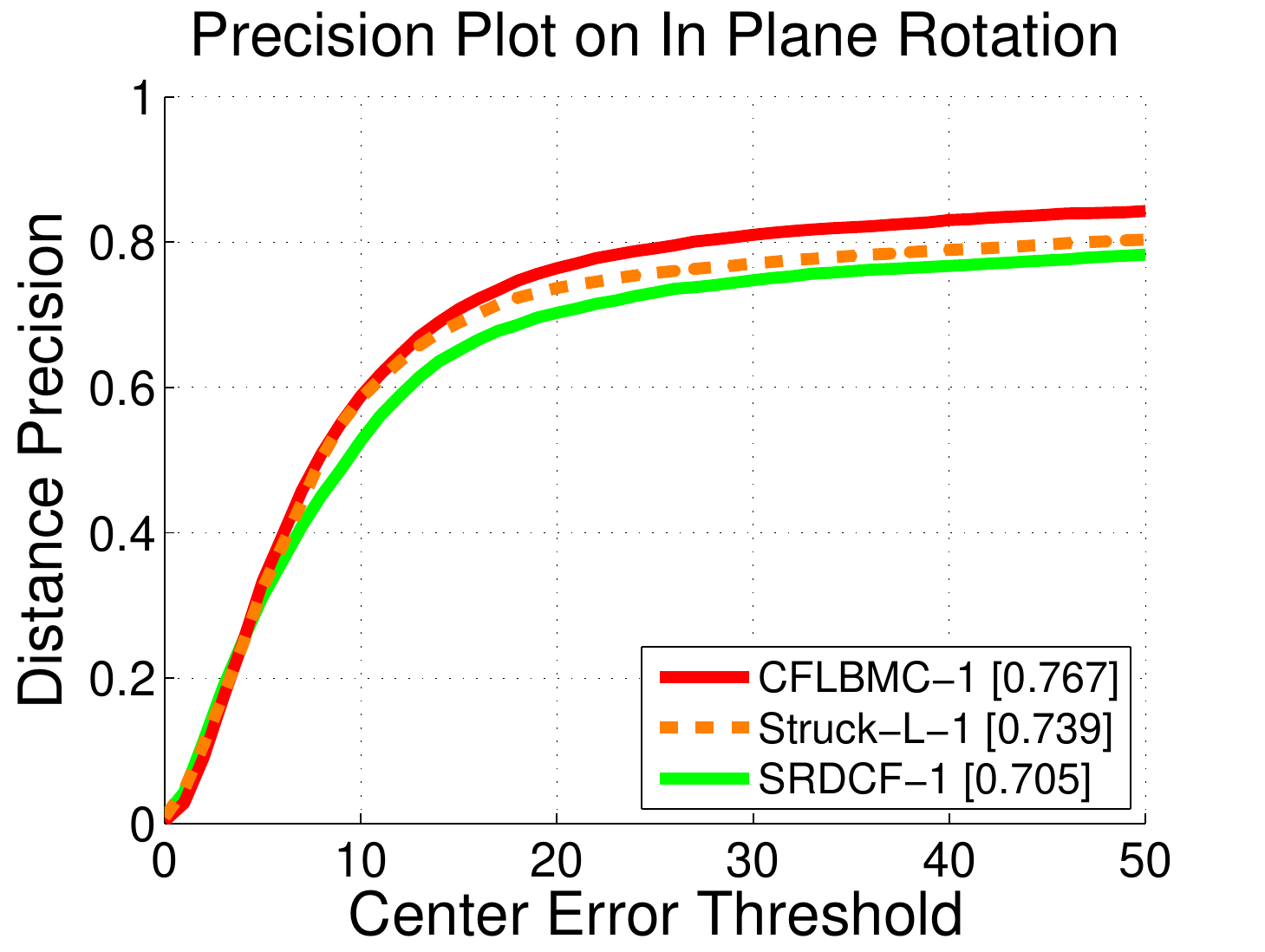}}
  \subfigure[]{\includegraphics[width=1.6in]{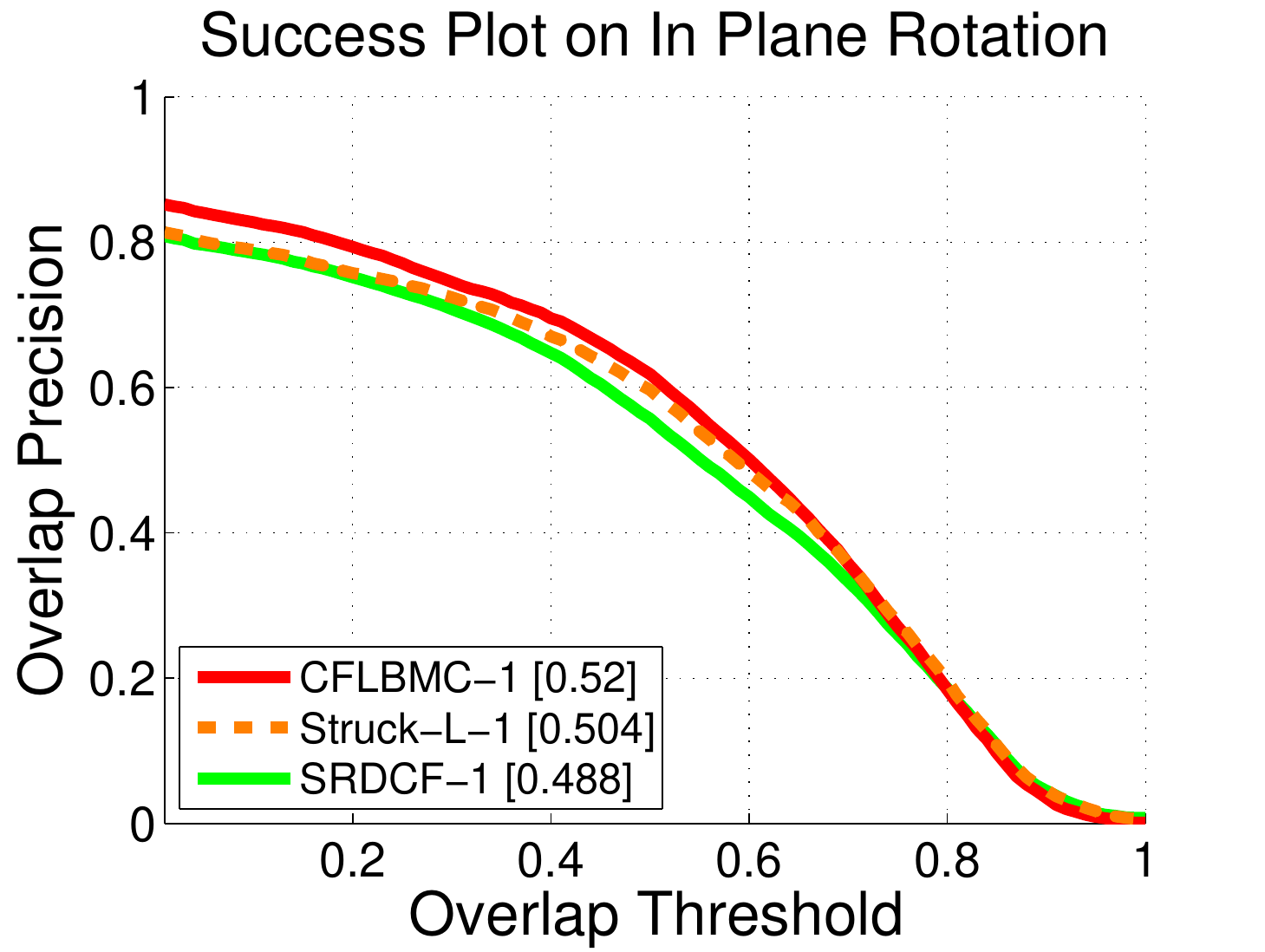}}
   \subfigure[]{\includegraphics[width=1.6in]{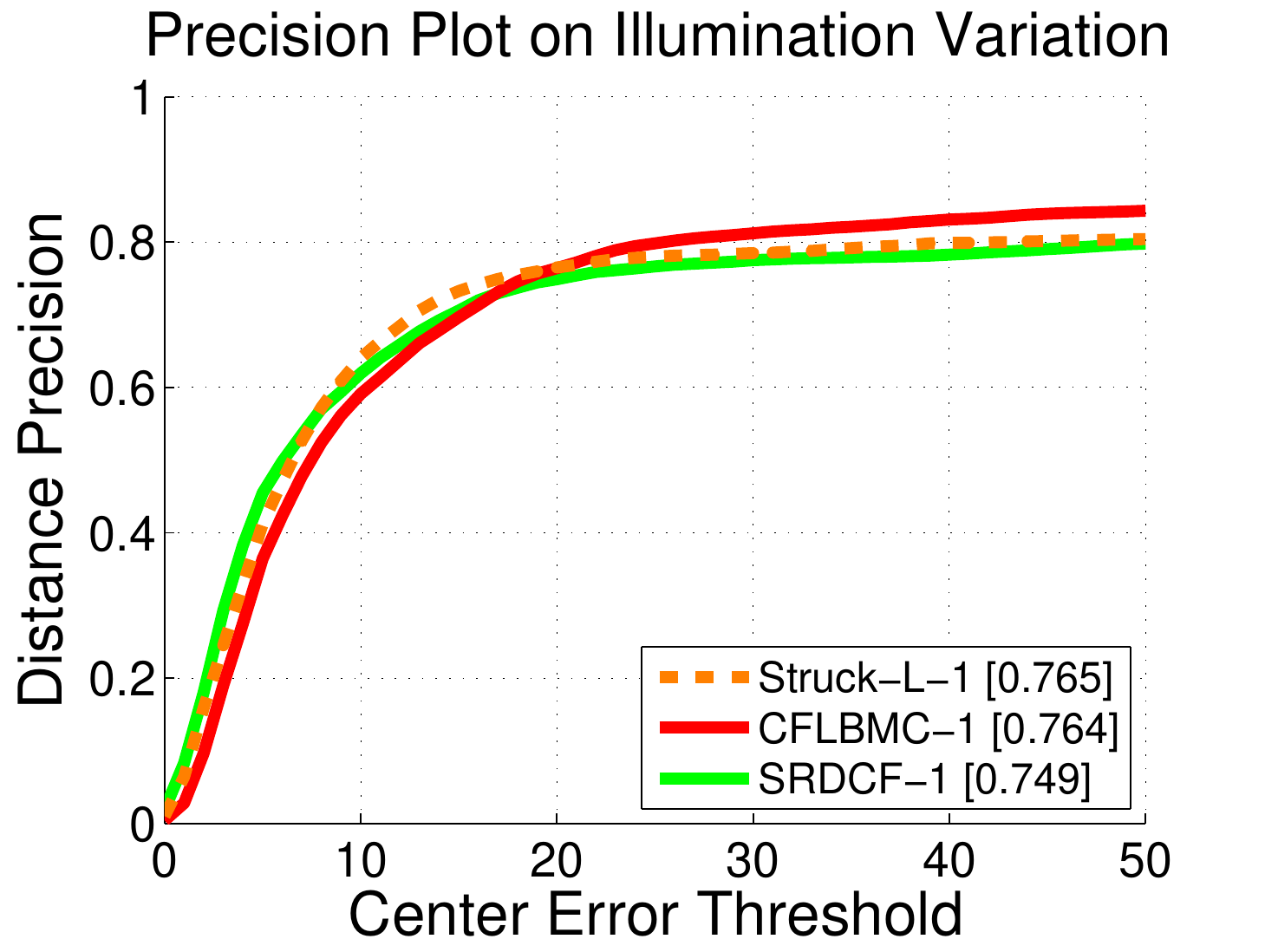}}
  \subfigure[]{\includegraphics[width=1.6in]{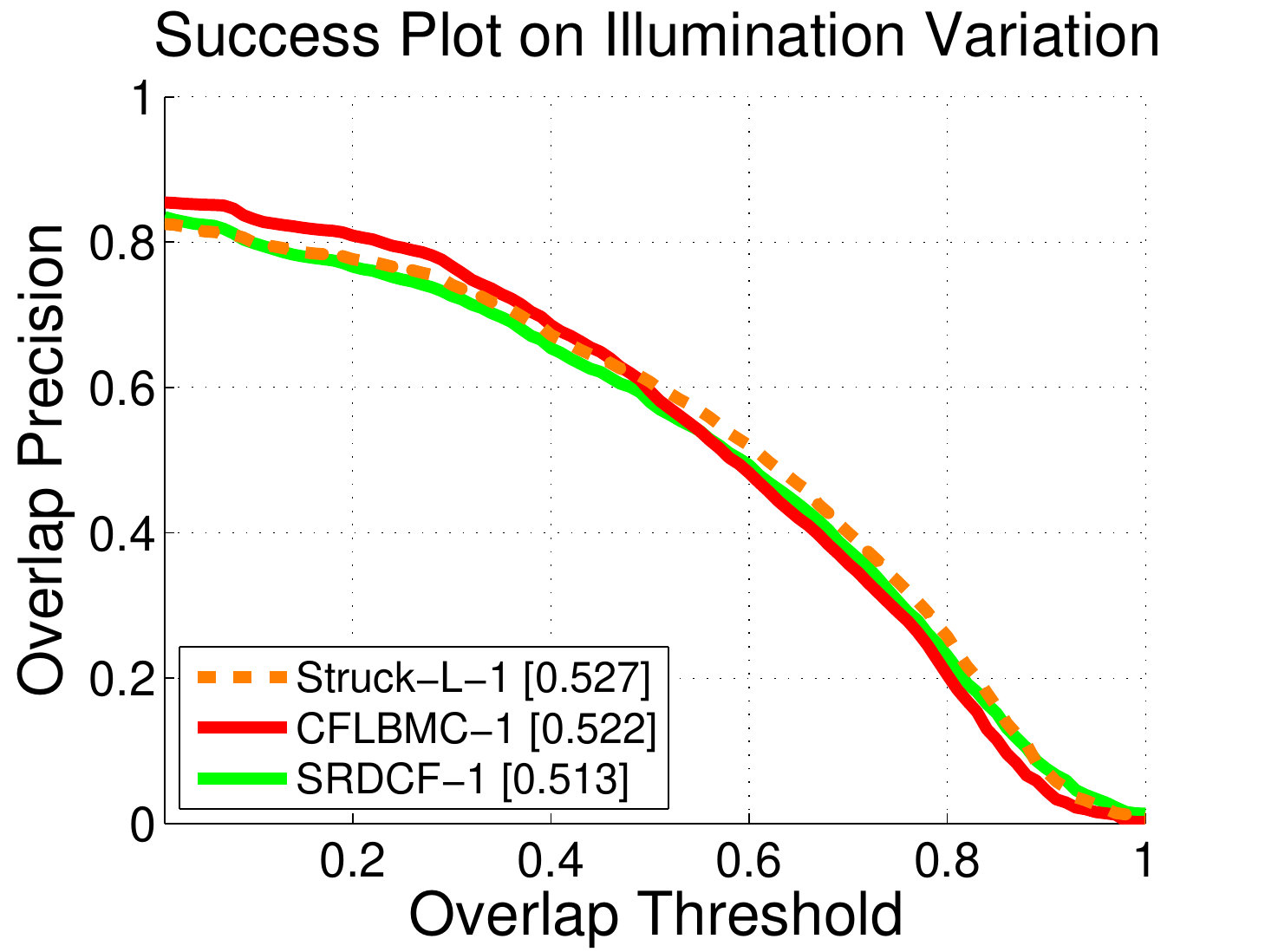}}

   \subfigure[]{ \includegraphics[width=1.6in]{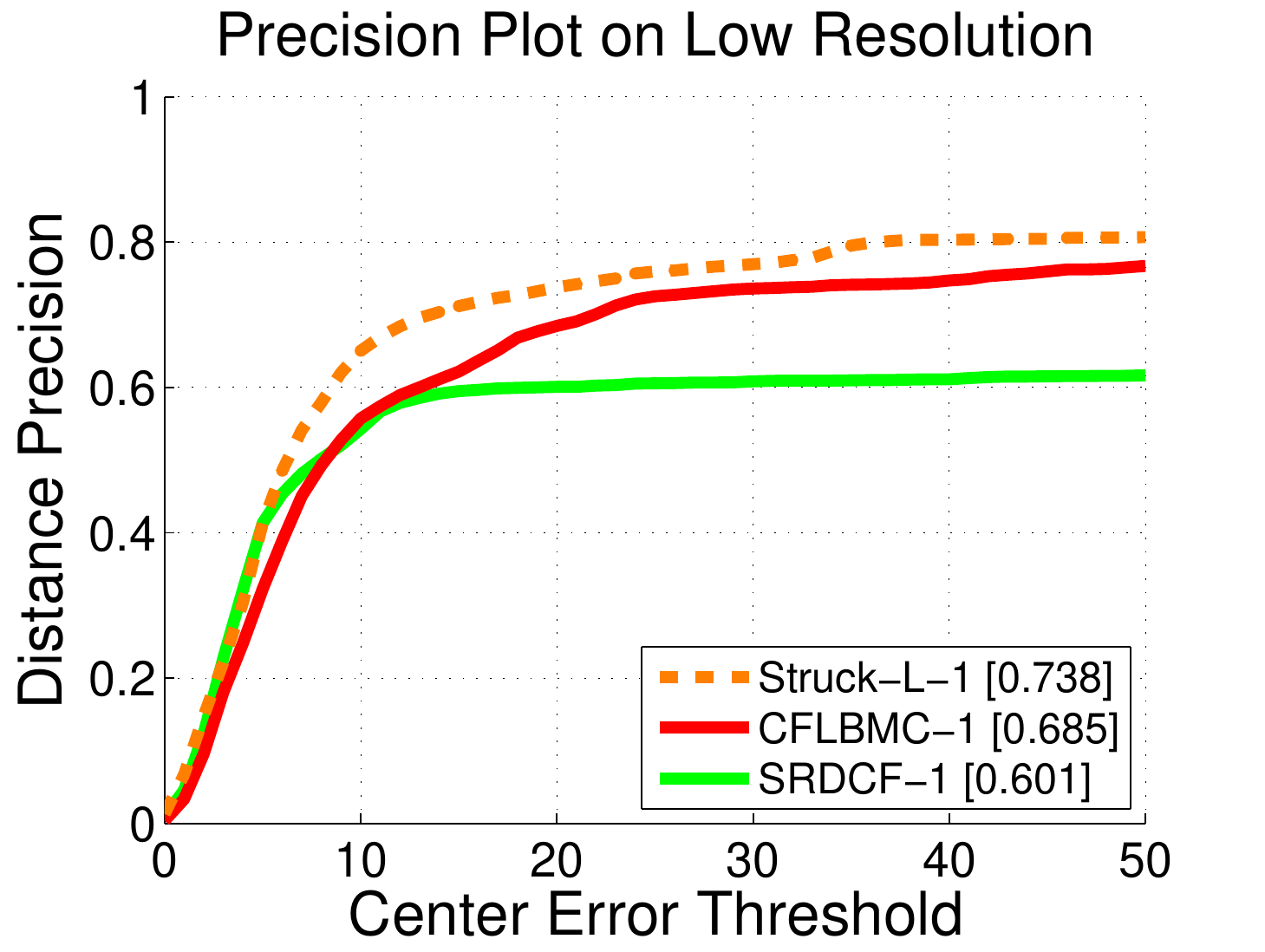}}
  \subfigure[]{\includegraphics[width=1.6in]{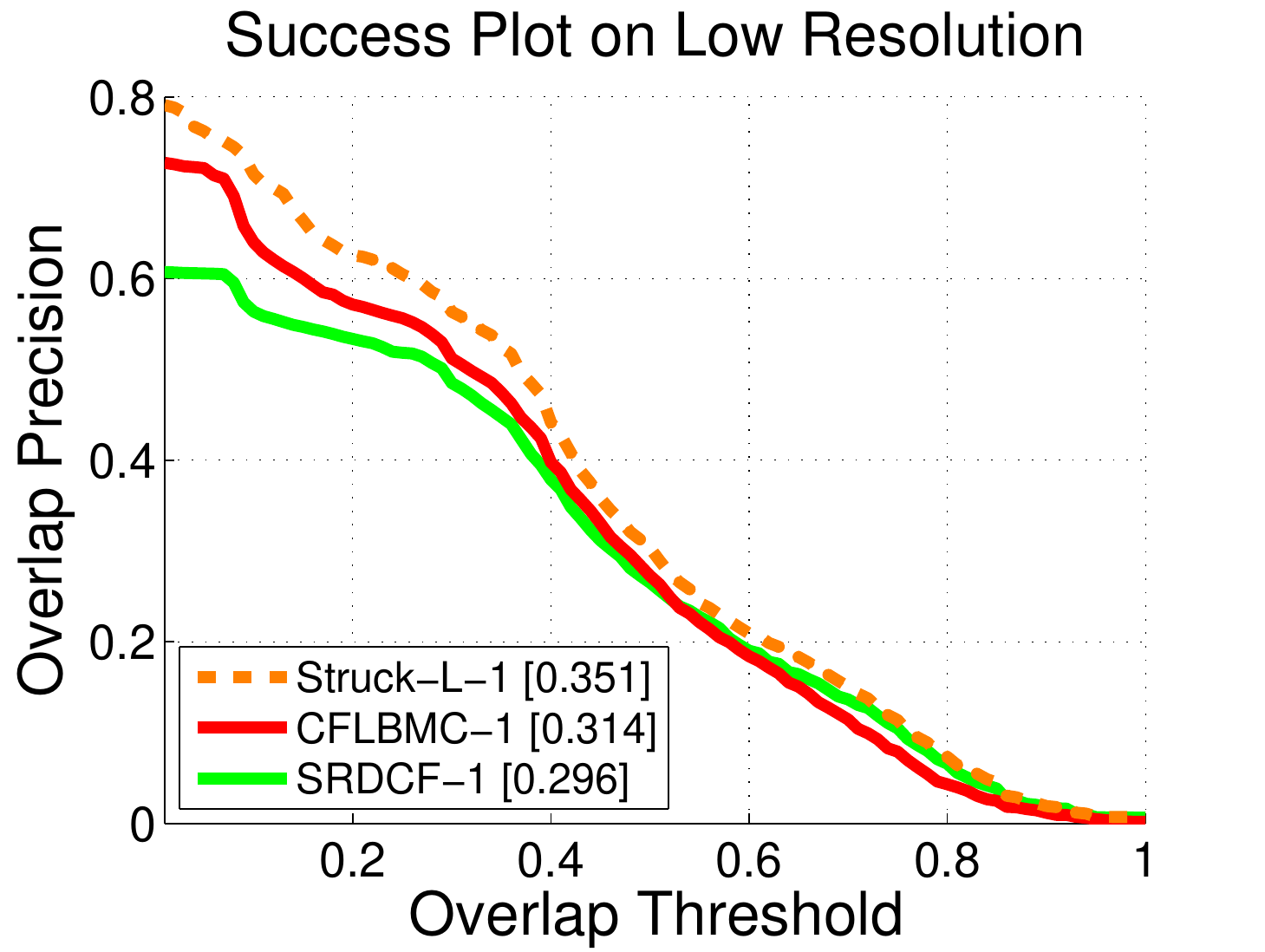}}
   \subfigure[]{ \includegraphics[width=1.6in]{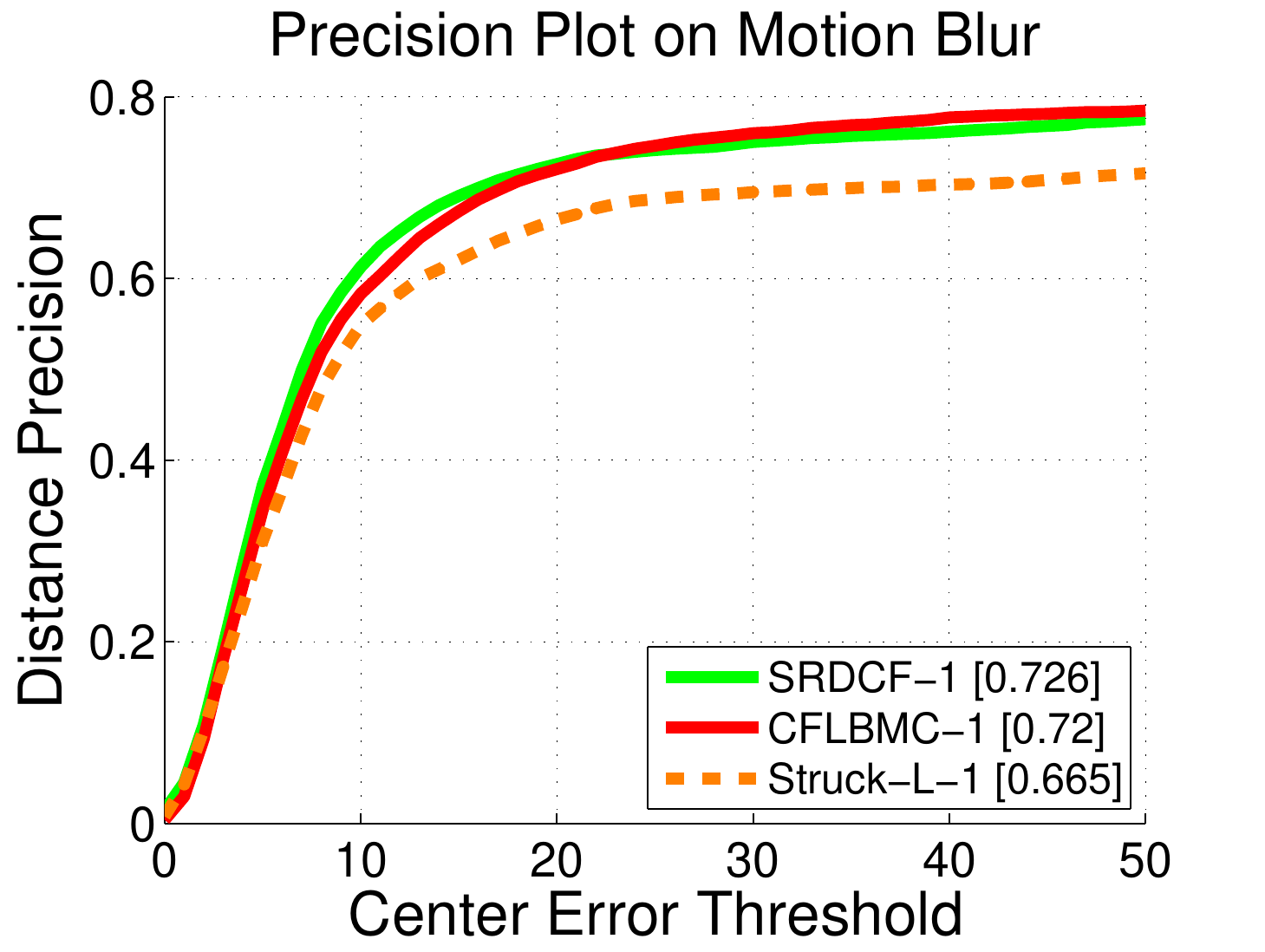}}
  \subfigure[]{\includegraphics[width=1.6in]{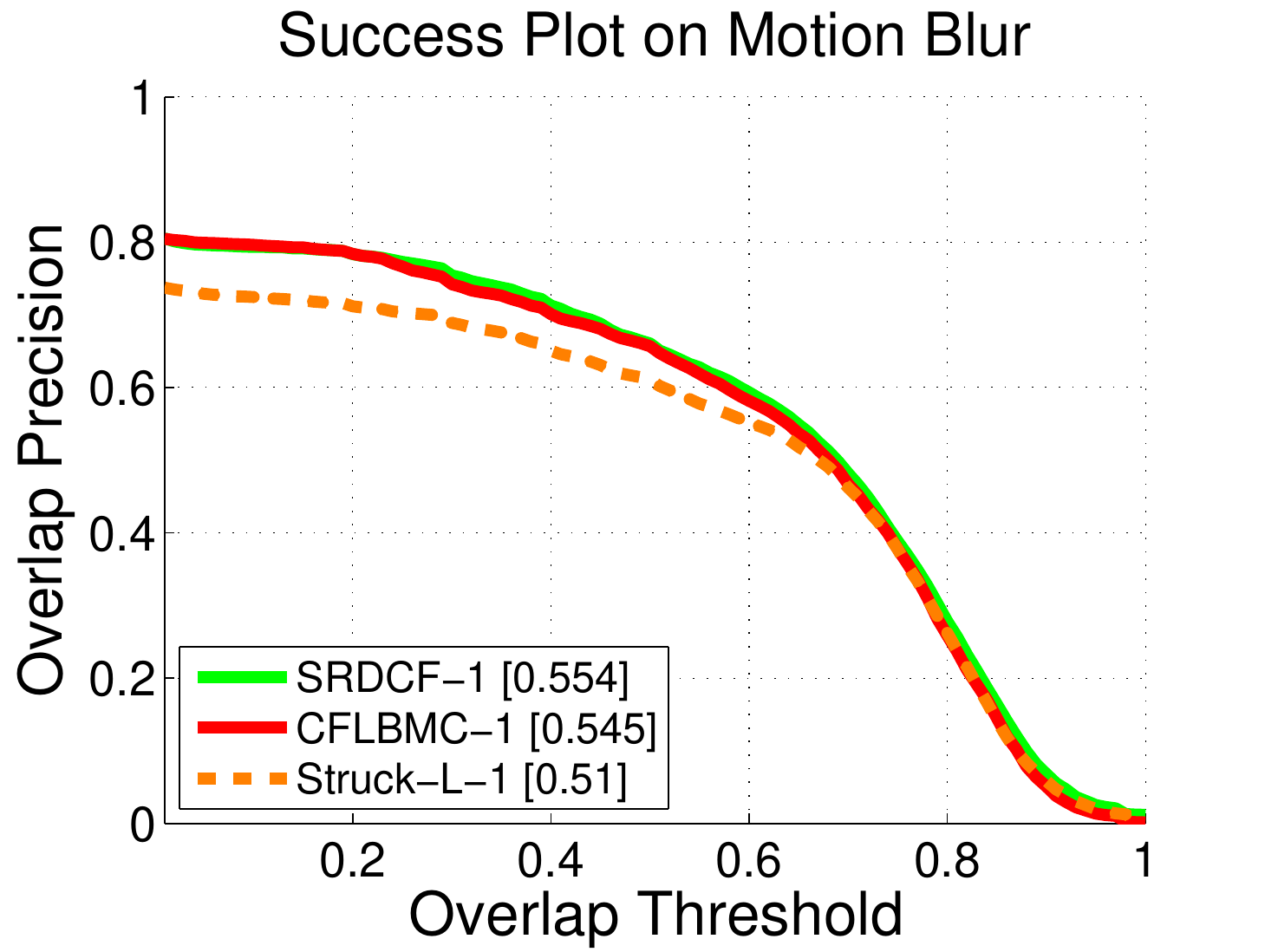}}

    \subfigure[]{\includegraphics[width=1.6in]{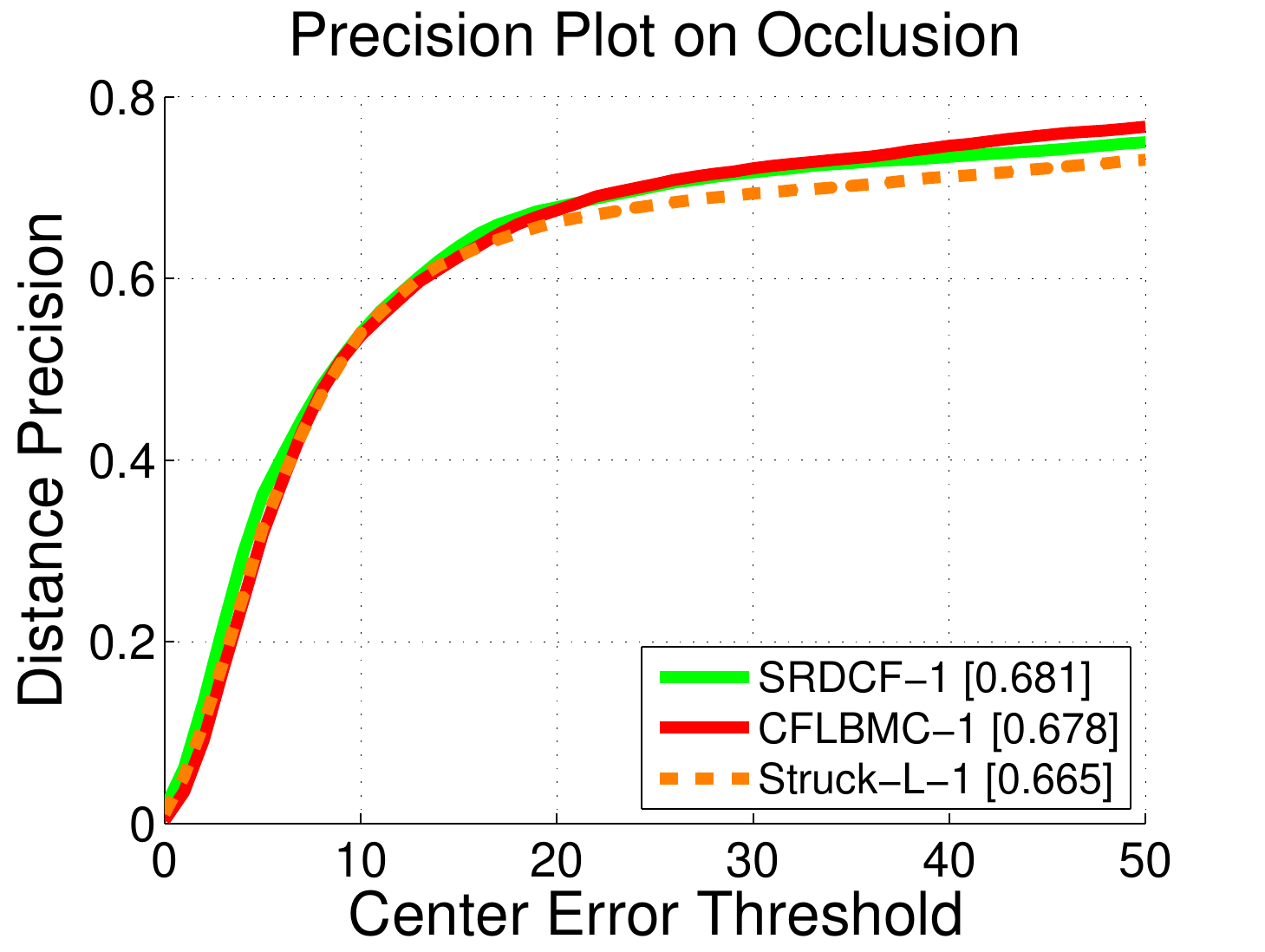}}
  \subfigure[]{\includegraphics[width=1.6in]{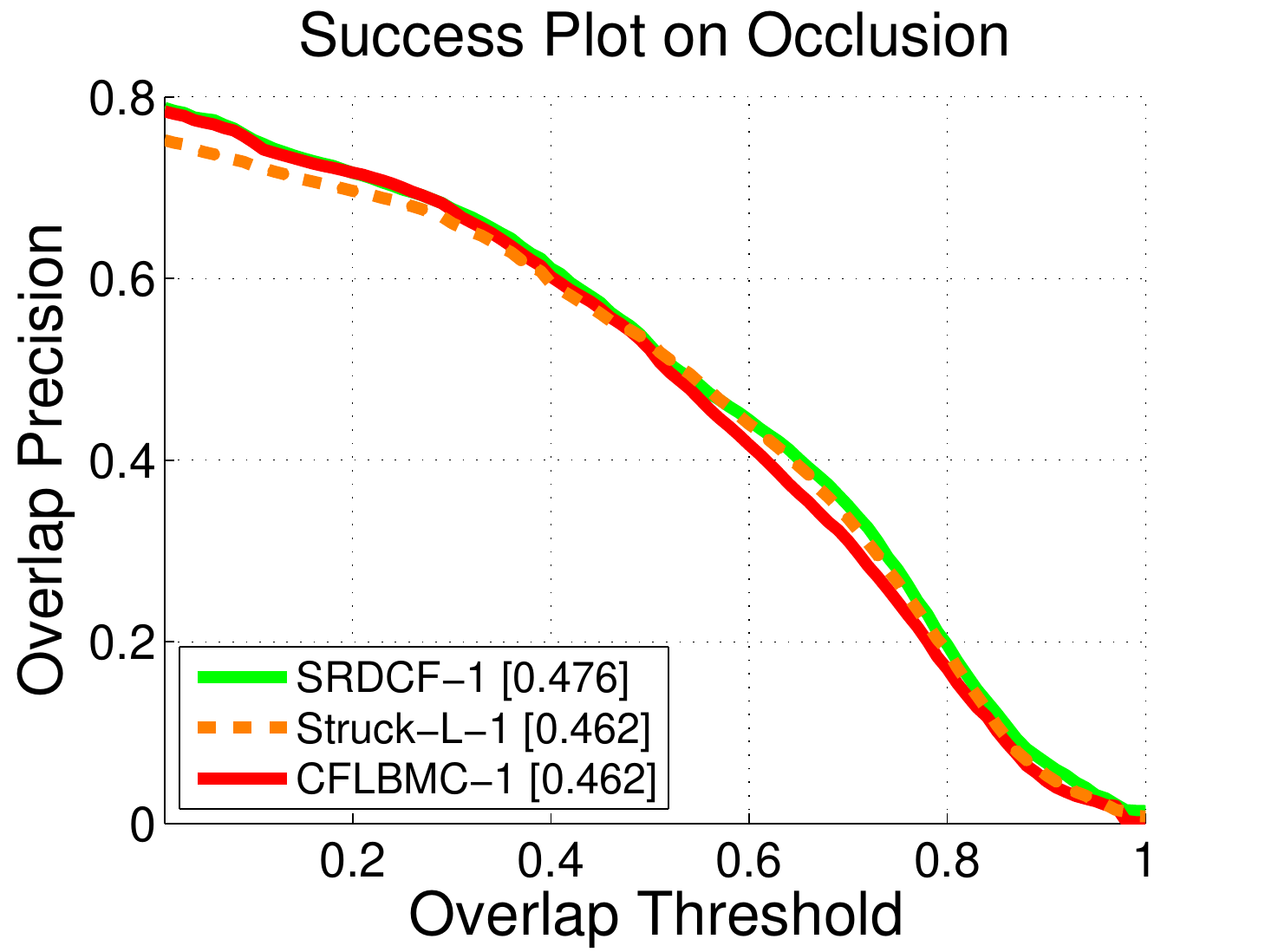}}
   \subfigure[]{ \includegraphics[width=1.6in]{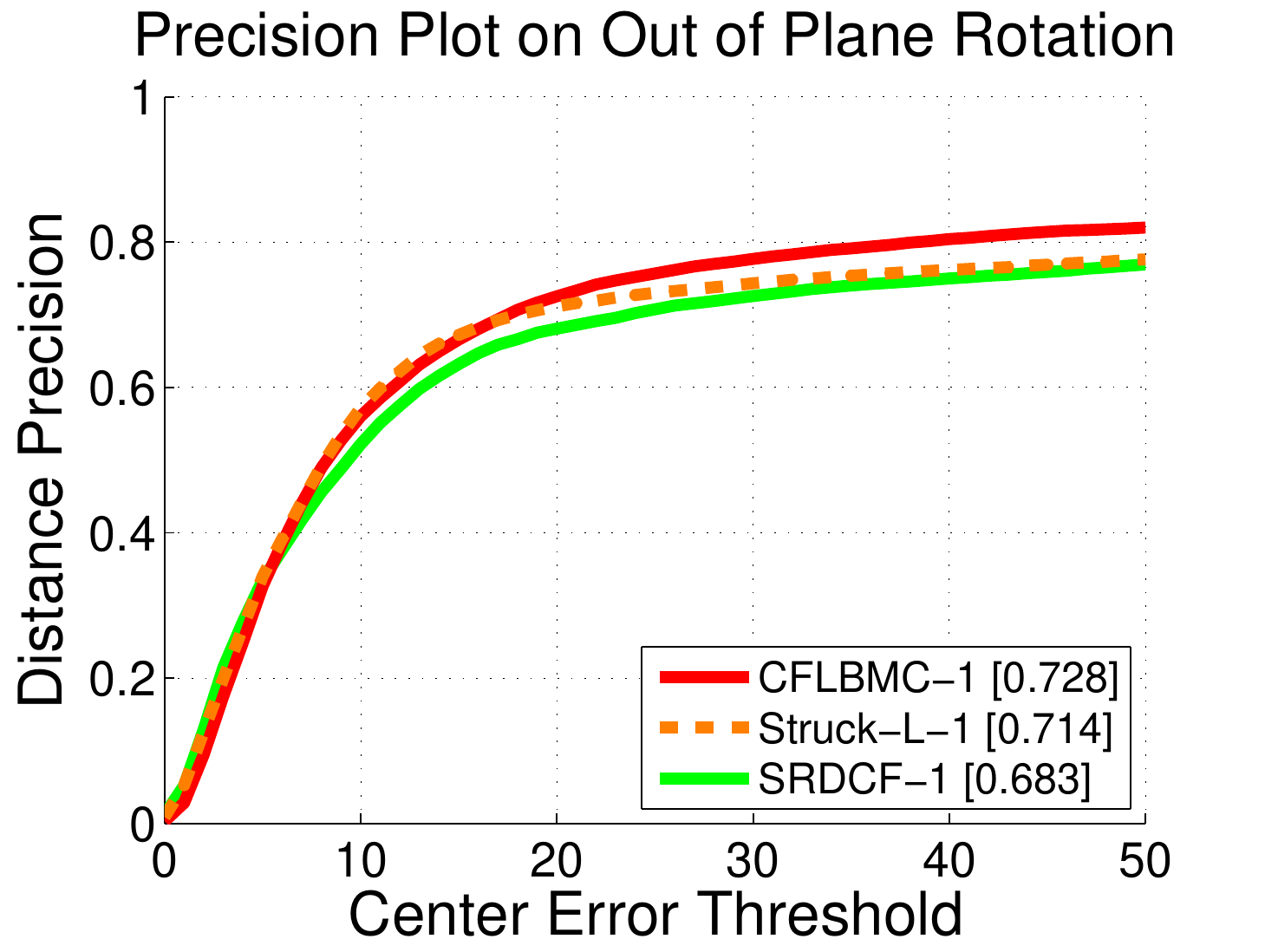}}
  \subfigure[]{\includegraphics[width=1.6in]{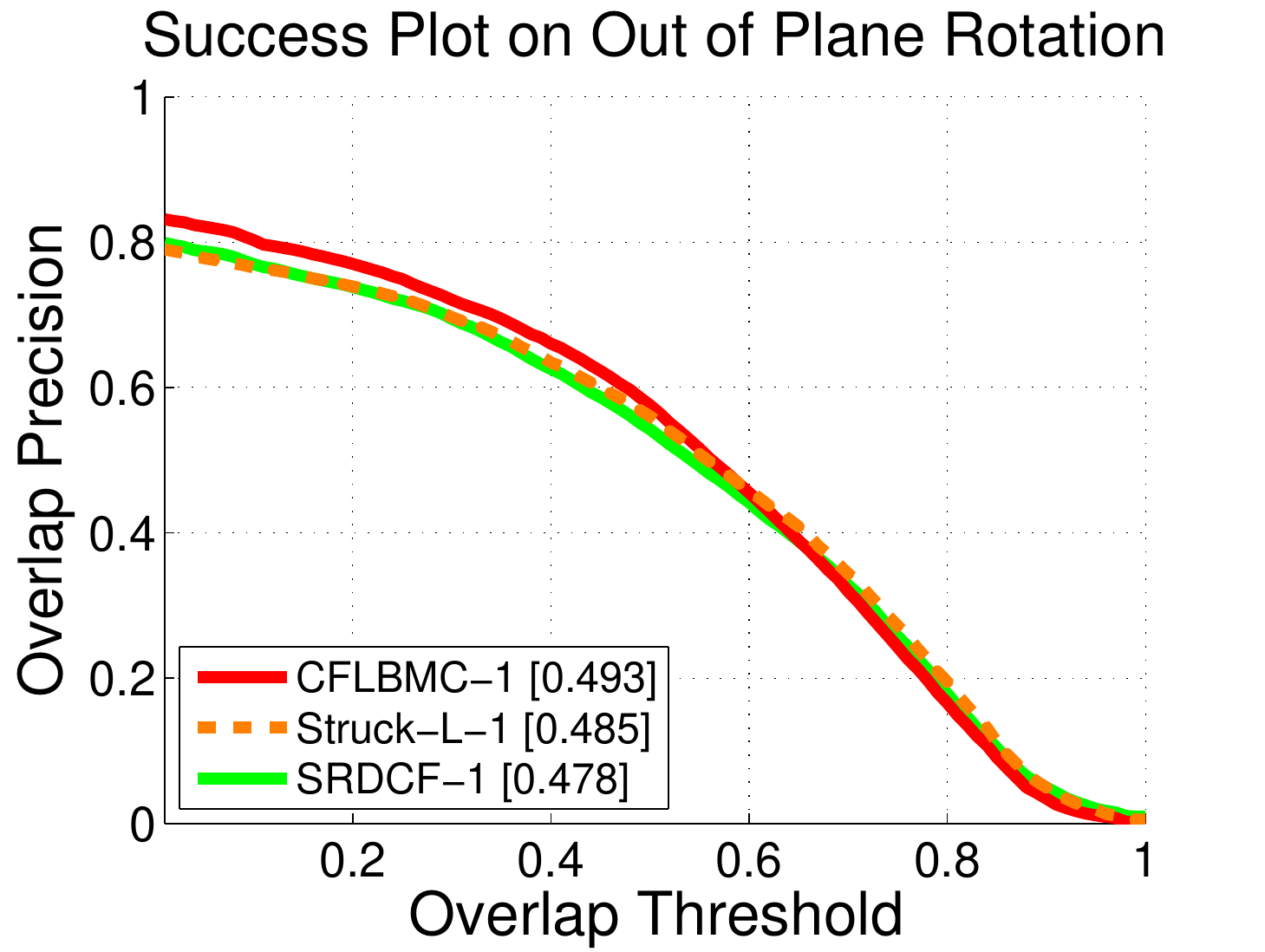}}

    \subfigure[]{\includegraphics[width=1.6in]{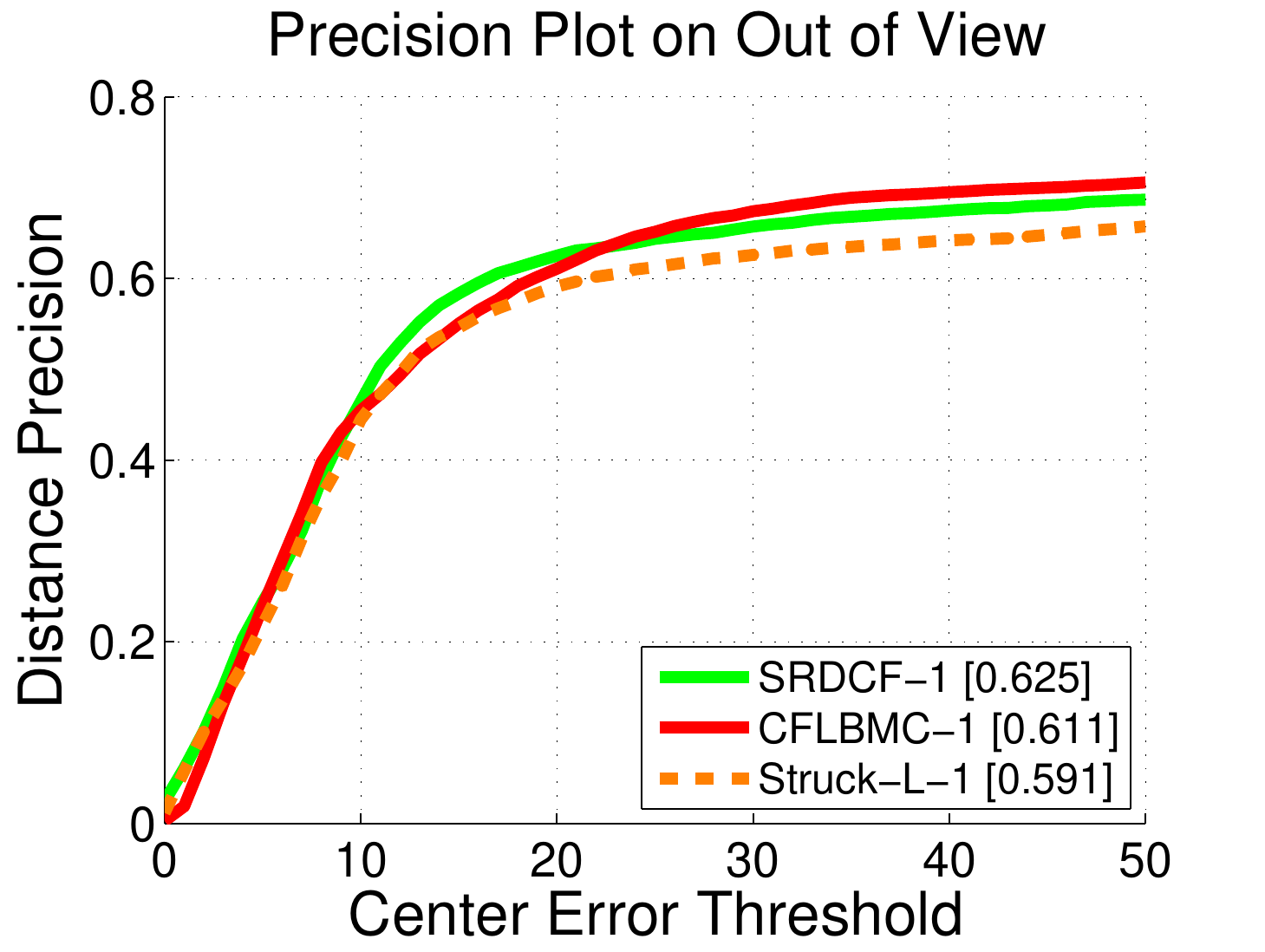}}
  \subfigure[]{\includegraphics[width=1.6in]{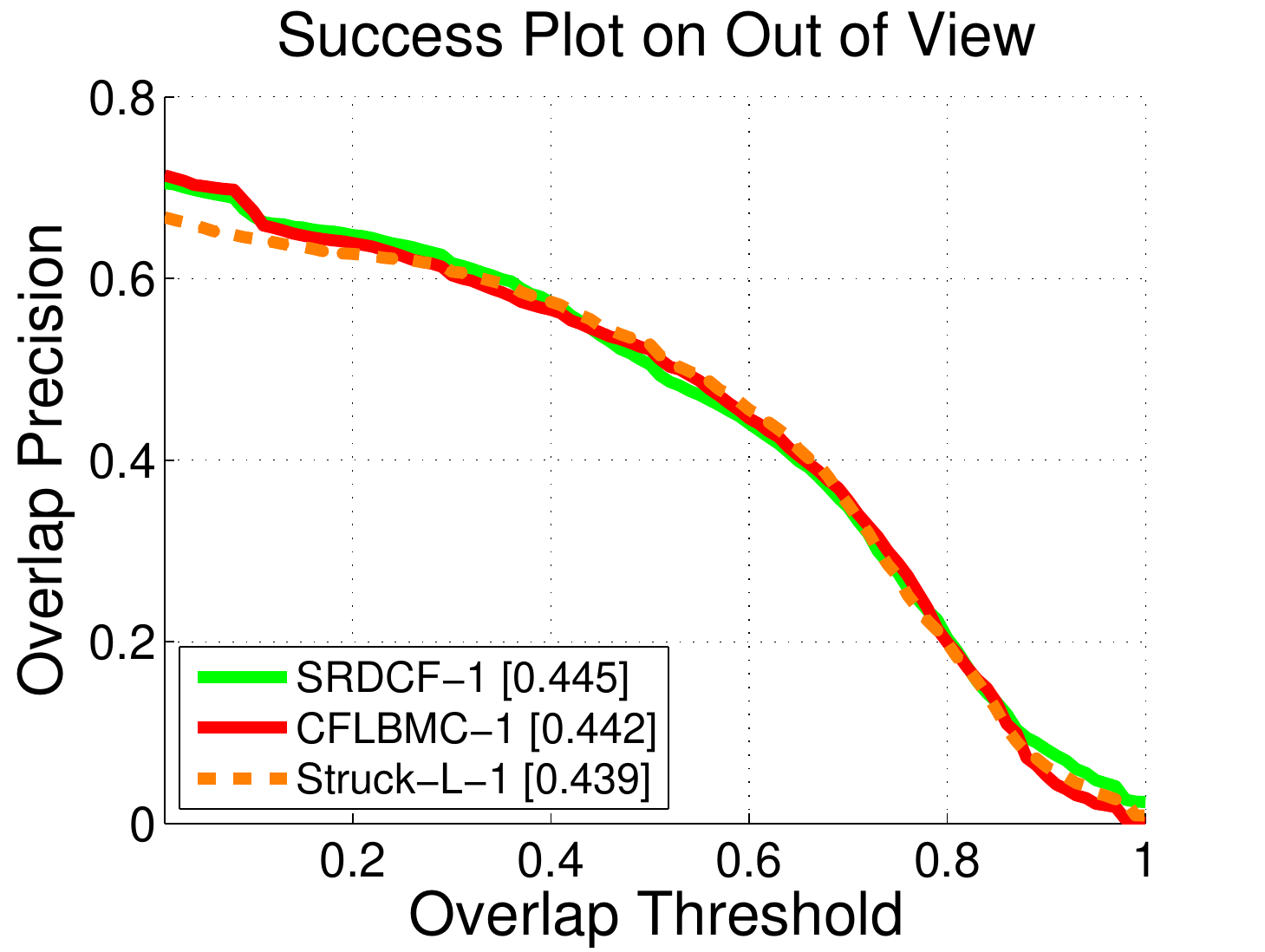}}
   \subfigure[]{\includegraphics[width=1.6in]{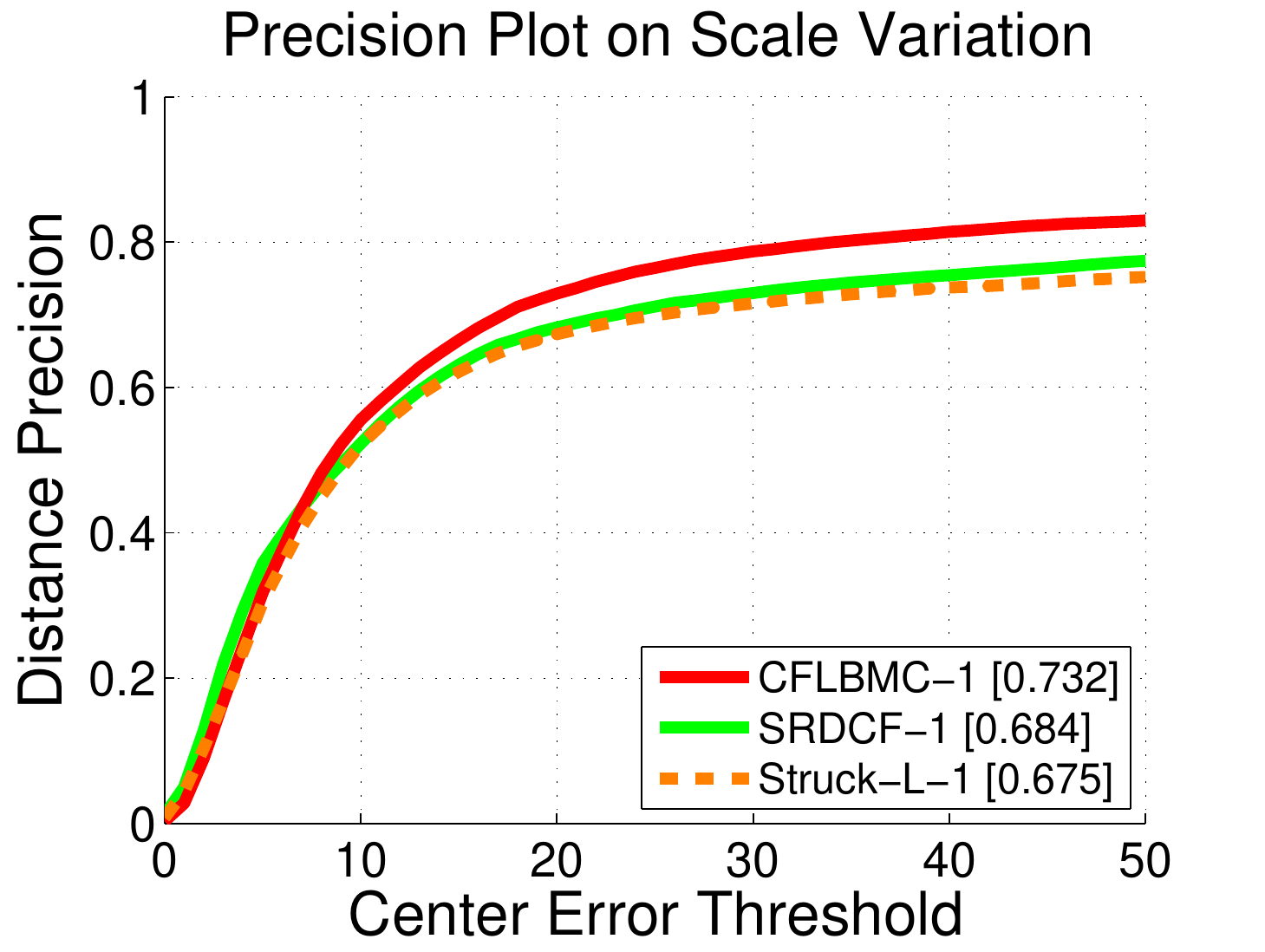}}
 \subfigure[]{\includegraphics[width=1.6in]{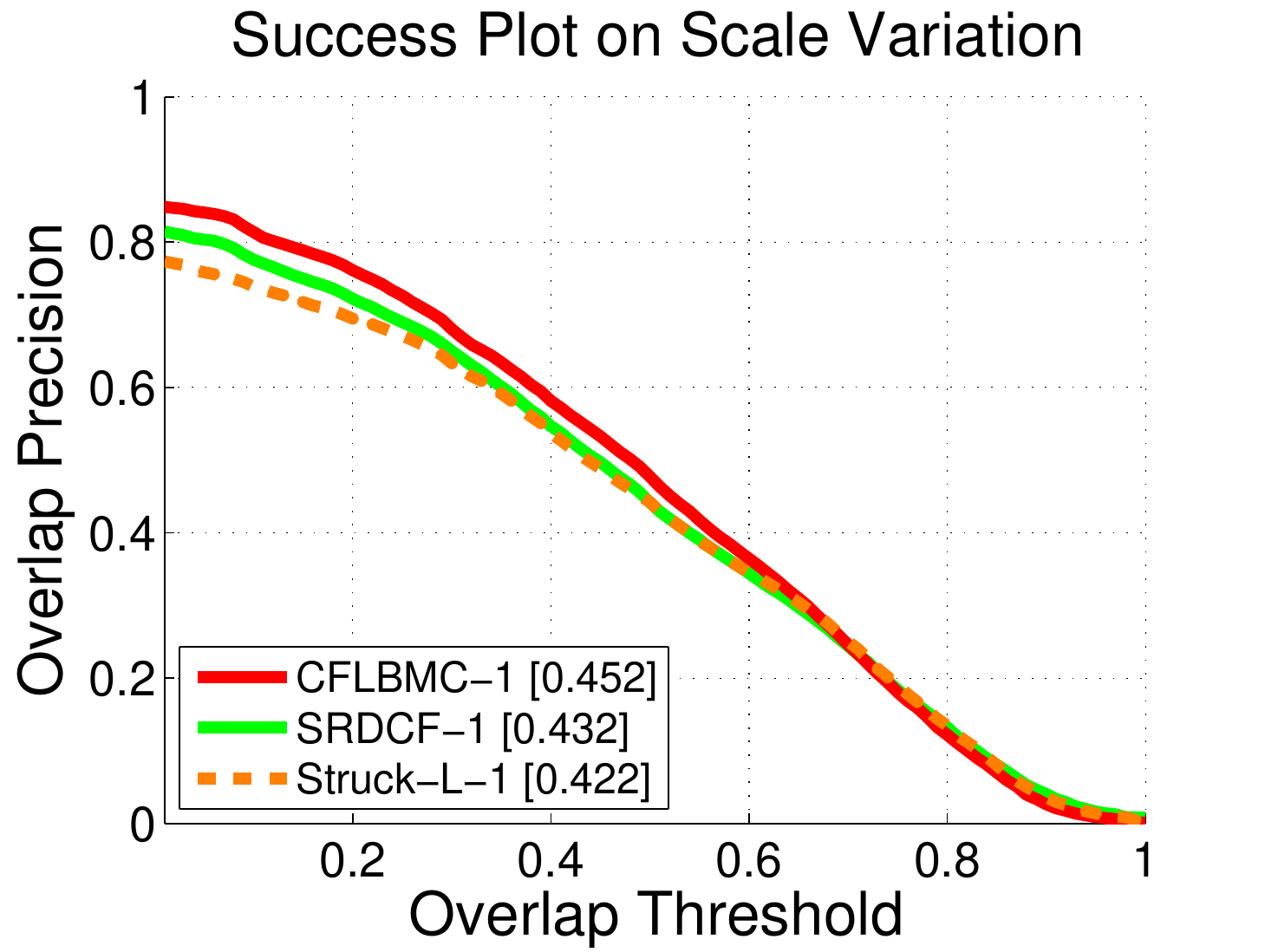}}
  \caption{The average precision plots and success plots of three trackers, SRDCF-1, CFLBMC-1, and Struck-L-1, on the whole benchmark and 11 annotated attributes of OTB100. The mean distance precision scores and AUCs of the trackers are reported in the legend. See text for details.}
  \label{fig:noscale100}
\end{figure*}

Figs.~\ref{fig:noscale50} and~\ref{fig:noscale100} (a) and (b) show the experimental results of SRDCF-1, CFLBMC-1, and Struck-L-1 on OTB50 and OTB100. It is seen that their precision plots and success plots are almost coincident. Fig.~\ref{fig:noscale50} shows that SRDCF-1, which performs best at $t_c=20$ in distance precision, is only $0.4\%$ higher than the worst Struck-L-1. And Struck-L-1, which performs best in AUC, is only $0.8\%$ higher than the worst CFLBMC-1. In Figs.~\ref{fig:noscale100} (a) and (b), it is seen that the best CFLBMC-1 is only $2\%$ better than the worst Struck-L-1 in distance precision, while the best SRDCF-1 only performs $0.5\%$ better than the worst Struck-L-1 in AUC. It can be drawn from the experiments that SRDCF-1, CFLBMC-1, and Struck-L-1 perform almost the same.

For further verifying the relation of SRDCF-1, CFLBMC-1, and Struck-L-1 experimentally, we ran them on the sequences of 11 attributes, respectively. And the results are reported in Figs.~\ref{fig:noscale100} (c) to (x).

It can be seen in Figs.~\ref{fig:noscale100} (c) to (x) that SRDCF-1, CFLBMC-1, and Struck-L-1 illustrate the approximate relation on the sequences of all attributes. If the overlap ratio related criteria are considered more reasonable than center error related ones, the inconsistences of the trackers are reflected mainly in the four attributes, Plane Rotation ((i) and (j)), Low Resolution ((m) and (n)), Motion Blur ((o) and (p)), and Scale Variation ((w) and (x)).

\subsubsection{Inconsistencies and their Reasons}
\label{sec:inconsistencyreasons}
\begin{figure}[t]
  \centering
  \includegraphics[width=3in]{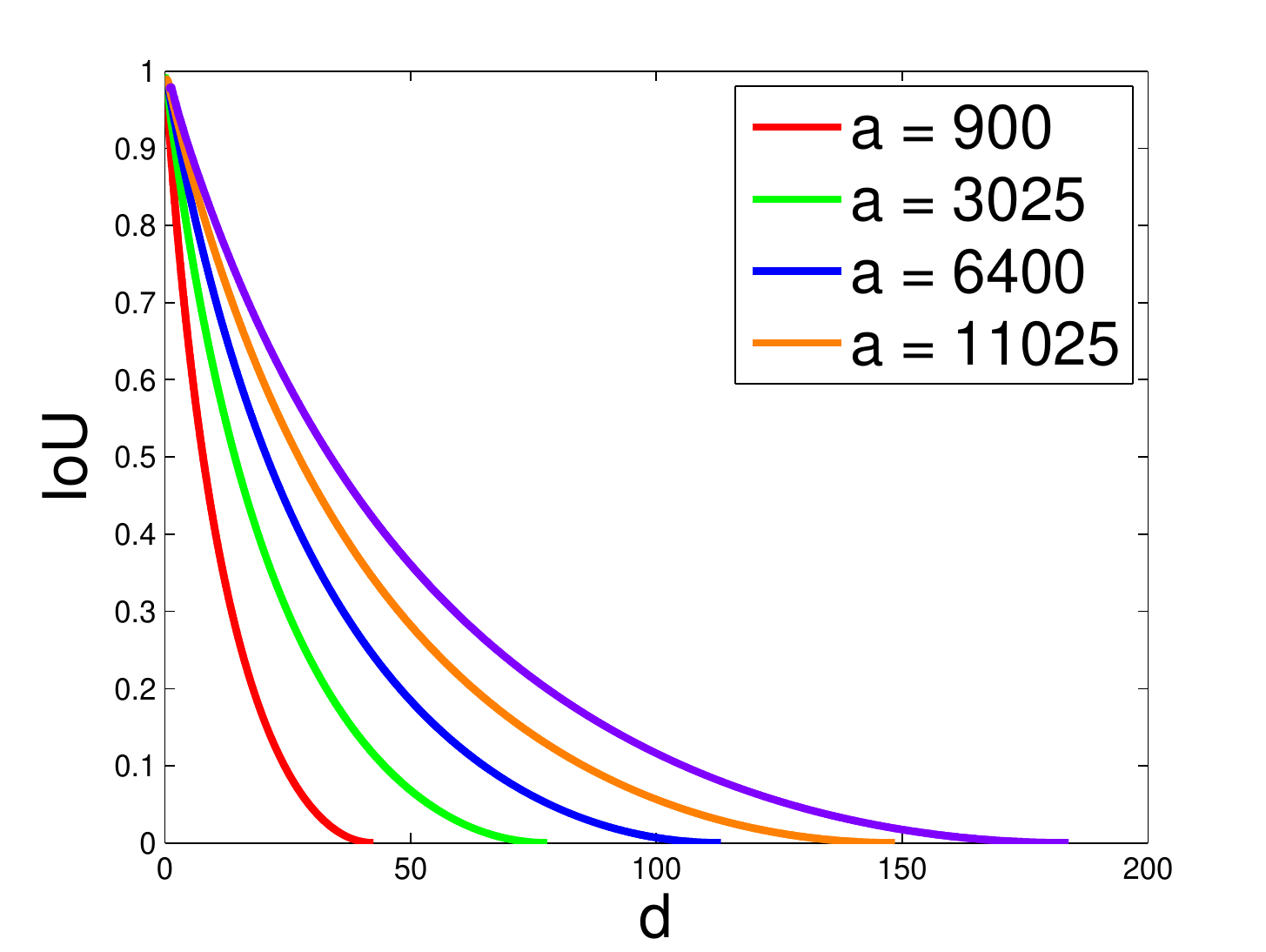}
  \caption{
  The mathematical characteristic of overlap ratio. $d$ is the distance between the centers of ground truth and located object. $a$ is the area of object bounding box.}
  \label{fig:relation}
\end{figure}

After observing their performances on each of the 100 sequences, we found that the following sequences result in the differences of the trackers' performances. These sequences are Bird1, Bird2, BlurOwl, Bolt, CarDark, ClifBar, Coke, Couple, Crowds, Diving, Football, Freeman1, Freeman4, Girl, Human2, Human4-2, Jogging-2, Jumping, KiteSurf, Lemming, Liquor, Panda, RedTeam, Rubik, Shaking, Subway, and Trans. After qualitatively analyzing the detailed differences of SRDCF-1, CFLBMC-1, and Struck-L-1 carefully, we draw that the following six details lead to the observed inconsistencies.
\begin{itemize}
  \item[1.] The spatial regularization leads to differences. SRDCF-1 adopts quadratic function, while CFLBMC-1 uses a masking function, and Struck-L-1 implicitly uses a masking function but with sparser sampling than CFLBMC-1 in training. Therefore, SRDCF-1 can learn some local background, while CFLBMC-1 and Sturck-L-1 can not. The following sequences show the inconsistency caused by the different spatial regularizations: Bird1, Bird2, BlurOwl, Bolt, ClifBar, Coke, Football, Freeman1, Freeman4, Girl, Lemming, Liquor, Panda, RedTeam, Rubik, and Shaking.
  \item[2.] The sizes of searching regions lead to differences. SRDCF-1 and CFLBMC-1 use the square region of 4 times area of the object bounding box, while Struck-L-1 only searches the object within a region of 2.5 times width and height of the object bounding box. The following sequences show the inconsistency caused by the different sizes of searching regions: BlurOwl, Couple, Diving, Human4-2, and Jogging-2.
  \item[3.] The update methods lead to differences. SRDCF-1 and CFLBMC-1 utilize the same update method of exponential weighting, while Struck-L-1 updates its appearance model through adding and discarding positive and negative support vectors. Therefore, SRDCF-1 and CFLBMC-1 are rarely affected much by the historical appearances of the target far from the current frame, while Struck-L-1 is quite the reverse. The following sequences show the inconsistencies caused by different update methods: ClifBar, Coke, Couple, Freeman4, Girl, Human2, KiteSurf, Lemming, and Liquor.
  \item[4.] The localization details lead to differences. Since the cell size of HOG is $4 \times 4$, the localization accuracy of CFLBMC-1 is 4 pixels in both vertical and horizontal axes. Whereas SRDCF-1 invokes a sub-pixel fine-locating method to locate the object in 1 pixel accuracy. And Struck-L-1 makes use of dense sampling in locating the object, achieving 1 pixel accuracy. The following sequence shows the inconsistency caused by the different localization details: CarDark, and KiteSurf.
  \item[5.] The combination of spatial regularization and banding windows leads to differences. Both SRDCF-1 and CFLBMC-1 adopt Hann window, while Struck-L-1 uses no banding window\footnote{Hann window, a Gaussian-like function, is commonly used as banding window to eliminate the discontinuity between opposite edges of non-periodic images, when transferring them into frequent domain with FFT.}. When the duration of occlusion is long enough, the appearance models of SRDCF-1 and CFLBMC-1 will learn some background noise, weakening the discriminativity of their models. Moreover, Hann window will impair the HOG near the search region boundaries while SRDCF-1 and CFLBMC-1 are locating the object in any frame. Consequently, CFLBMC-1 may fail if the object is occluded for a long enough duration and re-appears near the boundaries of search region, because of the weakened discriminativity of its model and the disturbance of Hann window on HOG of search region. Nevertheless, SRDCF-1 and Struck-L-1 will succeed in these cases because of the assist of local background for SRDCF-1 and no disturbance of Hann window on HOG of search region for Struck-L-1. The following sequences involve such cases: Bolt, Coke, Joggling-2, Jumping, and Subway.
  \item[6.] The mathematical characteristic of the overlap ratio curve leads to obvious differences. The relation of distance of the centers of two bounding boxes and their overlap ratio is similar to an exponential function, as shown in Fig.~\ref{fig:relation}. A small distance change of the centers of ground truth and located object will result in a conspicuous change of overlap ratio, if the two bounding boxes are close enough. And the smaller the areas of bounding boxes, the more conspicuous this case is. The following sequence shows the inconsistency caused by such an exponential-like relation: Crowds.
\end{itemize}

Note that the first four observations involve in four different details on trackers, and the last two ones come from the combination of different details. Except for the above, we do not find any other factors to generate obvious differences in localization.

In the following, we will illustrate the aforementioned six reasons with representative sequences.

\begin{figure}[t]
  \centering
  \includegraphics[width=3.4in]{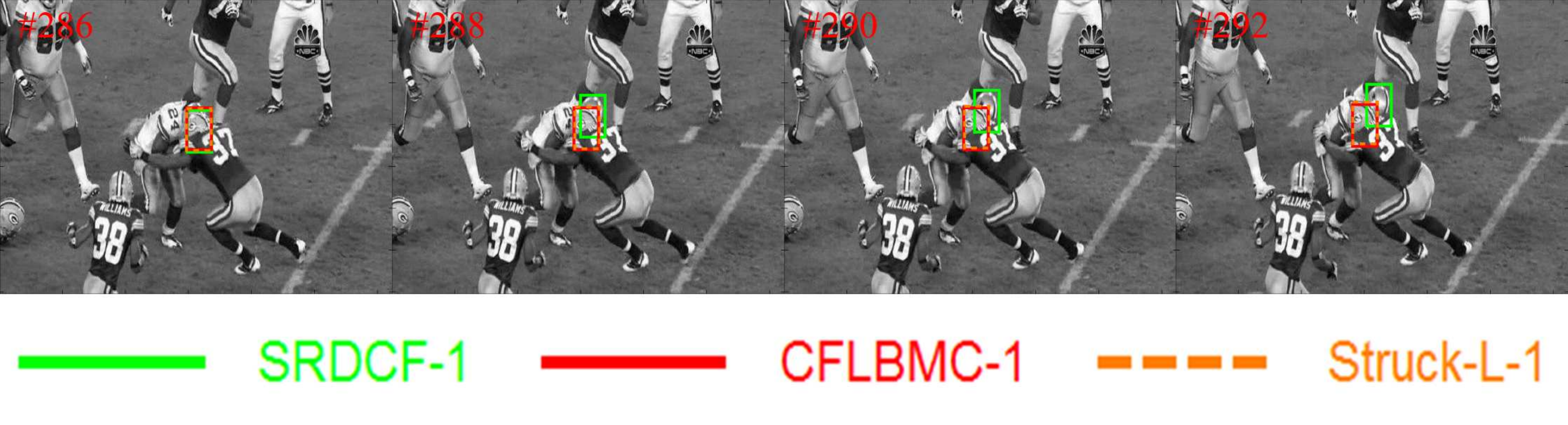}
  \caption{In football sequence, SRDCF-1 can catch the target object successfully because its appearance model is trained with the located objects and their local backgrounds. CFLBMC-1 and Struck-L-1 fail because they train their appearance models without local background. See text for details.}
  \label{fig:football}
\end{figure}

Fig.~\ref{fig:football} shows that CFLBMC-1 and Struck-L-1 may fail when the target is occluded by similar distracters, since they do not explore local backgrounds in training. Whereas, SRDCF-1 precisely catches the target object because its appearance model is trained with the object and its local background and there exist conspicuous differences between the local backgrounds of target object and distracters.

\begin{figure}[t]
  \centering
  \includegraphics[width=3.4in]{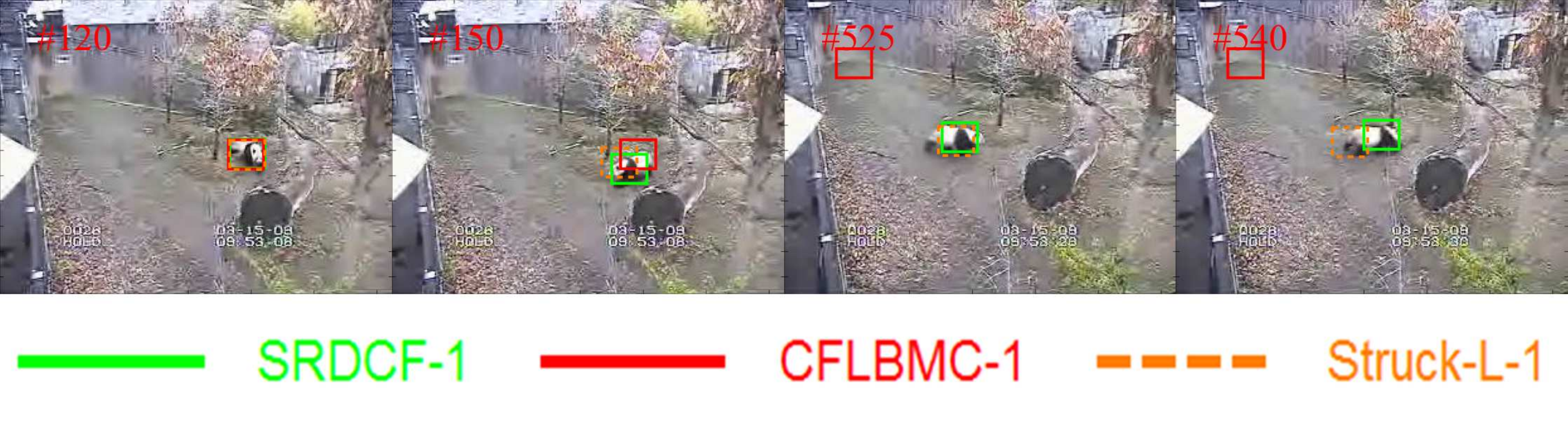}
  \caption{In panda sequence, SRDCF-1 catches the object successfully in the help of local background, while CFLBMC-1 and Struck-L-1 lose the panda, because they do not explore local background to train their appearance models.}
  \label{fig:Panda}
\end{figure}

Fig.~\ref{fig:Panda} shows another sequence where the relatively stable appearance of local background helps SRDCF-1 catch the panda robustly, although the panda's appearance changes greatly due to its rotation. Contrarily, CFLBMC-1 and Struck-L-1 explore no background, therefore lose the target while it rotates too much.

\begin{figure}[t]
  \centering
  \subfigure[]{\includegraphics[width=0.8in]{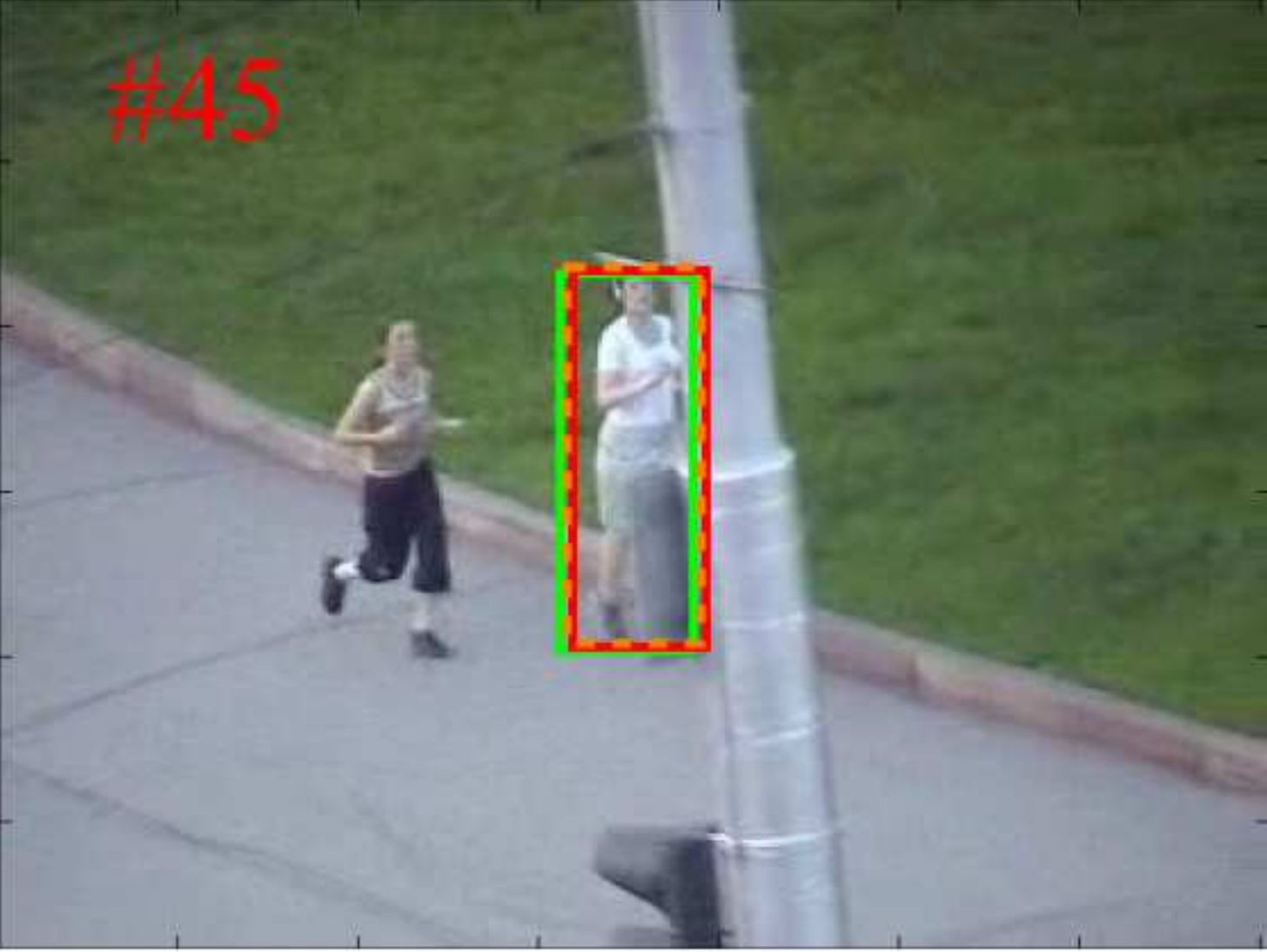}}
  \subfigure[]{\includegraphics[width=0.8in]{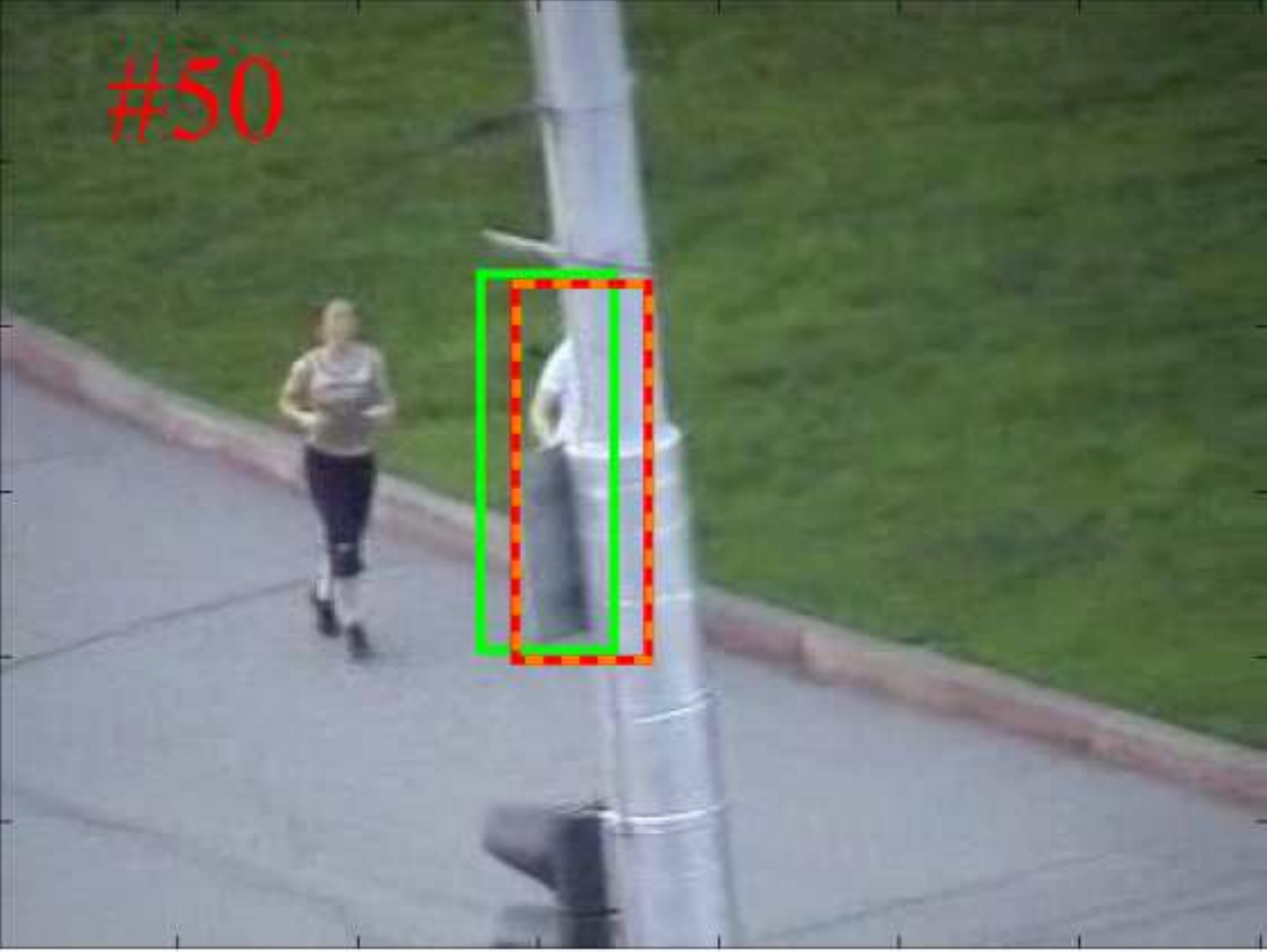}}
  \subfigure[]{\includegraphics[width=0.8in]{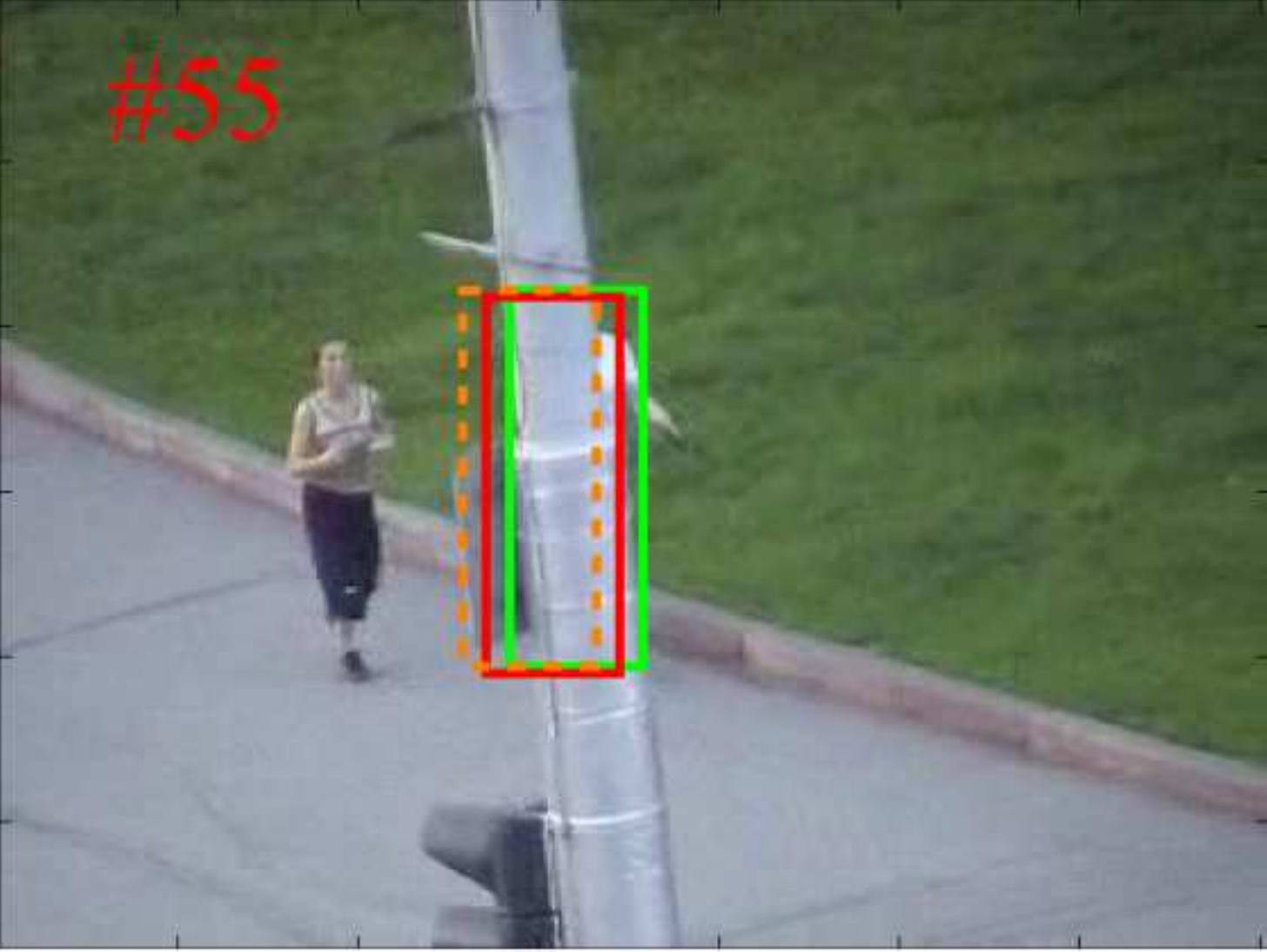}}
  \subfigure[]{\includegraphics[width=0.8in]{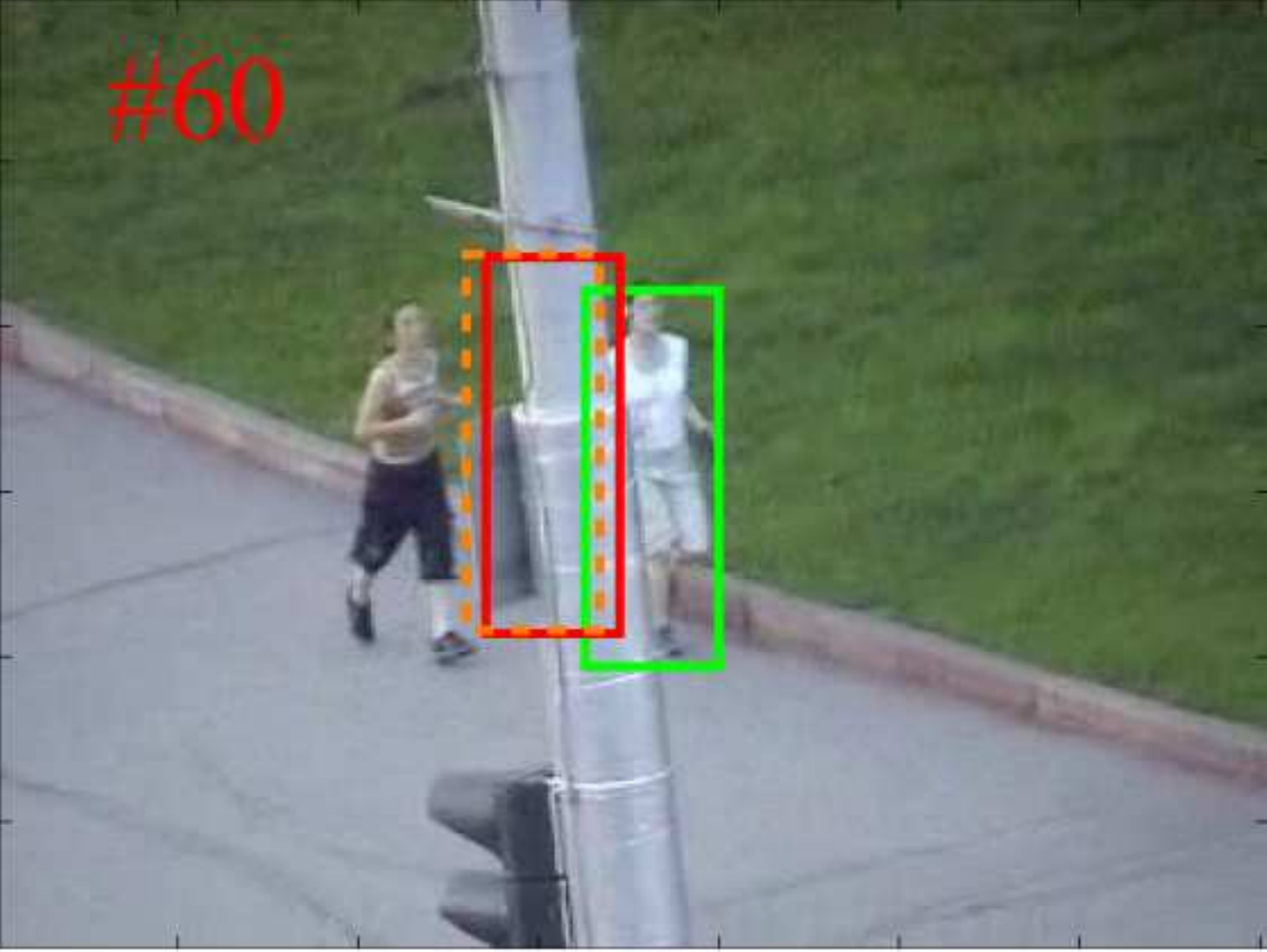}}\\
  \subfigure[]{\includegraphics[width=0.8in]{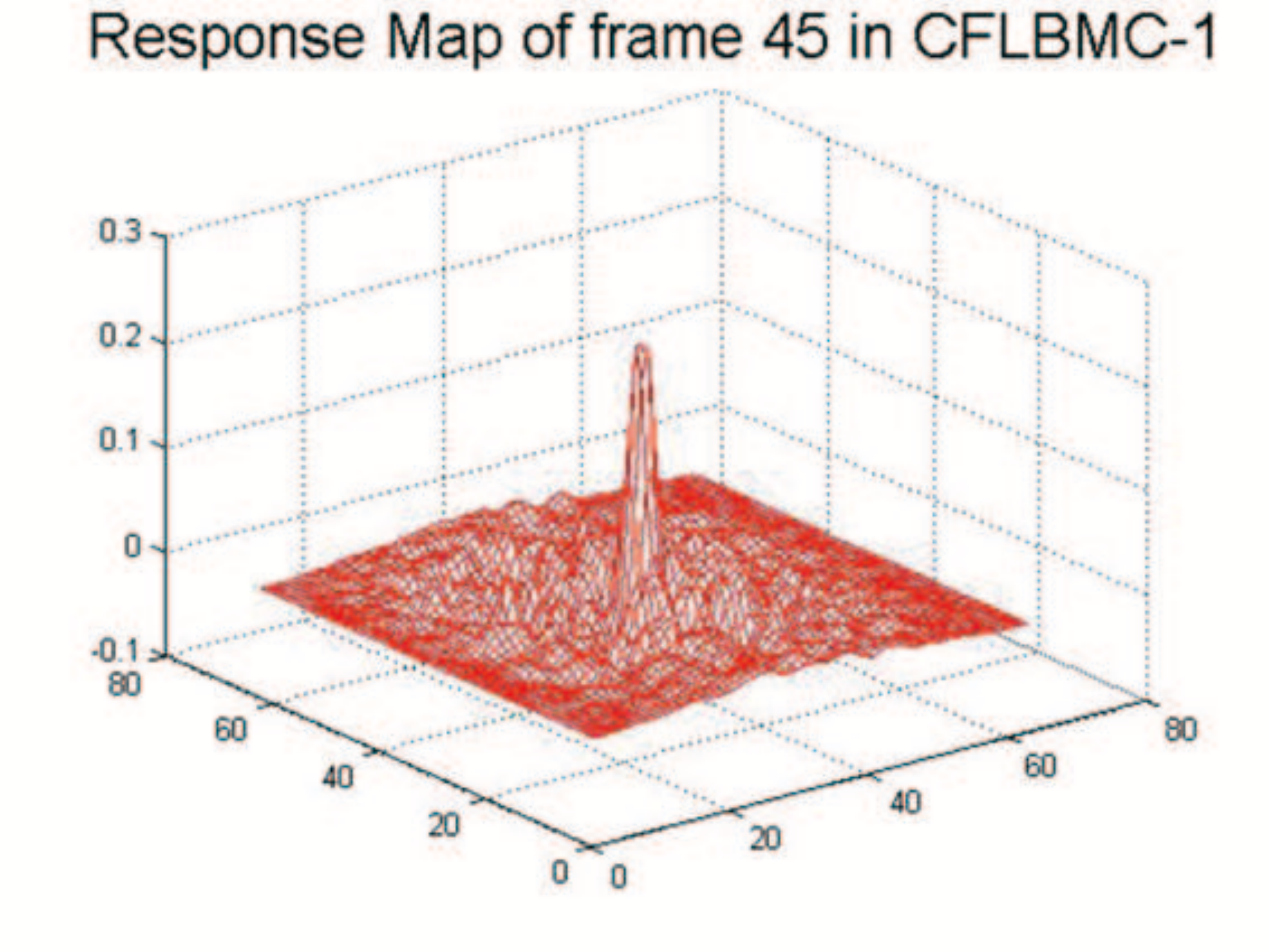}}
  \subfigure[]{\includegraphics[width=0.8in]{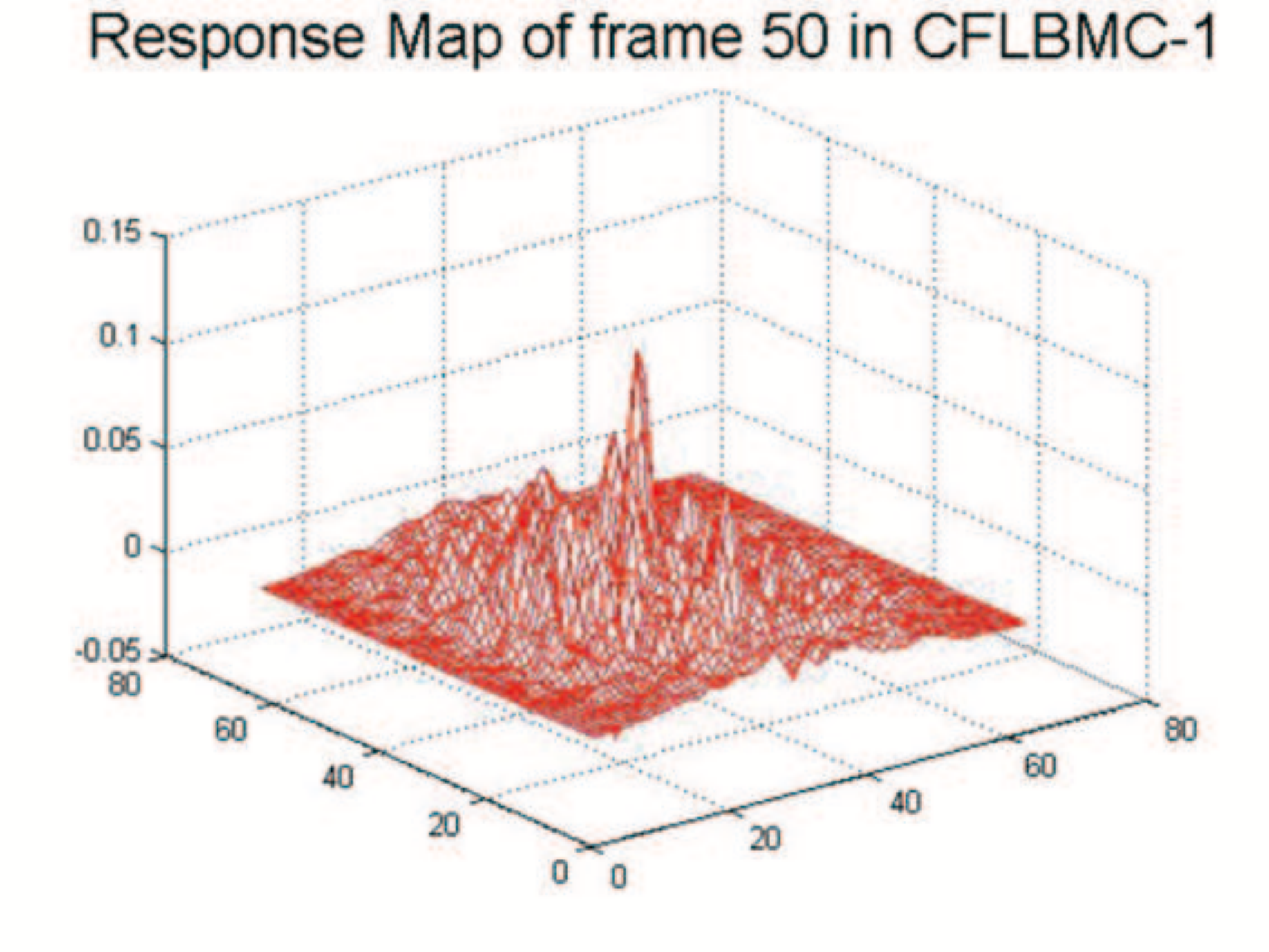}}
  \subfigure[]{\includegraphics[width=0.8in]{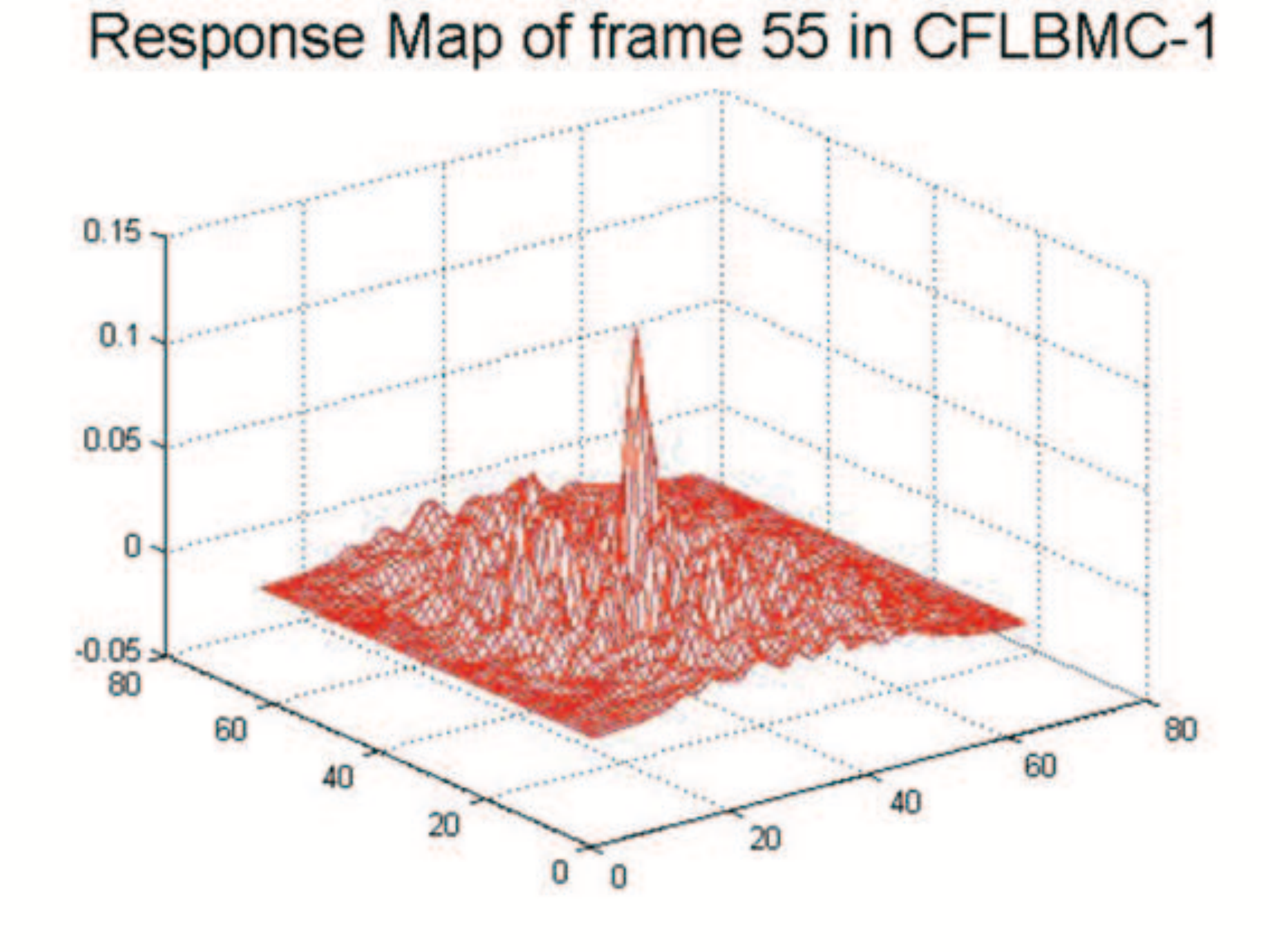}}
  \subfigure[]{\includegraphics[width=0.8in]{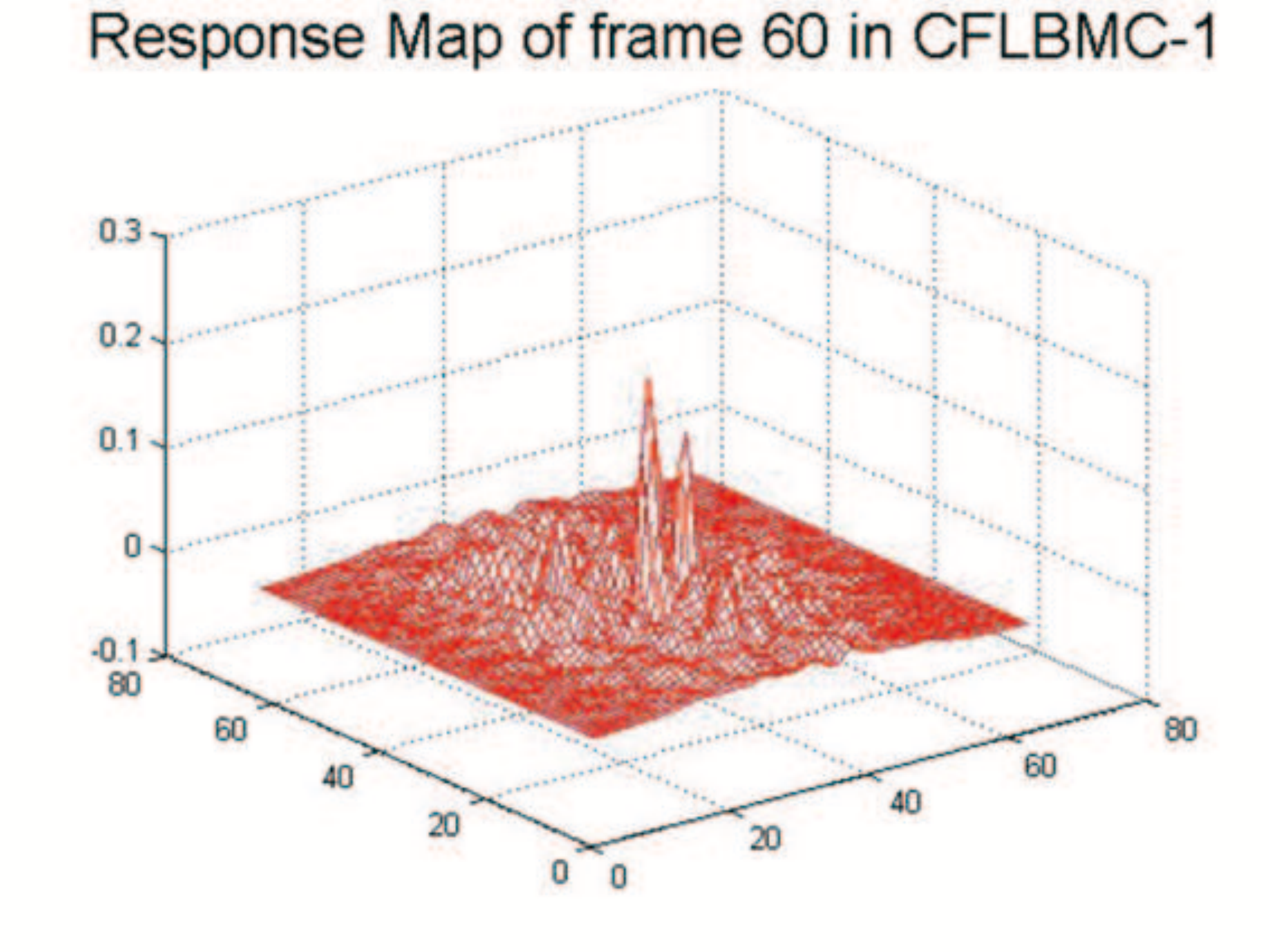}}\\
  \subfigure[]{\includegraphics[width=0.8in]{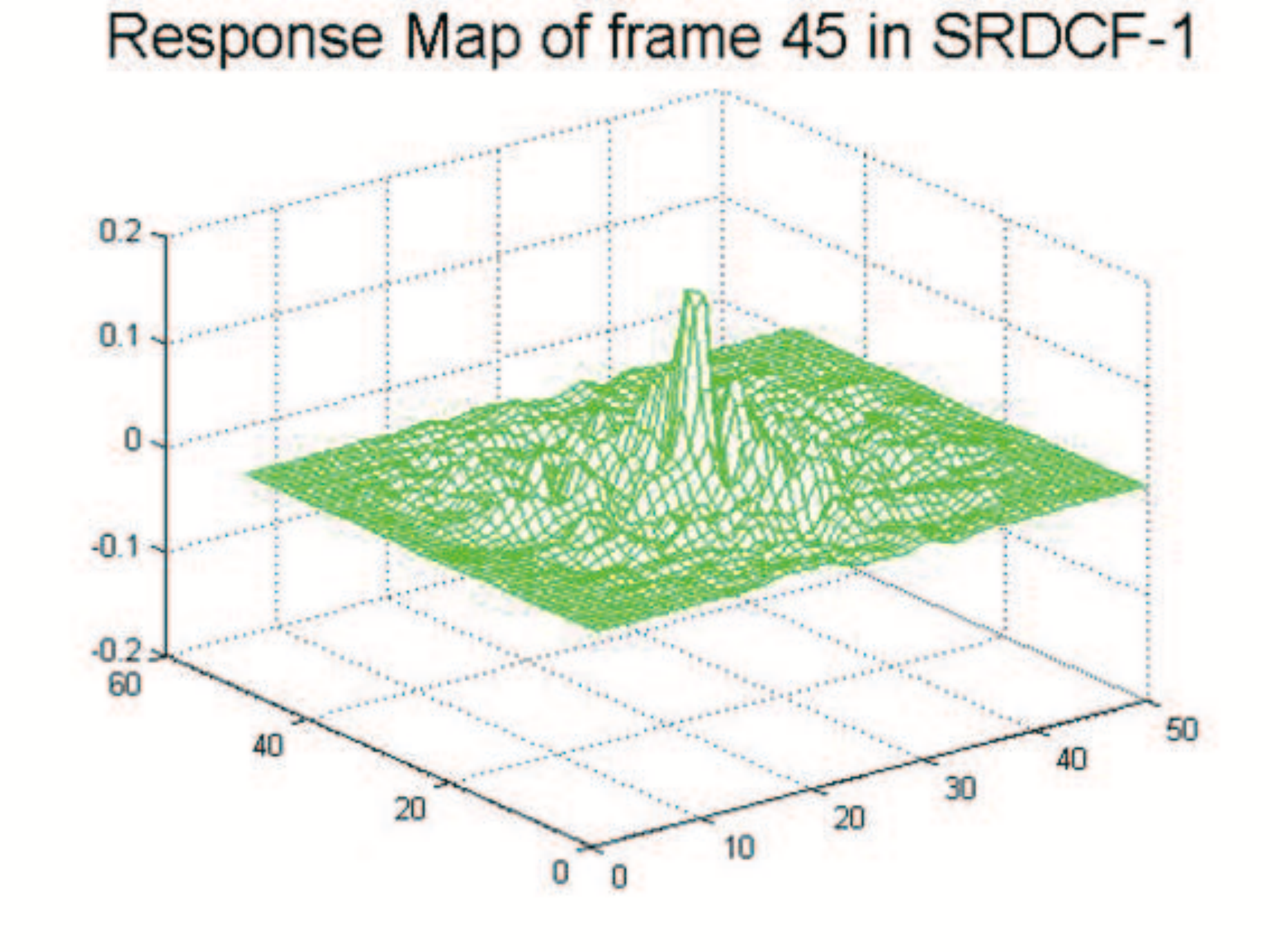}}
  \subfigure[]{\includegraphics[width=0.8in]{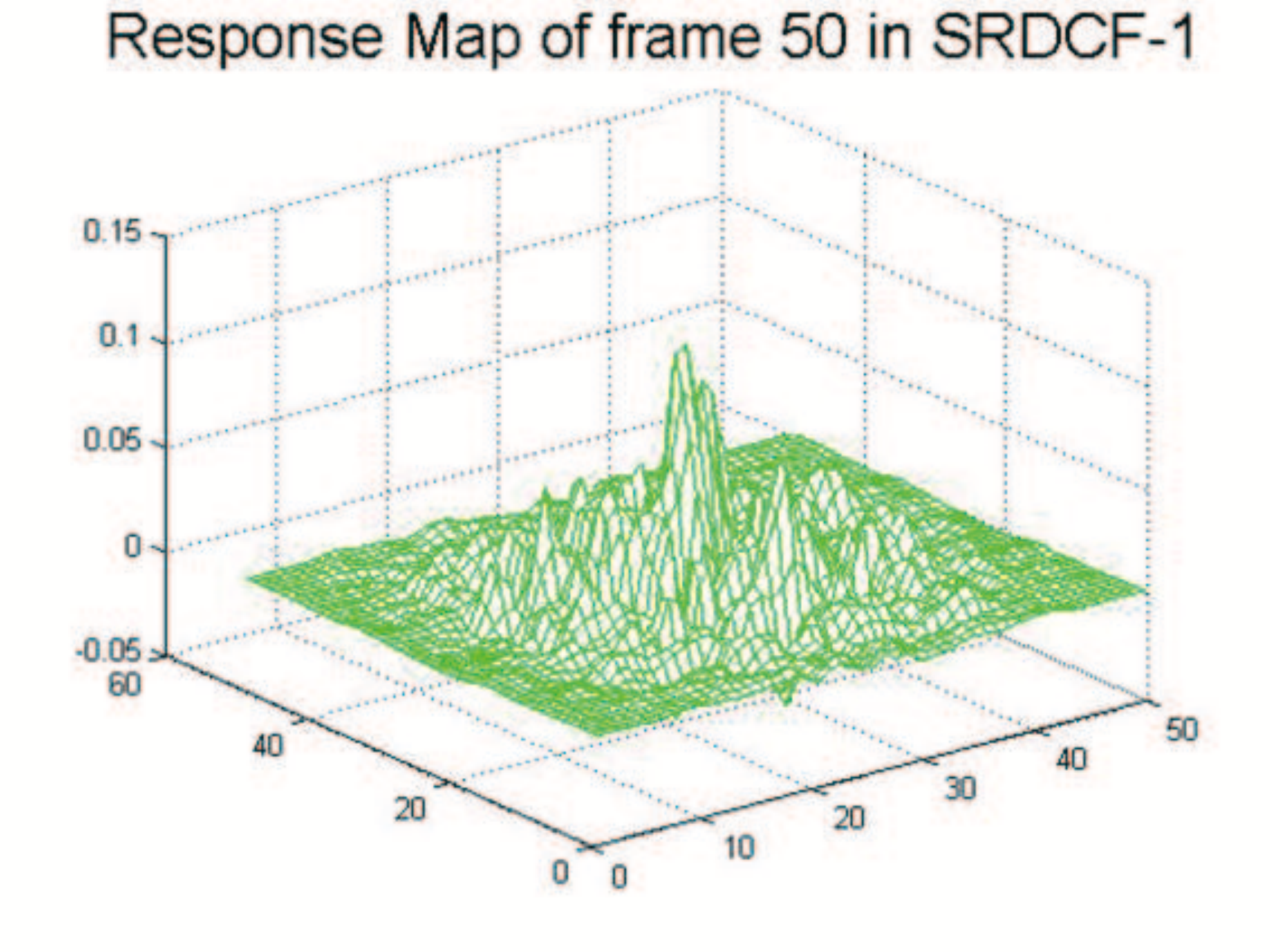}}
  \subfigure[]{\includegraphics[width=0.8in]{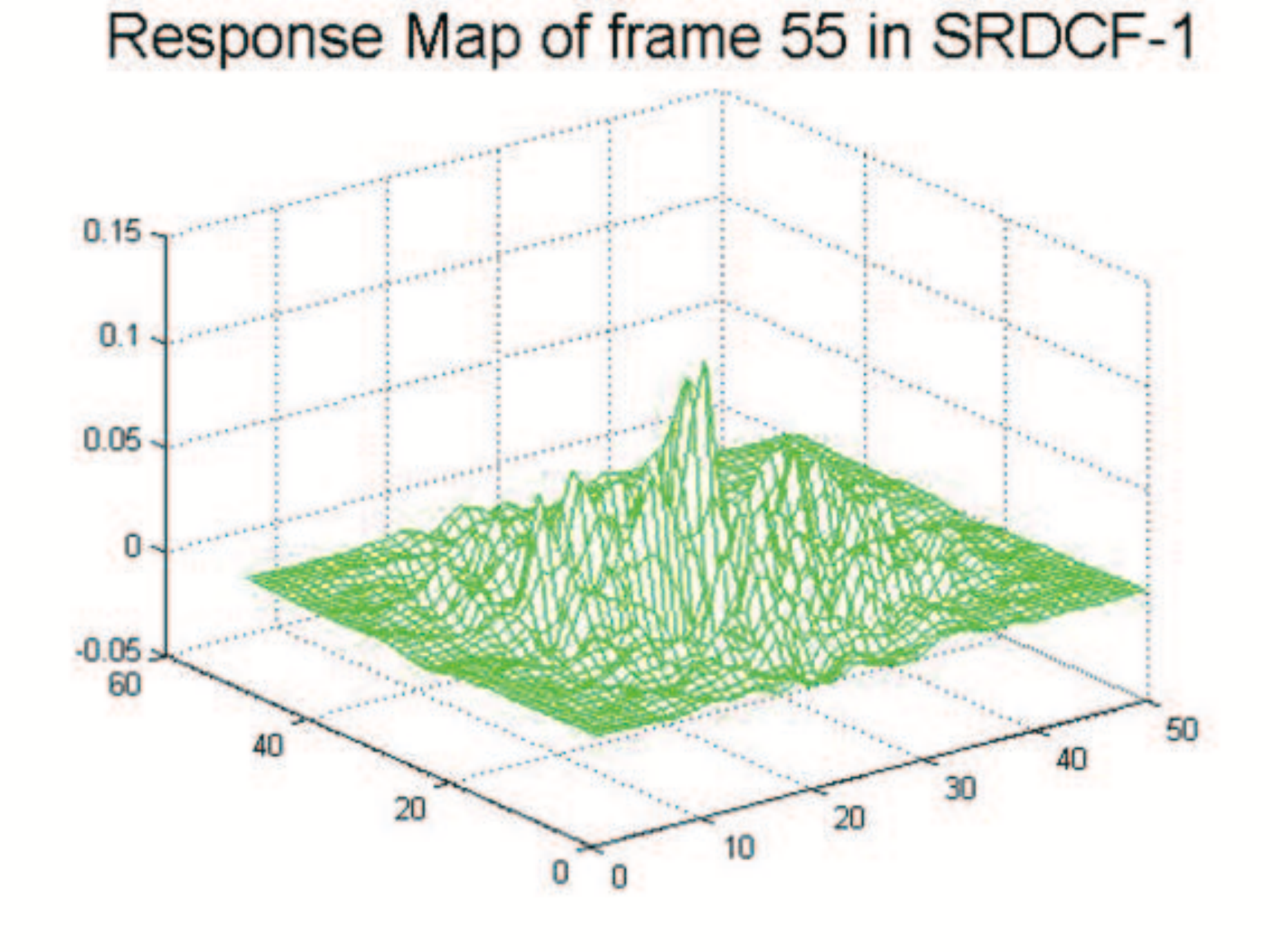}}
  \subfigure[]{\includegraphics[width=0.8in]{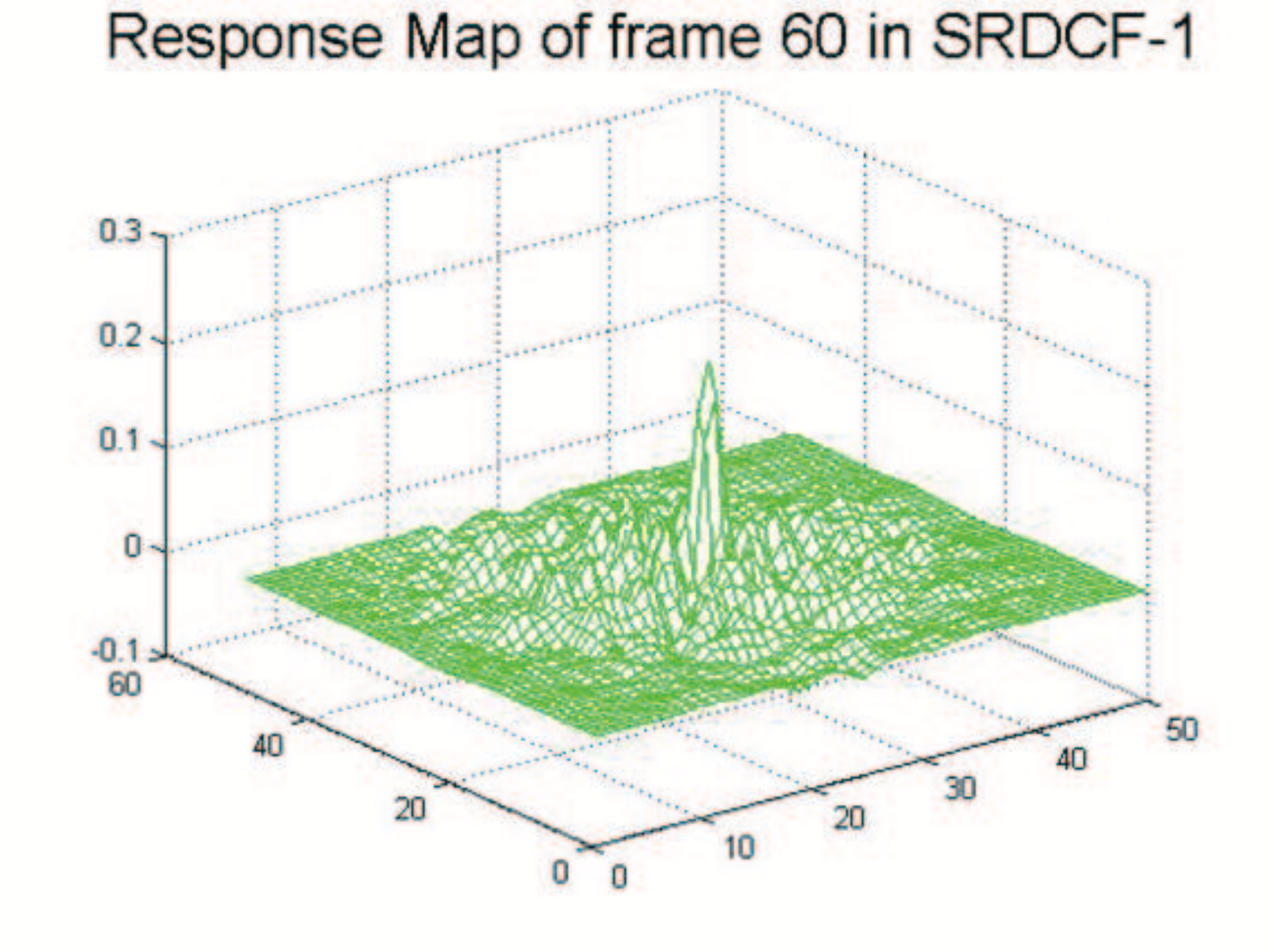}}\\
  \includegraphics[width=3.4in]{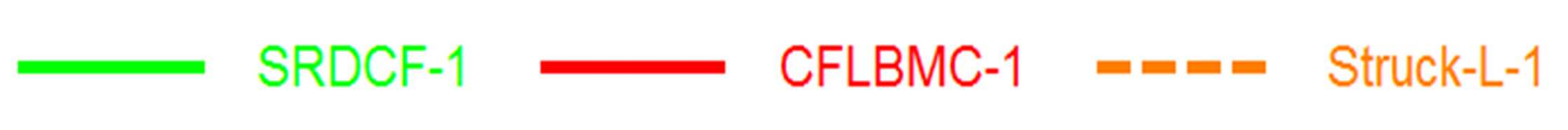}
  \caption{Jogging-2 sequence illustrates the coefficient effects of exploring background in training, update rate, and the sizes of search regions on location performances. See text for details.}
  \label{fig:Jogging-2}
\end{figure}

Fig.~\ref{fig:Jogging-2} shows an example to illustrate the coefficient effects of exploring background in training and the sizes of search regions on location performances. Observing the response maps of SRDCF-1 before and after occlusion in Figs.~\ref{fig:Jogging-2} (i) and (l), it is found that there is a dominant peak per map. This is because the object background varies little in these two frames and the update rate is really slow. Therefore, SRDCF-1 catches the target successfully after the occlusion is over. In contrast, Figs.~\ref{fig:Jogging-2} (e) and (h) tell us that there is a dominant peak for CFLBMC-1 before occlusion, whereas there exist two dominant peaks after occlusion: the highest one corresponds to part of post pole, and another to the target object. The reasons that the highest peak does not correspond to the target object are that the discriminativity of HOG is not strong enough to distinguish the object from its background in these frames and Hann window impairs the discriminativity further. Therefore, CFLBMC-1 fails in this sequence. Because the searching region of Struck-L-1 is not large enough, the target object gets out of the search region when it re-appears. Consequently, Struck-L-1 also fails in this sequence.

\begin{figure}[t]
  \centering
  \includegraphics[width=3.4in]{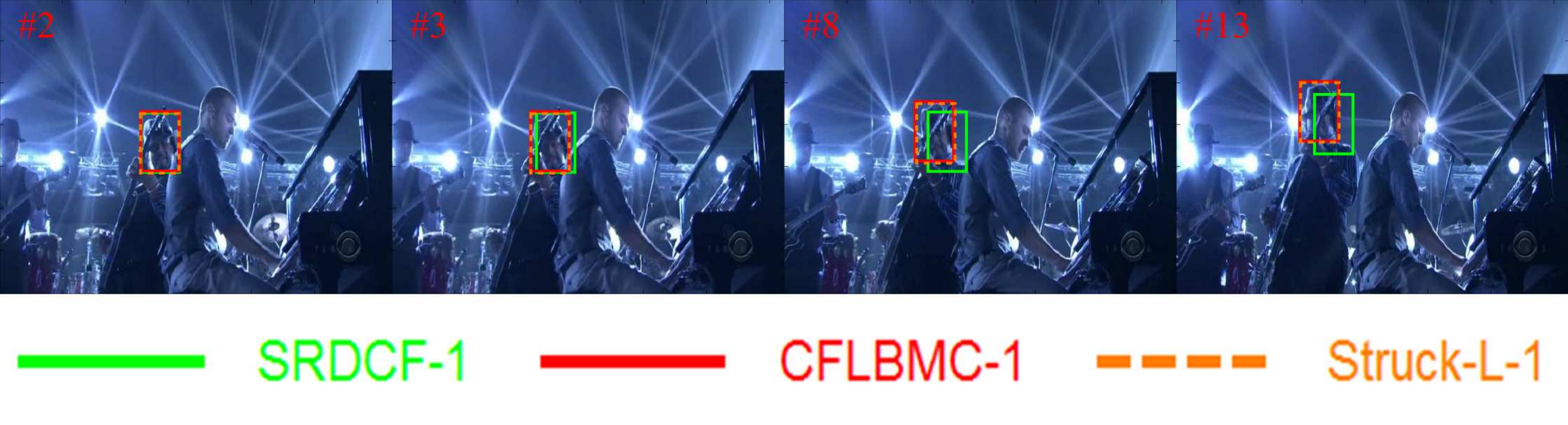}
  \caption{Shaking sequence shows the negative effect on SRDCF-1 when exploring the local background to train the appearance models. See text for details.}
  \label{fig:Shaking}
\end{figure}

Nevertheless, exploring local backgrounds to train SRDCF-1 leads to failure sometimes, as shown in Fig.~\ref{fig:Shaking}. Indeed, compared with the edges of the target object, those of its local background are more distinct in initial frames. Therefore, when involving local background in training, to distinguish different samples SRDCF-1 will depend more on the local background than on the object itself. Conspicuous changes of the local background will deviate SRDCF-1 from the object. Whereas, CFLBMC-1 and Struck-L-1 only train their appearance models with the object itself. Therefore they are able to catch the object robustly in such case.

\begin{figure}[t]
  \centering
  \includegraphics[width=3.4in]{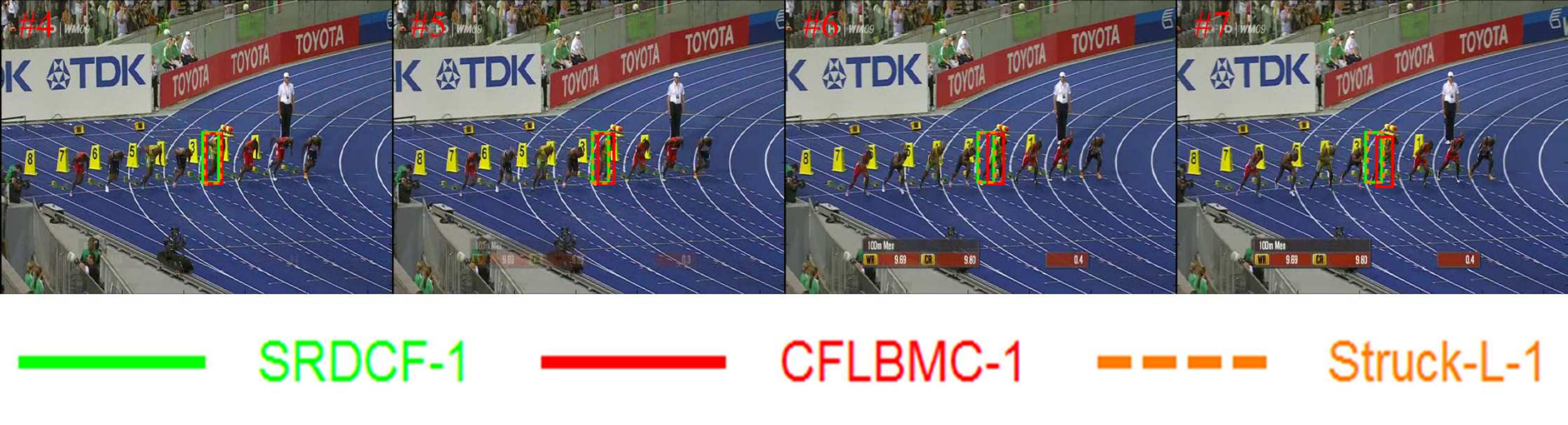}
  \caption{Bolt sequence illustrates the negative effects of distinct background and inadequate samples on SRDCF-1 and Struck-L-1, respectively. See text for details.}
  \label{fig:Bolt}
\end{figure}

Fig.~\ref{fig:Bolt} shows a similar example where there exists a distinct curve in Bolt's background. It is difficult for SRDCF-1 to catch the object because the background curve dominates the response map. Through training its model without exploring background, CFLBMC-1 successfully locates Bolt. Because there are no enough training and updating samples for Struck-L-1 in initial frames due to sparse sampling, the rapid change of Bolt's pose results in its failure.

\begin{figure}[t]
  \centering
  \includegraphics[width=3.4in]{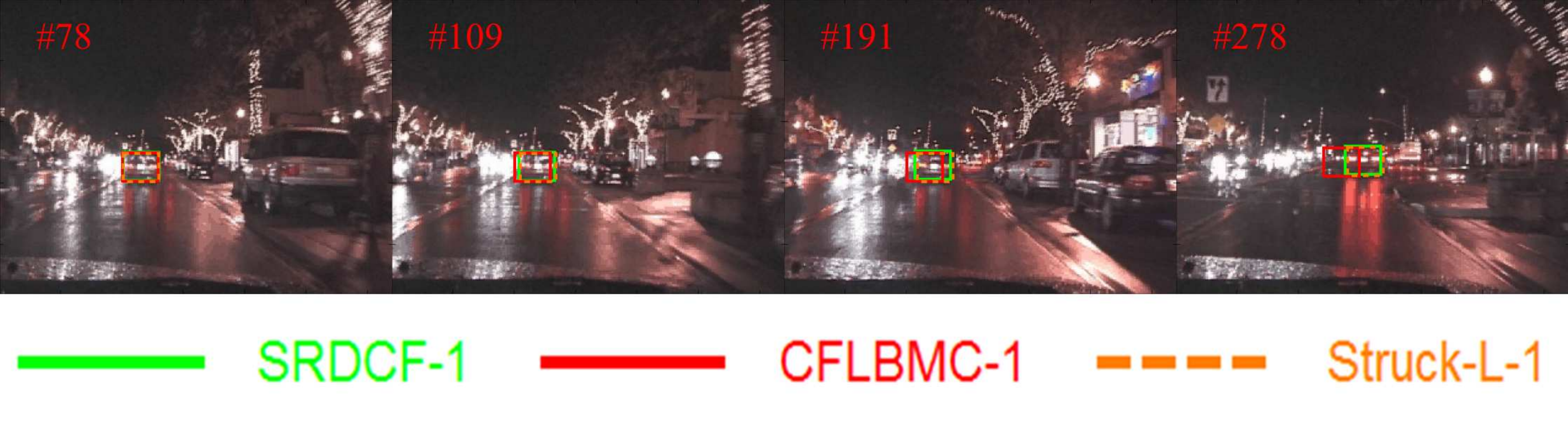}
  \caption{In CarDark sequence, too low location resolution of CFLBMC-1 results in its gradual failure. See text for details.}
    \label{fig:CarDark}
\end{figure}

As aforementioned, the minimal resolution of the employed HOG is $4\times4$ pixels in our experiments. In most of sequences, such a resolution does not affect the performance of the three trackers much. Nevertheless, in CarDark sequence, the performance of CFLBMC-1 is negatively affected by the low resolution, as shown in Fig.~\ref{fig:CarDark}. This is because the target car is small and blur, and its background involves many distinct edges. A small deviation from the correct location means much to the small target car. Therefore, the bounding box located by CFLBMC-1 slowly and gradually shifts away from the target car and sticks to the background. On the other hand, SRDCF-1 developed a sub-pixel method to refine its location accuracy up to a pixel in every frame, and Struck-L-1 densely samples the candidate object bounding boxes. Therefore, they successfully catch the target car in the sequence.

\begin{figure}[t]
  \centering
  \includegraphics[width=3.4in]{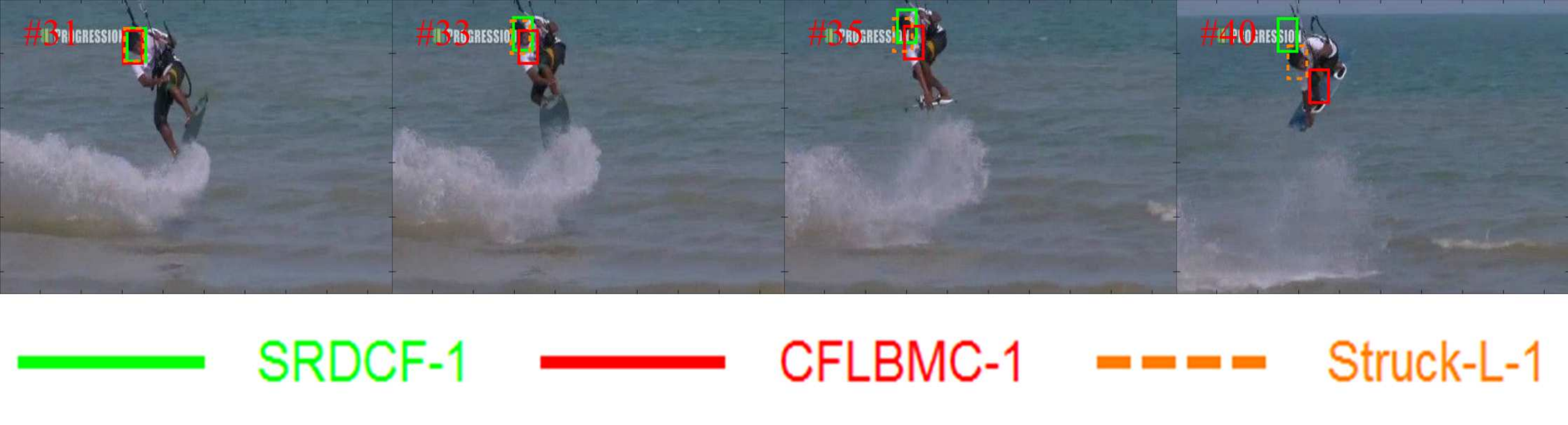}
  \caption{KiteSurf sequence shows the positive effect of dense sampling on Struck-L-1. CFLBMC-1 fails because of its sparse sampling. SRDCF-1 fails because of the negative effect of the distinct edges of the word on it. See text for details.}
  \label{fig:KiteSurf}
\end{figure}

Fig.~\ref{fig:KiteSurf} shows another example where the low resolution of locations makes CFLBMC-1 fail while the dense candidate samples make Struck-L-1 catch the object. In fact, Struck-L-1 will fail whenever the sampling density of its candidates lowers a half or more. The failure of SRDCF is different in this sequence. Because SRDCF explores the local background in training, it can catch the object although not as stably as Struck-L-1 does, after CFLBMC-1 fails. Nevertheless, strong edges of the left corner word, PROGRESSION, mislead the update of SRDCF's appearance model, resulting in the failure of SRDCF.

\begin{figure}[t]
  \centering
  \subfigure[]{\includegraphics[width=0.8in]{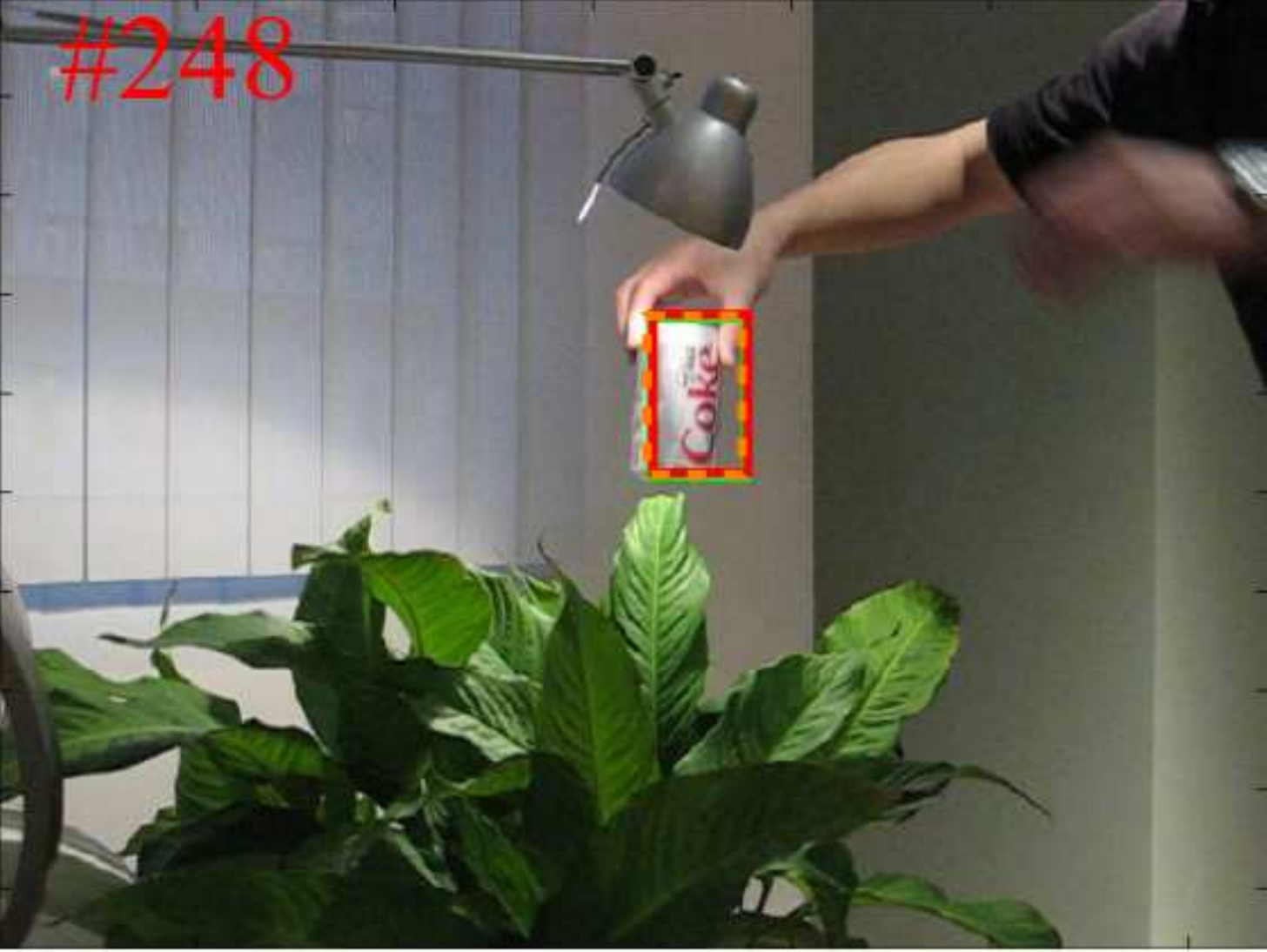}}
  \subfigure[]{\includegraphics[width=0.8in]{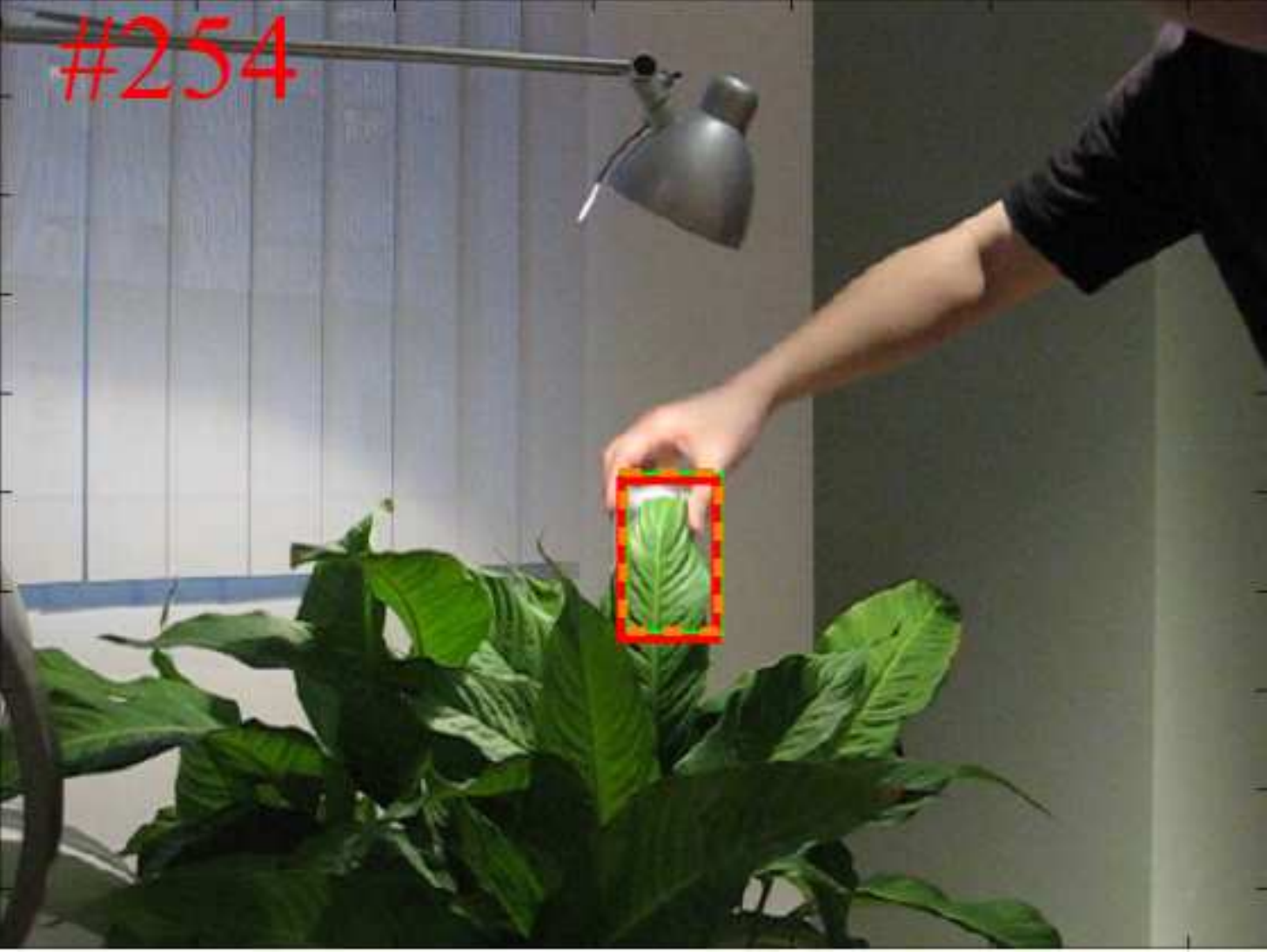}}
  \subfigure[]{\includegraphics[width=0.8in]{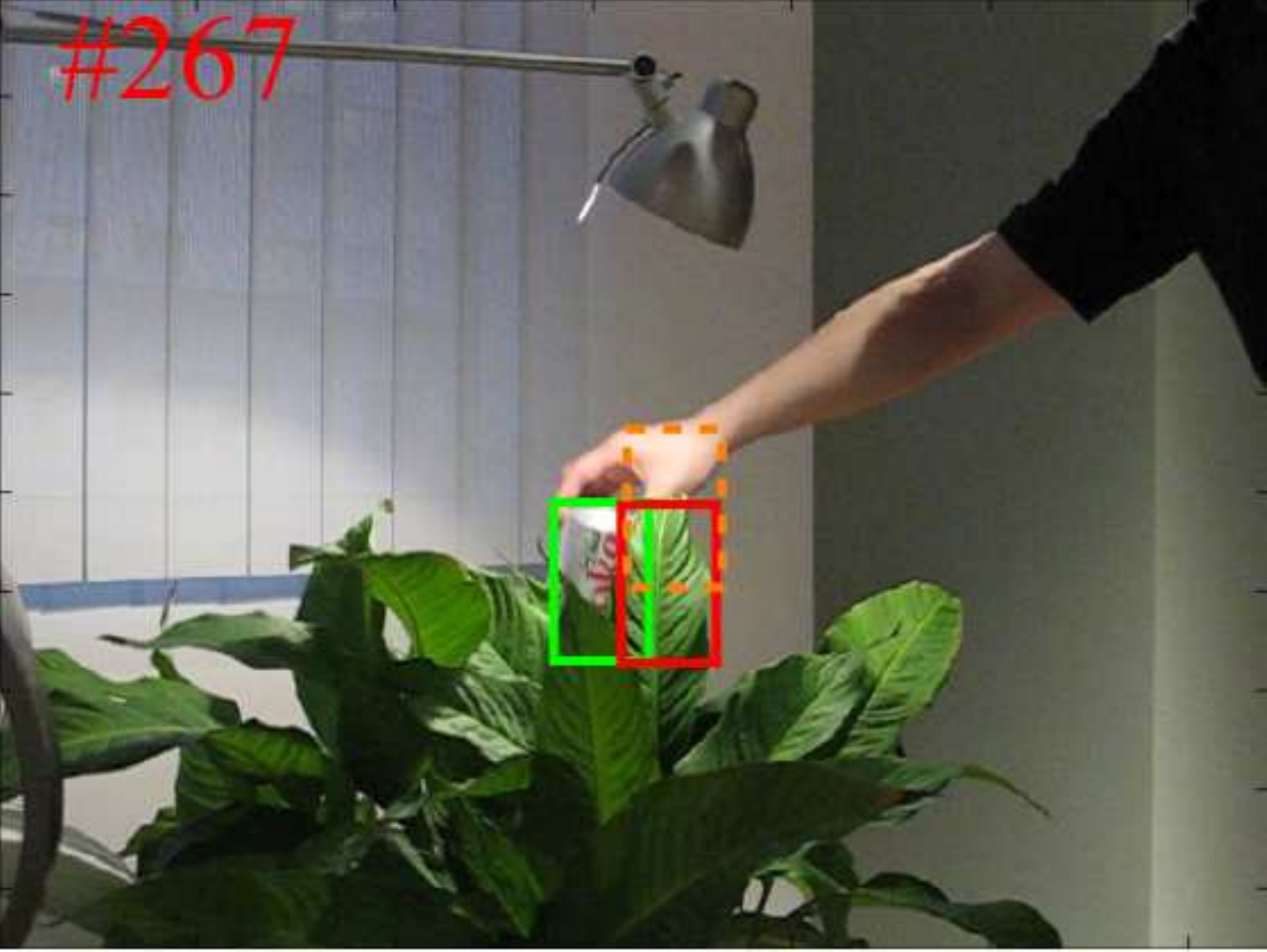}}
  \subfigure[]{\includegraphics[width=0.8in]{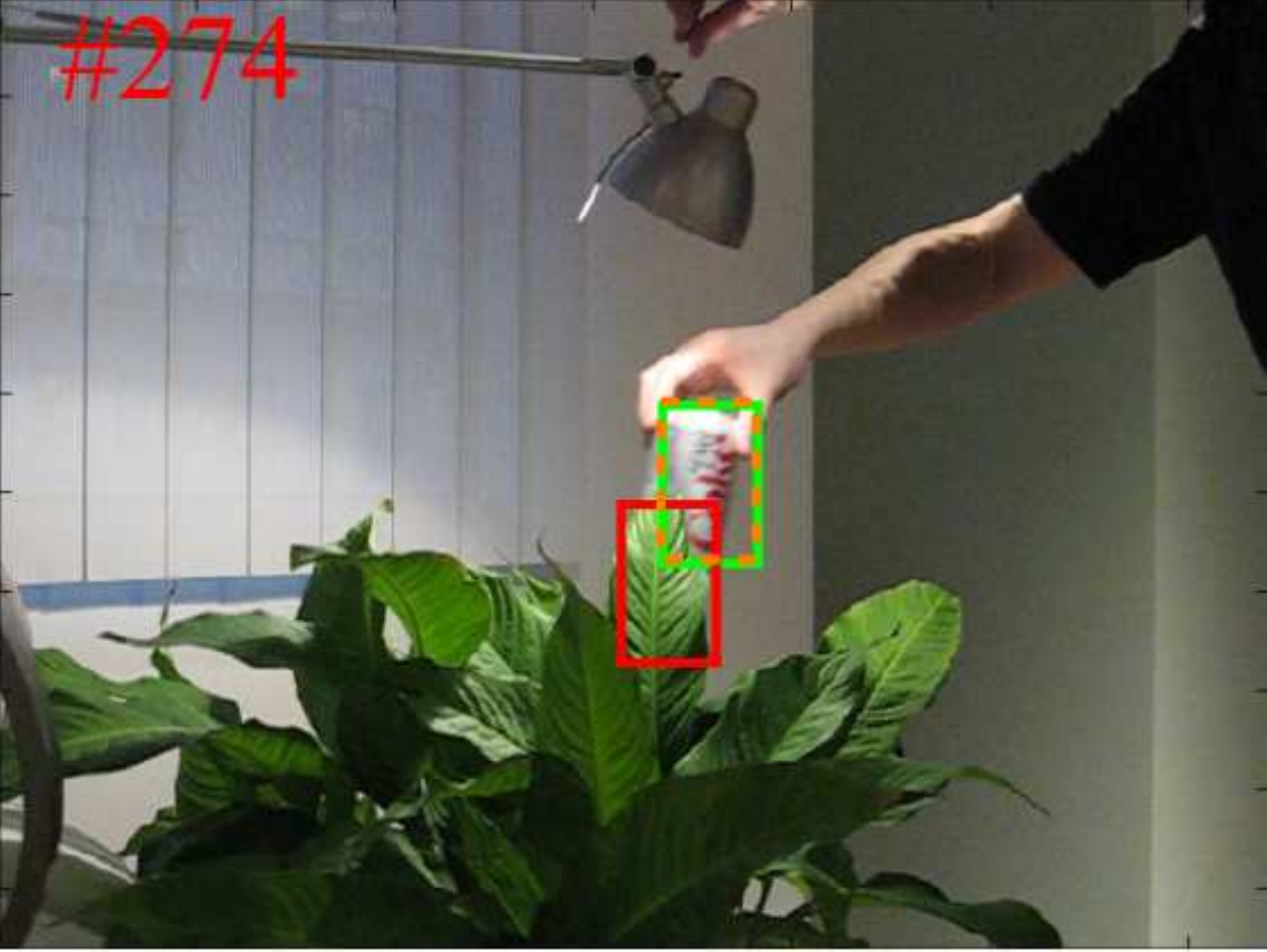}}\\
  \subfigure[]{\includegraphics[width=0.8in]{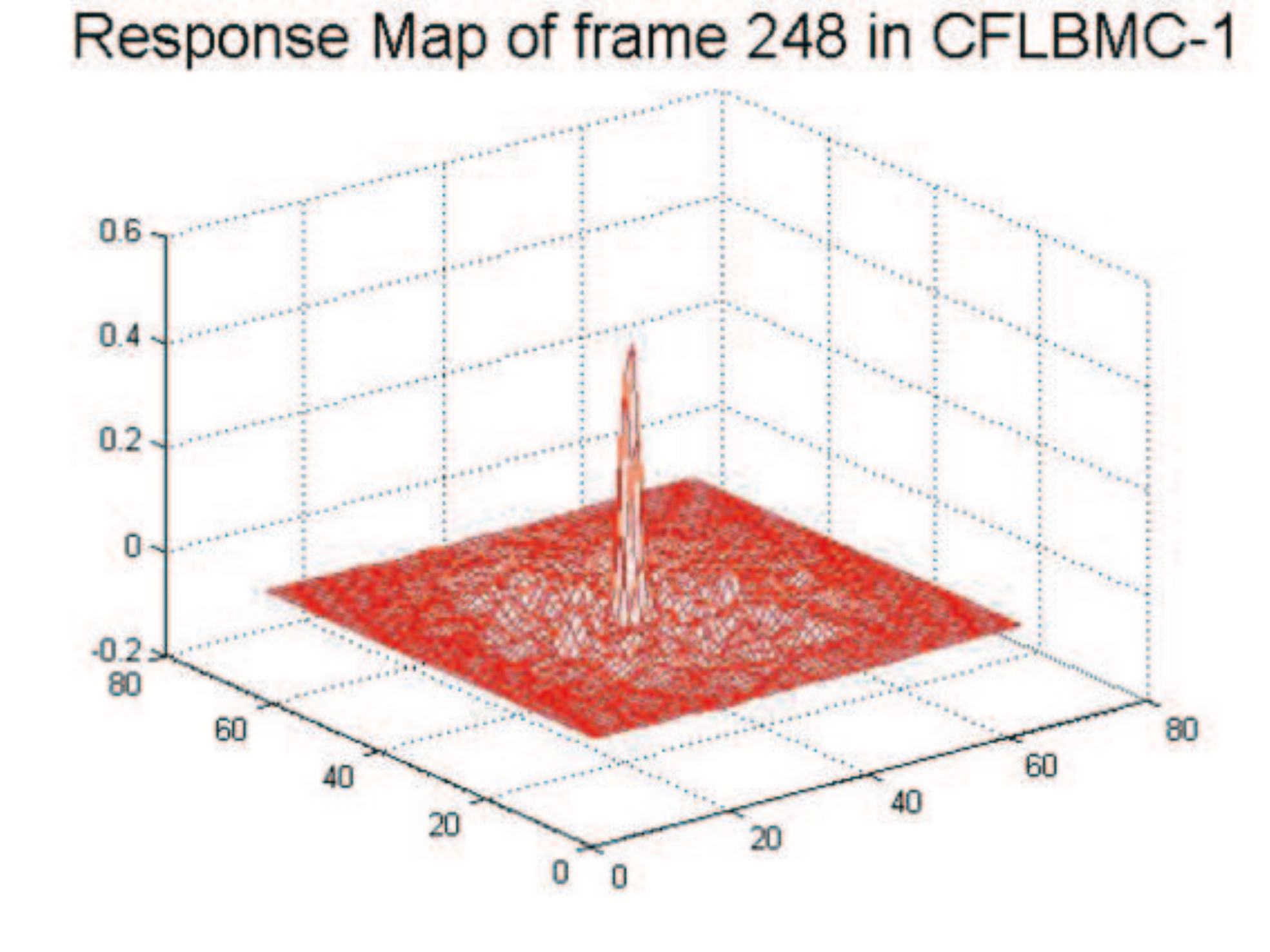}}
  \subfigure[]{\includegraphics[width=0.8in]{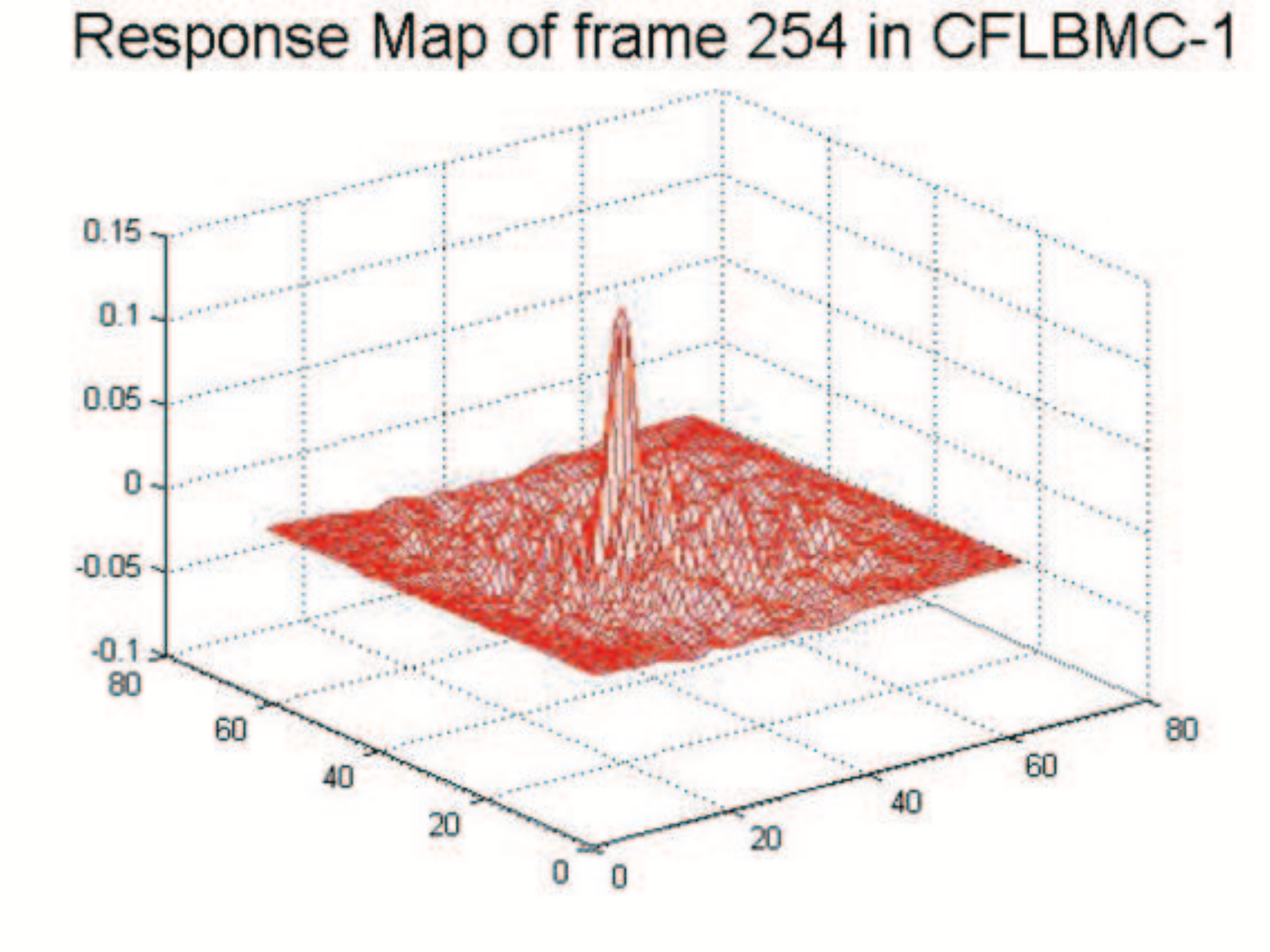}}
  \subfigure[]{\includegraphics[width=0.8in]{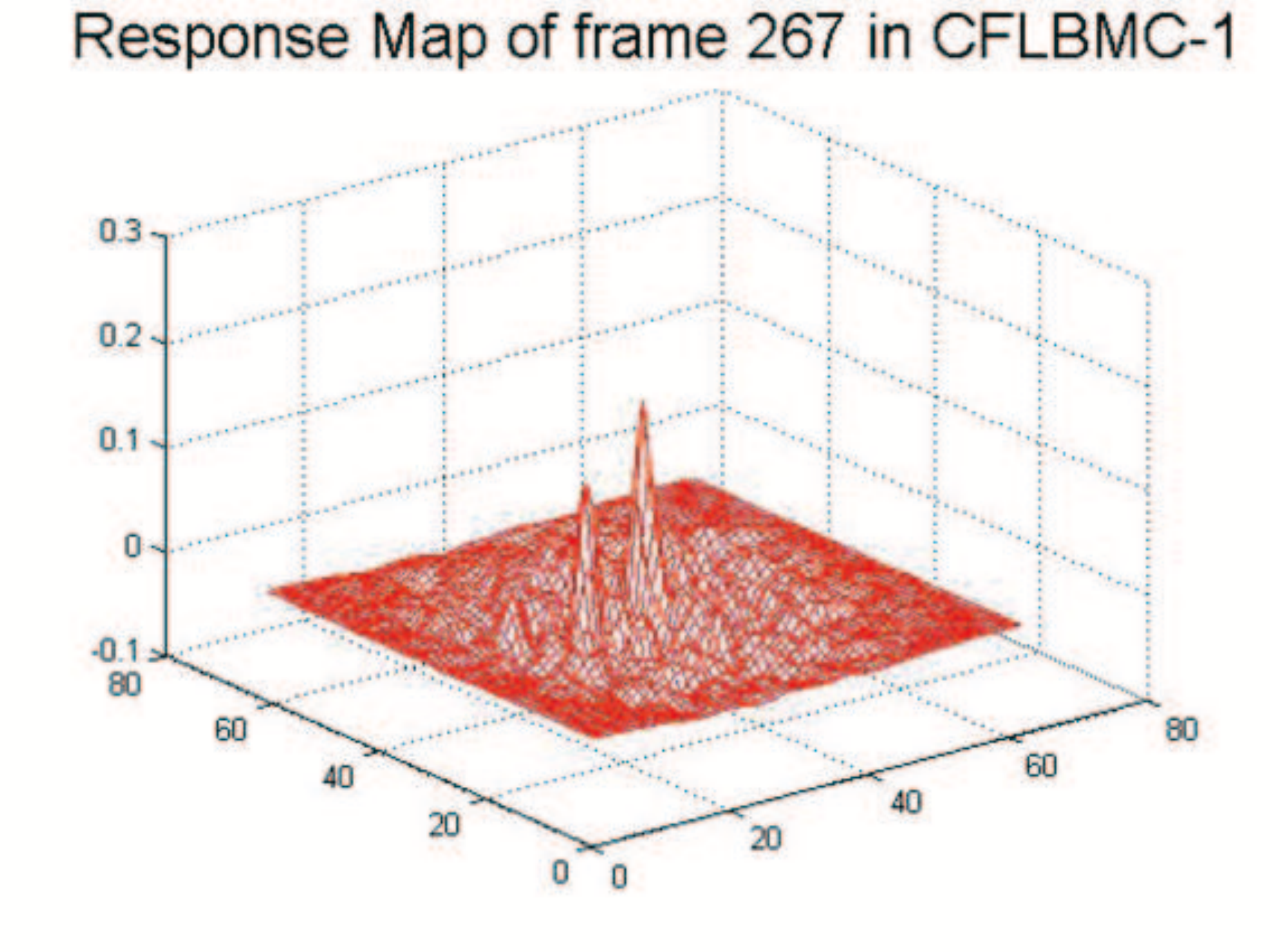}}
  \subfigure[]{\includegraphics[width=0.8in]{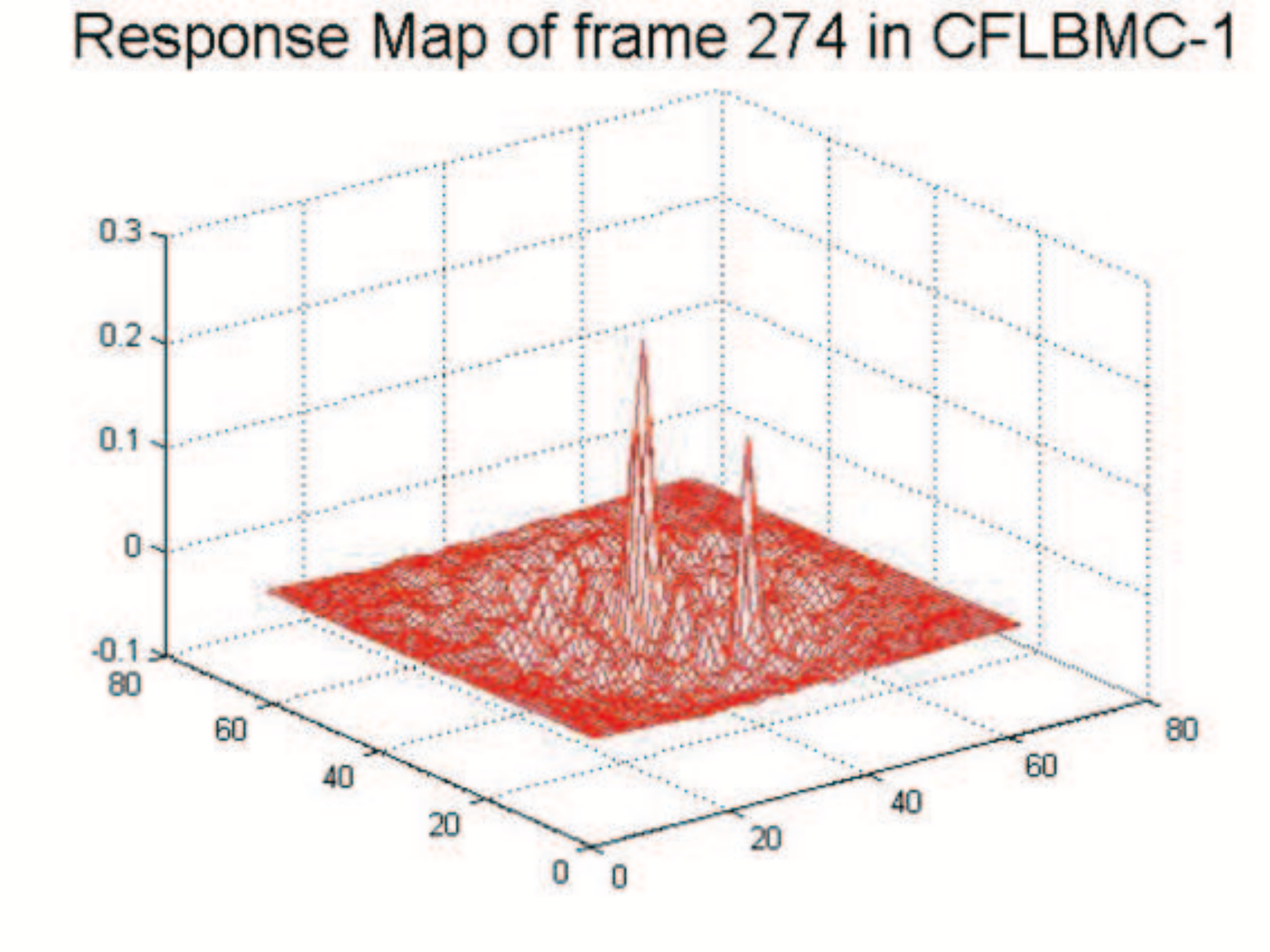}}\\
  \subfigure[]{\includegraphics[width=0.8in]{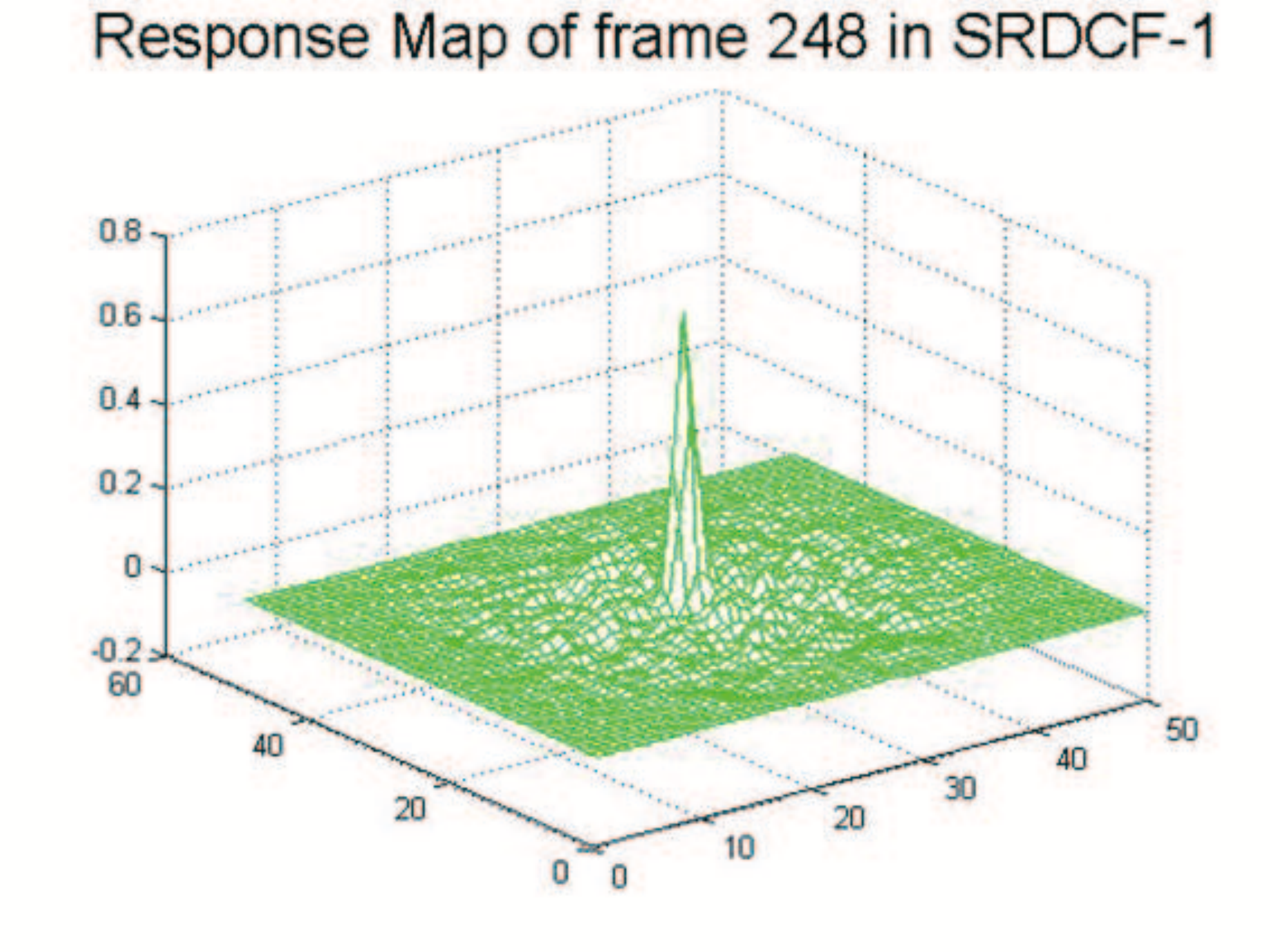}}
  \subfigure[]{\includegraphics[width=0.8in]{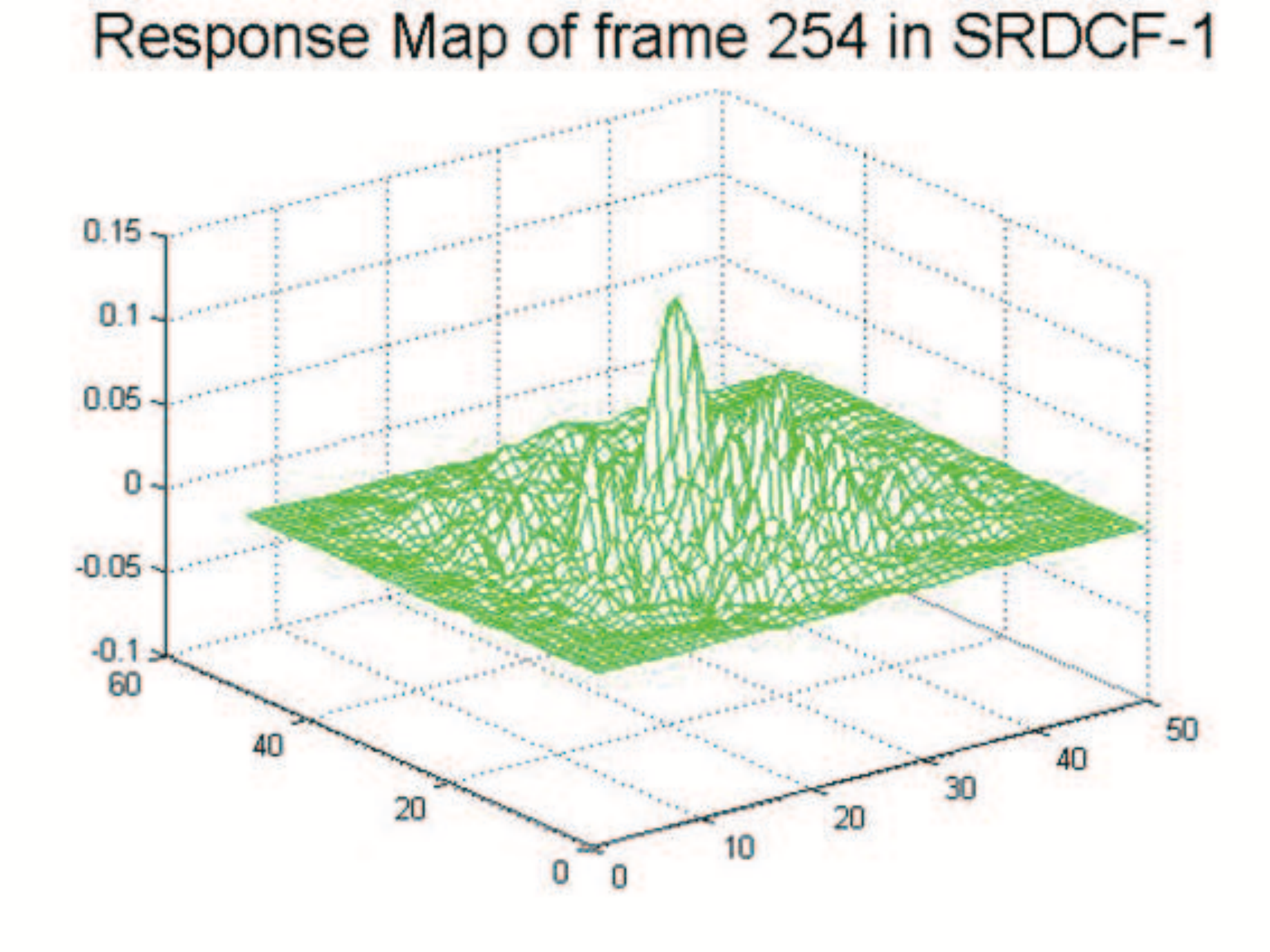}}
  \subfigure[]{\includegraphics[width=0.8in]{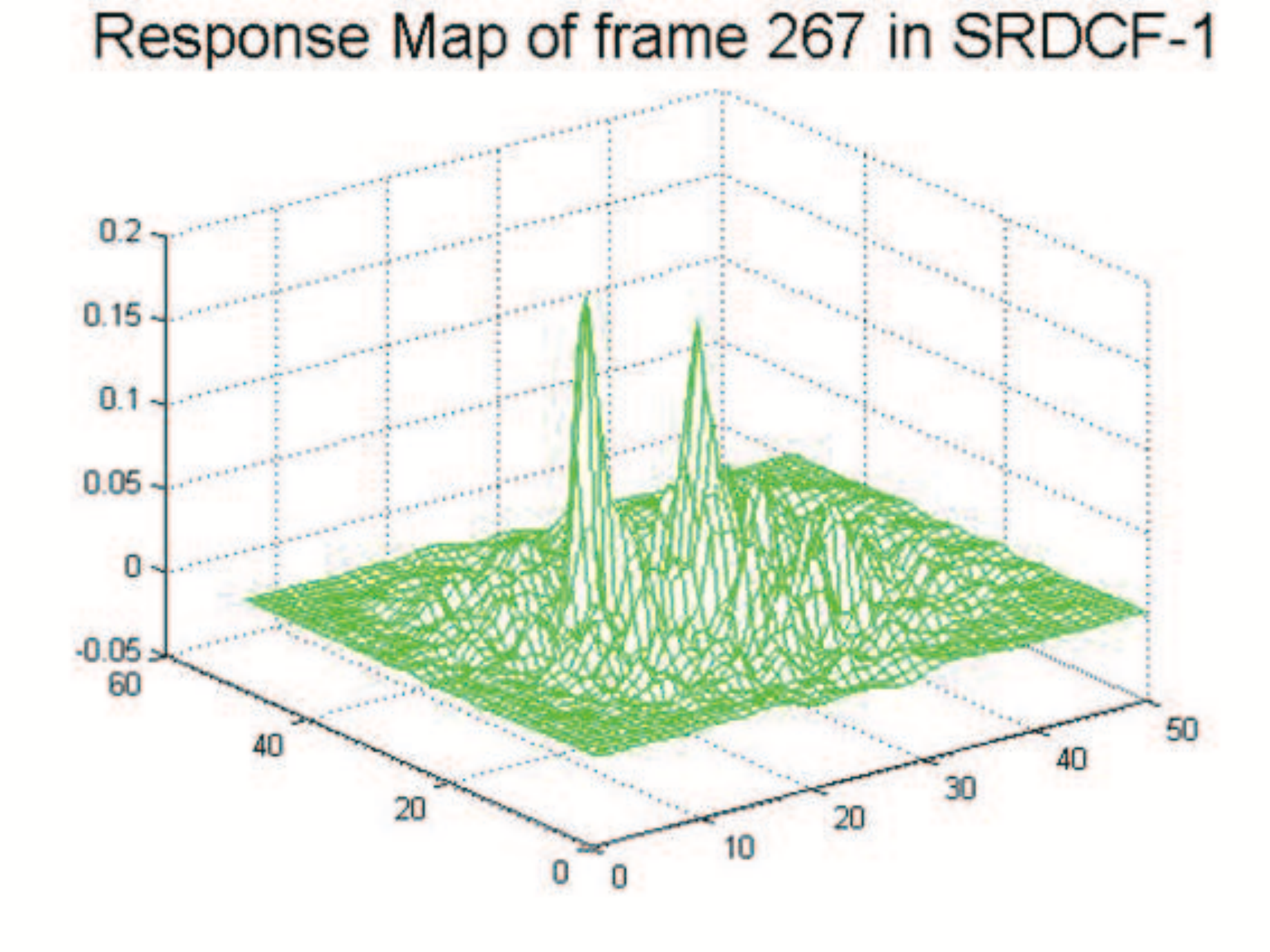}}
  \subfigure[]{\includegraphics[width=0.8in]{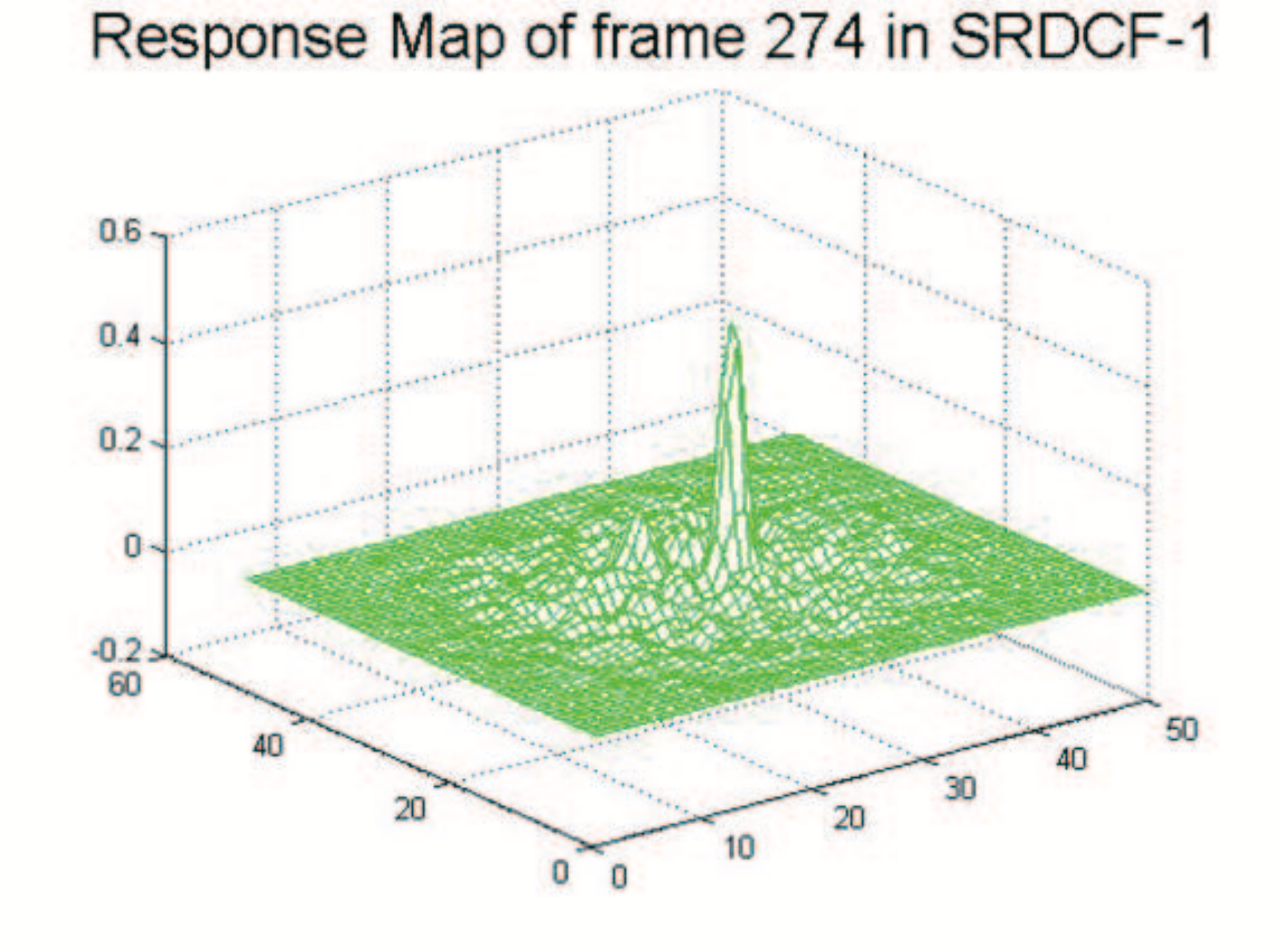}}\\
  \includegraphics[width=3.4in]{line.pdf}
  \caption{The coefficient effects on SRDCF-1 and CFLBMC-1 generated by both spatial regularization and Hann windows. See text for details.}
  \label{fig:Coke}
\end{figure}

The coefficient effects on SRDCF-1 and CFLBMC-1 generated by both spatial regularization and Hann windows are illustrated in Fig.~\ref{fig:Coke}. By comparing Figs.~\ref{fig:Coke} (e) and (f), and (i) and (j), it is found that the long term occlusion generates a dominant peak on the background leaf, therefore weakens the model discriminativities of both SRDCF-1 and CFLBMC-1, even though the local background is explored by SRDCF-1. And it is clear that the HOG of search region multiplied by Hann window will cripple the features near the search region boundaries and may also confuses the descriptions of the target object and its background. Consequently, the response maps of CFLBMC-1 contain two dominant peaks, and the highest one always corresponds to the leaf, as shown in Figs.~\ref{fig:Coke} (g) and (h). Whereas, the response maps of SRDCF-1 always contain the highest peaks which consistently correspond to the target coke with the help of local background, as shown in Figs.~\ref{fig:Coke} (k) and (l). Because Struck-L-1 always keeps reliable support vectors during tracking without Hann window multiplication, it is able to grasp the object robustly.

\begin{figure}[t]
  \centering
  \includegraphics[width=3.4in]{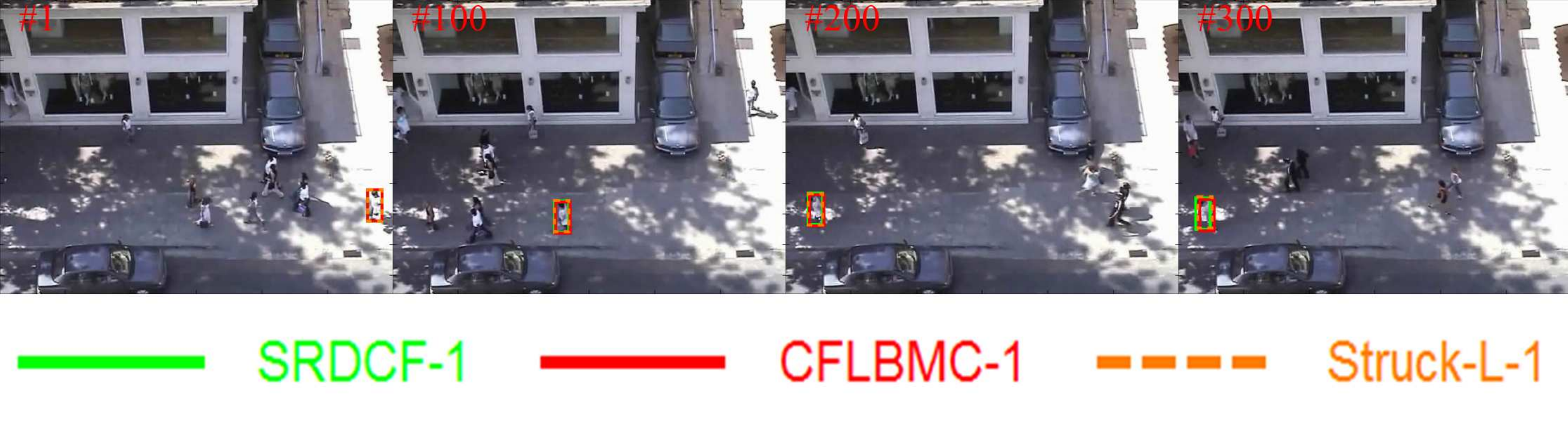}
  \caption{In Crowds sequence, the differences of SRDCF-1, CFLBMC-1, and Struck-L-1 are caused by the evaluation criterion, overlap ratio. See text for details.}
  \label{fig:Crowds}
\end{figure}

\begin{figure}[t]
  \centering
  \includegraphics[width=1.7in]{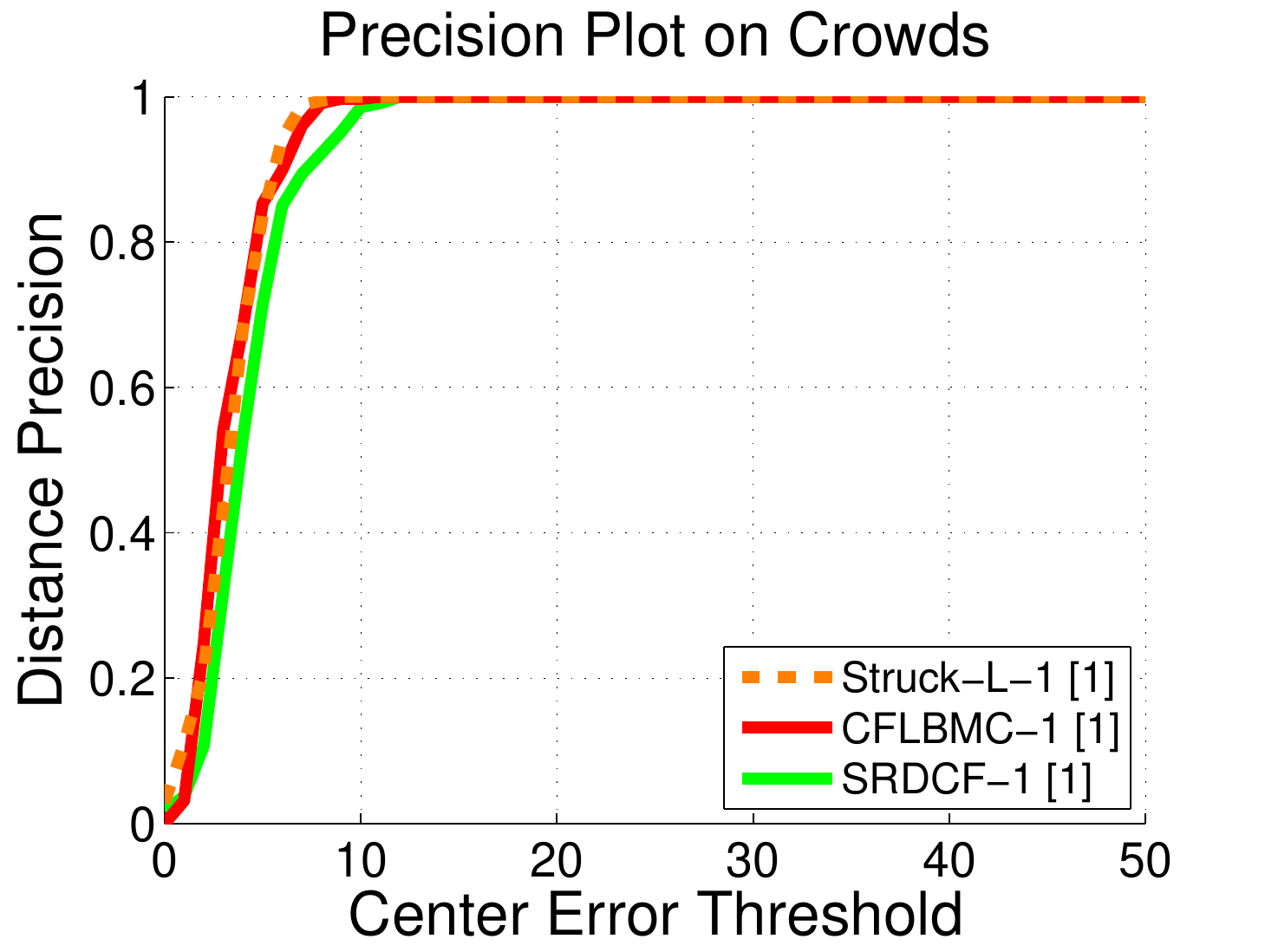}
  \includegraphics[width=1.7in]{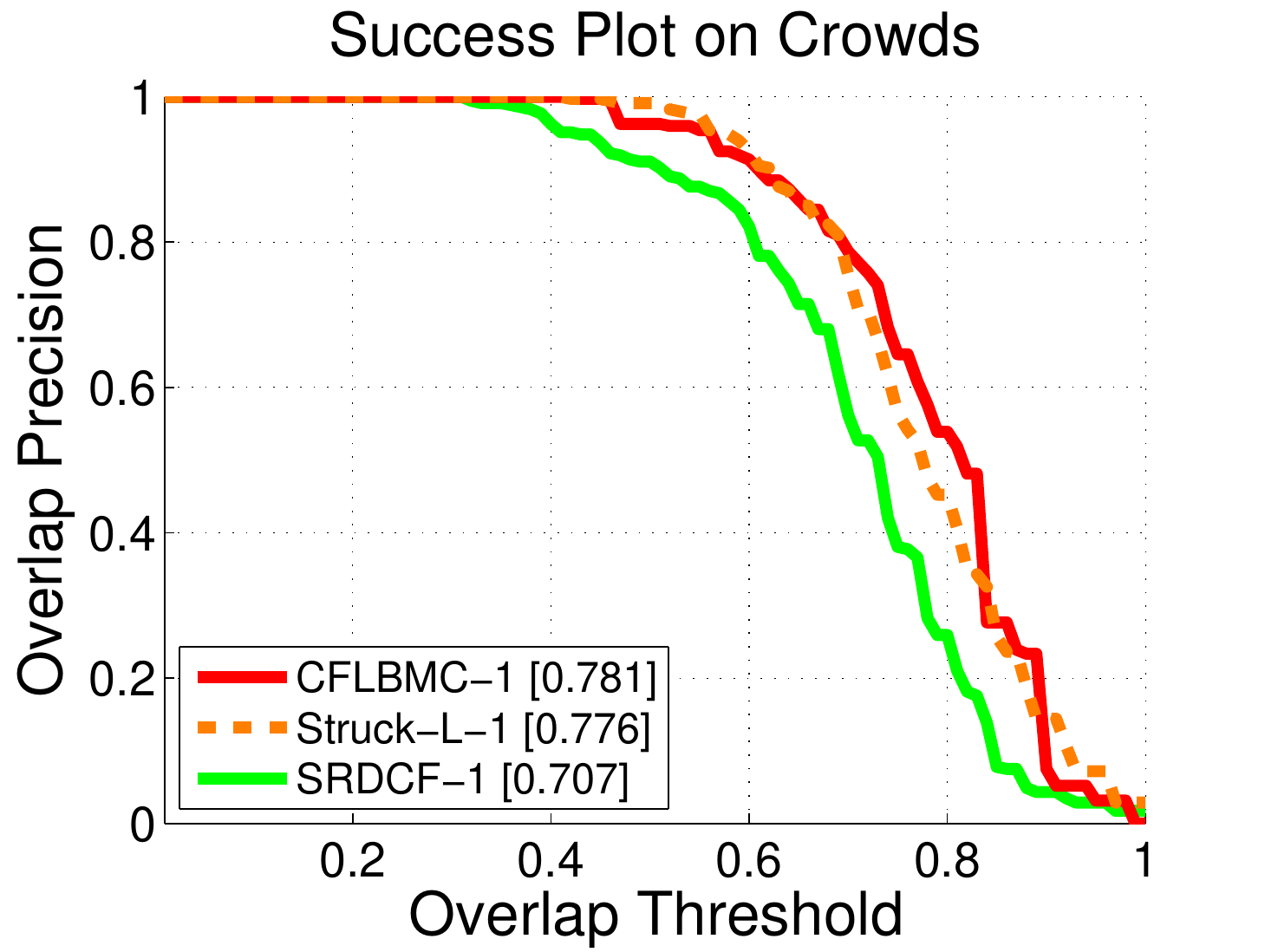}
  \caption{The precision and success plots of SRDCF-1, CFLBMC-1, and Struck-L-1 on Crowds sequence. The mean distance precision scores and AUCs of the trackers are reported in the legend. The differences of the three trackers are caused by the mathematical characteristic of overlap ratio. See text for details.}
  \label{fig:Crowds-plots}
\end{figure}

At last, it should be pointed out that the evaluation criterion, overlap ratio, will augment the fine distinctions of trackers when they all achieve high performances and the target object is small enough. It can be seen in Fig.~\ref{fig:Crowds} that SRDCF-1, CFLBMC-1, and Struck-L-1 achieve really close accuracies. Intuitively, there should not be obvious differences between their success plots. Nevertheless, large distinctions occur in their success plots, although their precision plots only exhibit small differences, as shown in Fig.~\ref{fig:Crowds-plots}. Note that the target object almost does not undergo scale changes in this sequence.

\subsubsection{Ingredients Influencing Performance}
\label{sec:improveperformance}
According to the above observations and analysis on sequences, we summarize the ingredients which will influence the performance as follows.
\begin{itemize}
  \item[1.] Exploring the local background of target object sometimes benefits and sometimes degrades the location accuracy. In the cases where the target appearance varies drastically or is occluded with little change of background, SRDCF-1 can locate the target more accurately than the other two trackers by means of learned local background. In other cases, involving local background in training may be harmful for the location accuracy. Therefore, the average difference between the location accuracies of SRDCF and CFLBMC on a benchmark is decided by the percentage of the sequences in which exploring local background will benefit the location accuracy.
  \item[2.] The large search region always benefits the location accuracy. If the target is occluded and then re-appears or moves a large distance in adjacent frames, the large search region may help the tracker catch the target correctly.
  \item[3.] The update method adopted by Struck benefits the location accuracy more than that adopted by SRDCF and CFLBMC, because the historical samples possess identical weight and the effect of $N$ similar samples on the appearance model approximately equals that of a single sample in Struck. This strategy may help Struck catch the target successfully after the target is missed for a long enough period.
  \item[4.] The location accuracy up to one pixel always benefits the location performance.
\end{itemize}


According to the above summarization, it can be concluded that, as far as the location performance is concerned, Struck is the most flexible and powerful because it is able to involve the local background in its samples, and there is no obstacle in enlarging its search region. In addition, Struck is able to employ non-linear kernels to improve its performance further, whereas SRDCF and CFLBMC are not. Nevertheless, the unacceptable computational cost prevents Struck from exploring effective ingredients to improve itself, because Struck can not employ FFT to accelerate its training and locating greatly.

Although a great many of fine differences of SRDCF, CFLBMC, and Struck may lead to small inconsistence in performance, that the effects of fine differences on performance are positive or negative is dependent on the characteristic of the specific sequence. Statistically, such inconsistence may be expected to vanish if the benchmark is large enough and diverse enough. Or, the consistence of SRDCF, CFLBMC, and Struck in experimental performance on a benchmark may be a measure of the scale and diversity of the benchmark.

It should be pointed out that the optimization objectives of SRDCF and CFLBMC are similar, whereas their optimization procedures are quite different, resulting in the efficiency of SRDCF is over ten times higher than that of CFLBMC. Note that Galoogani~\etal~\cite{galoo15} reported that the fps of CFLB is about 100. This is because CFLB only employs gray as its feature. Due to the multi-channel characteristic of CFLBMC, the same optimization algorithm utilized by CFLBMC and CFLB makes the former much slower than the latter.

\subsection{Struck with Gaussian Kernel}
\label{sec:struckwithgaussian}

\begin{figure}[t]
  \centering
  \includegraphics[width=1.7in]{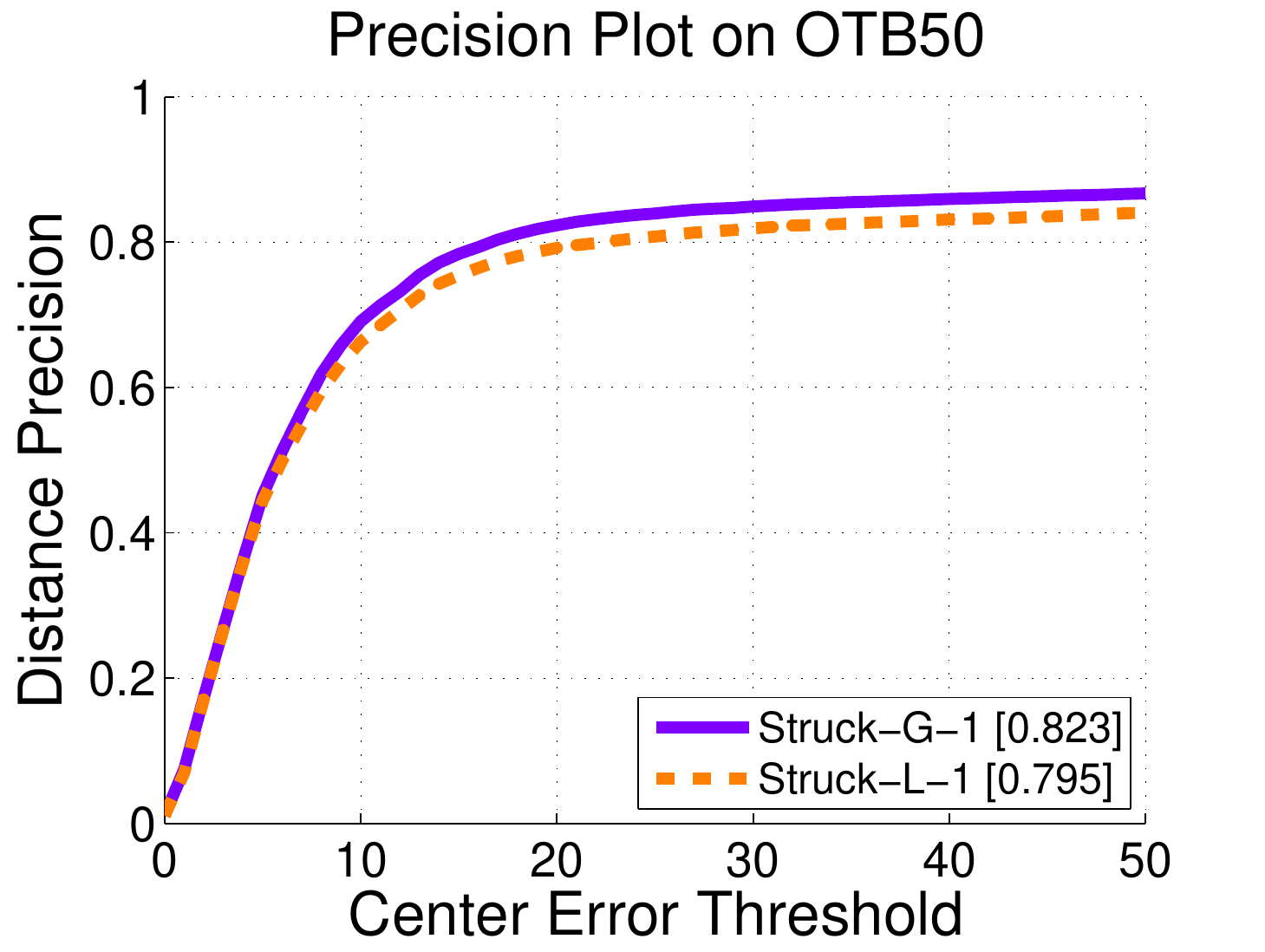}
  \includegraphics[width=1.7in]{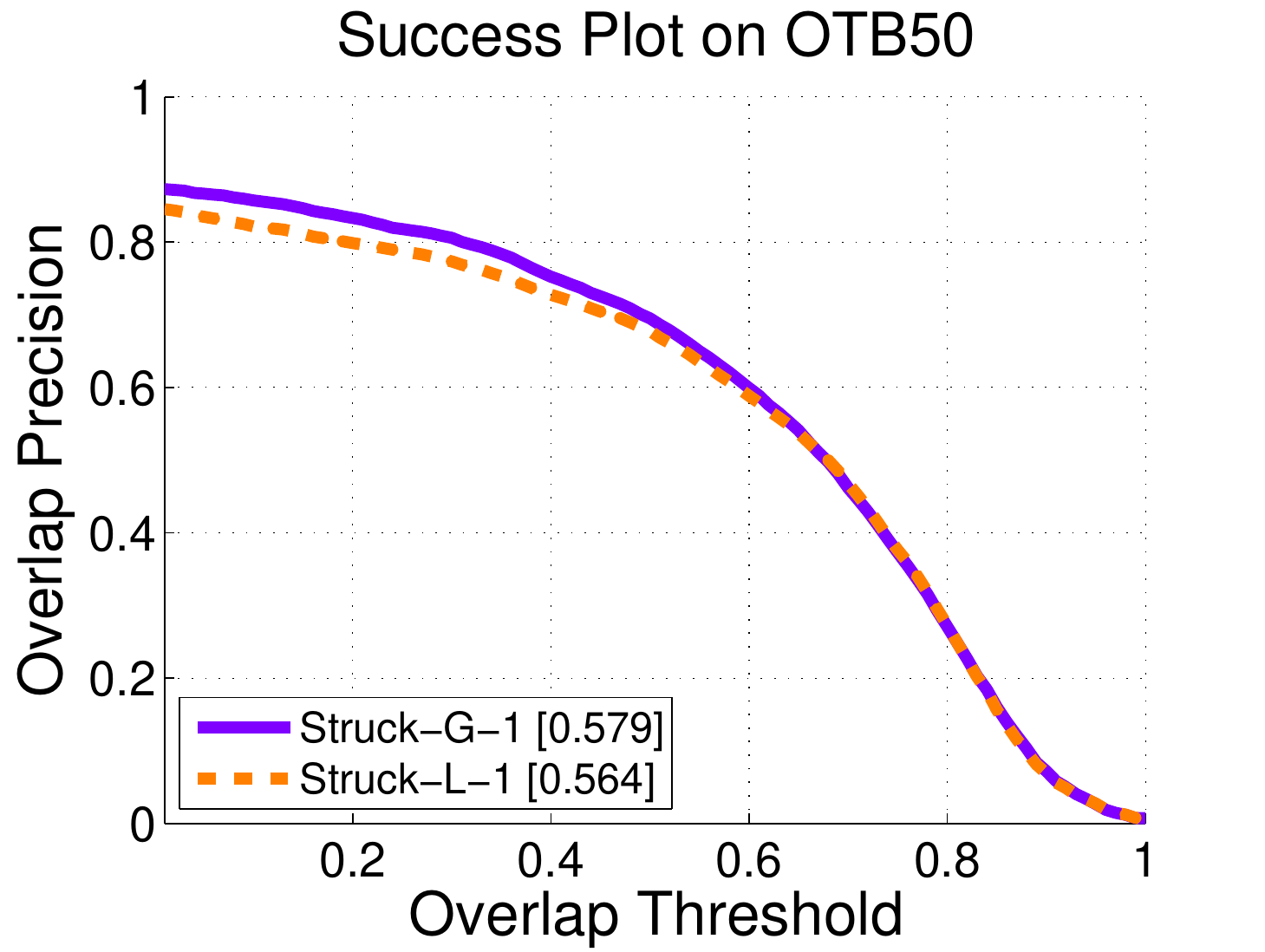}
  \caption{The average precision plots of Struck-L-1 and Sturck-G-1 on OTB50. The mean distance precision scores and AUCs of the trackers are reported in the legend. See text for details.}
  \label{fig:Struck_Gauss}
\end{figure}

To verify the positive effect of the non-linear kernel on performance, the linear kernel is replaced by a Gaussian one in the modified Struck in this section. The Struck with the Gaussian kernel and without scale estimation, denoted as Struck-G-1, is compared with Struck-L-1. It is seen in Fig.~\ref{fig:Struck_Gauss} that Struck-G-1 outperforms Struck-L-1 consistently on OTB50. Of course, the total processing time of Struck-G-1 is about $40\%$ more than that of Struck-L-1 on the benchmark.

\subsection{Struck and CFLBMC with Scale Estimation}
\label{sec:struckvscflbmcwithscale}

\begin{figure*}[t]
  \centering
  \includegraphics[width=1.7in]{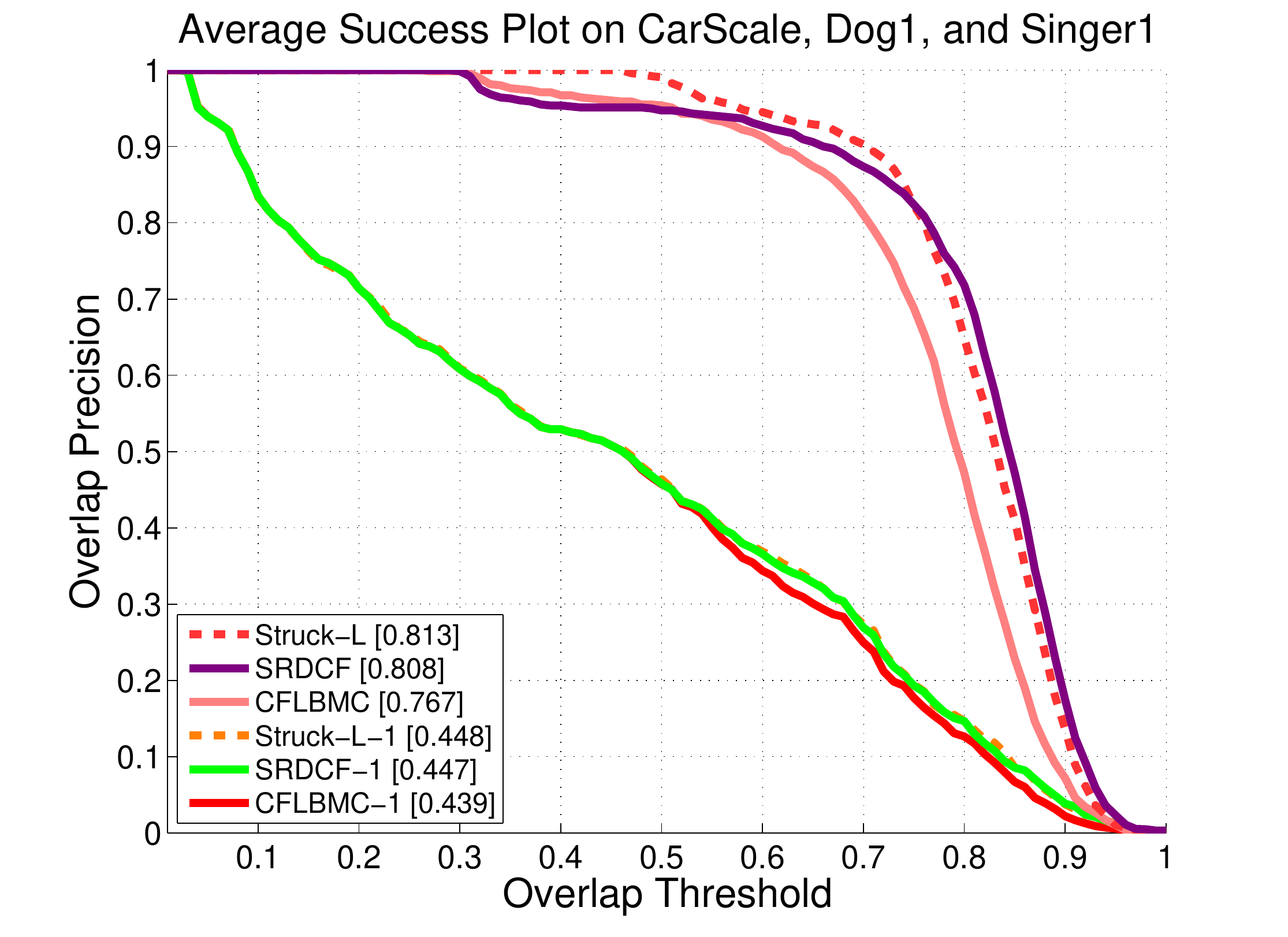}
  \includegraphics[width=1.7in]{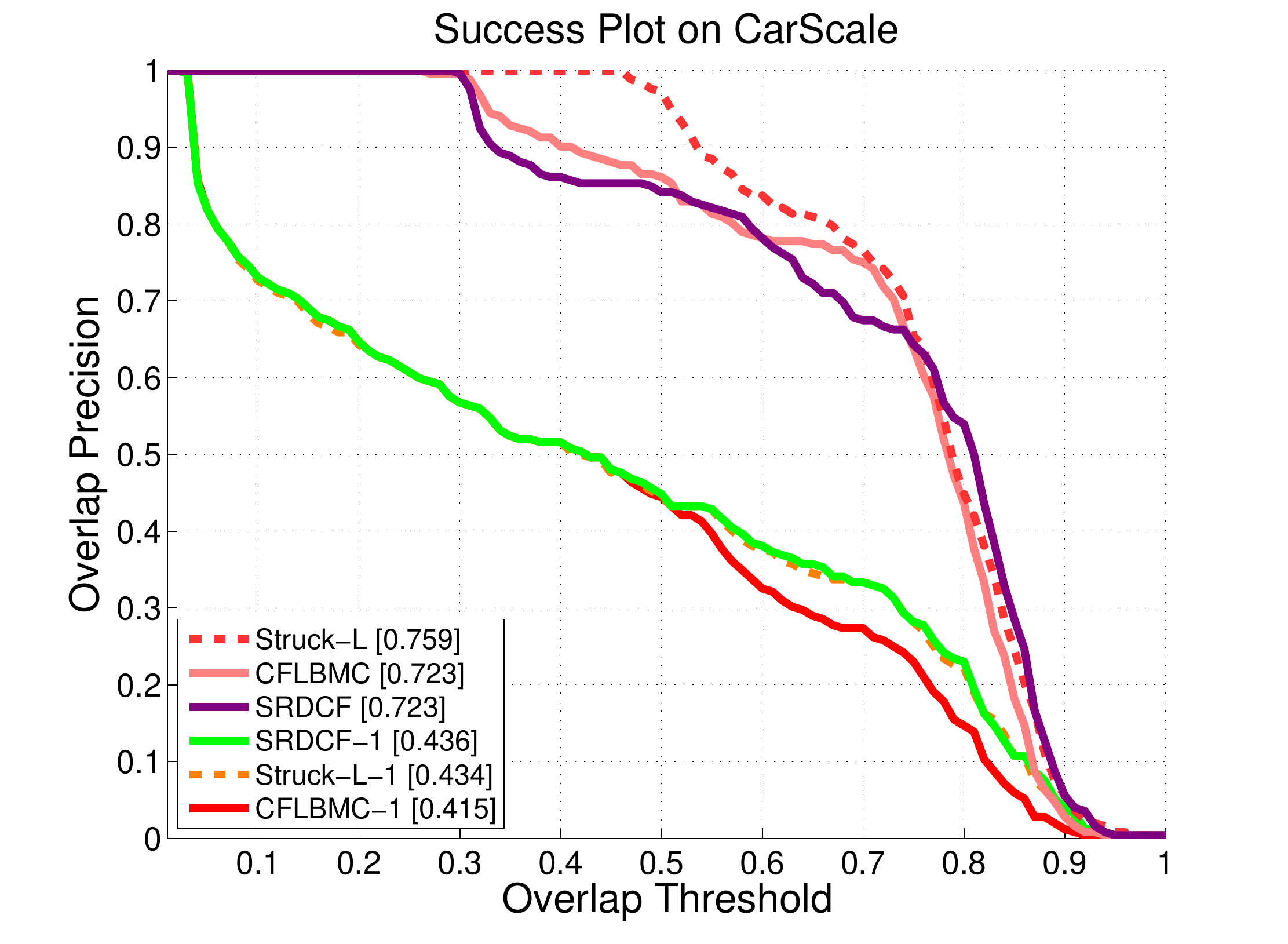}
  \includegraphics[width=1.7in]{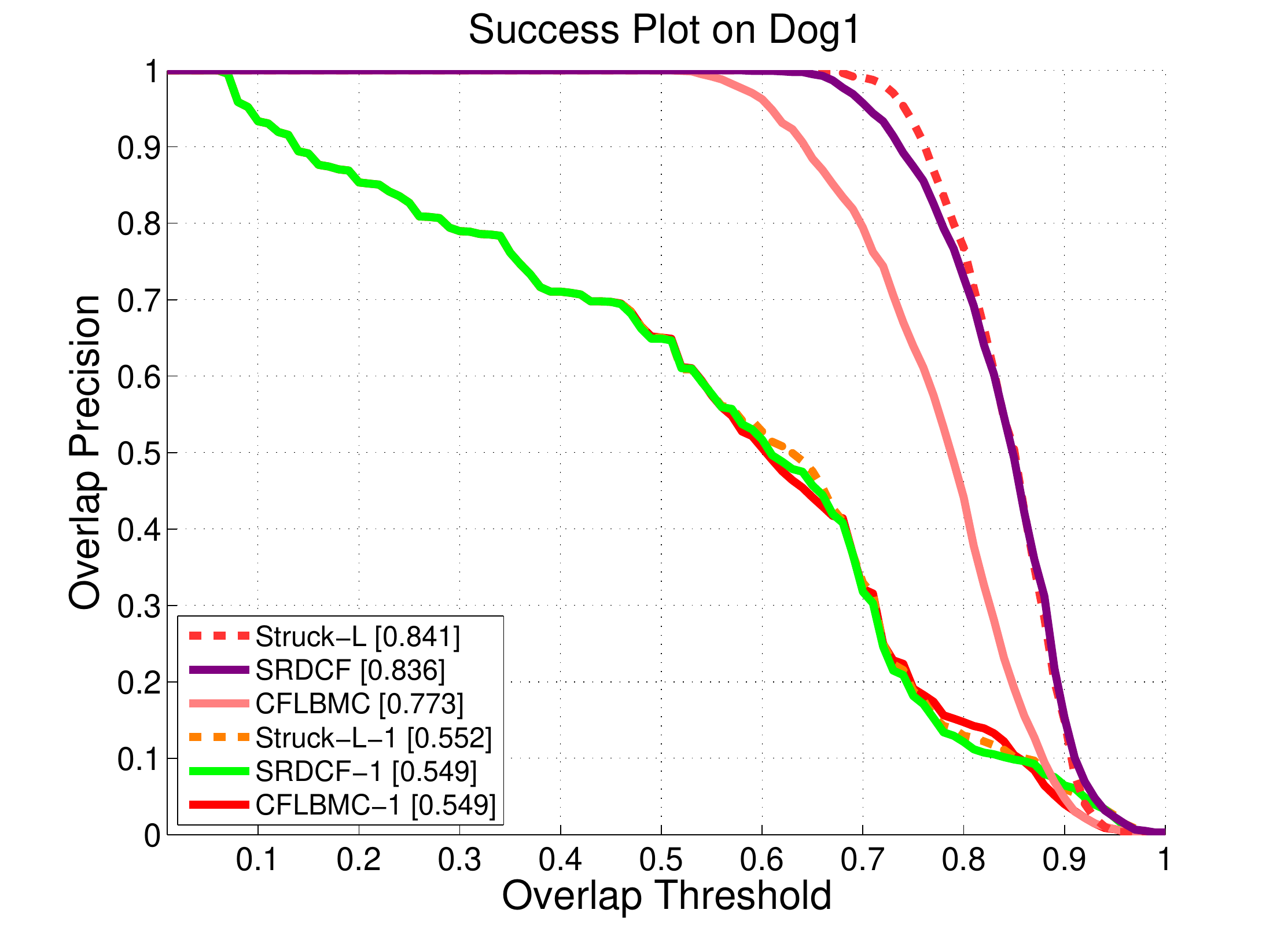}
  \includegraphics[width=1.7in]{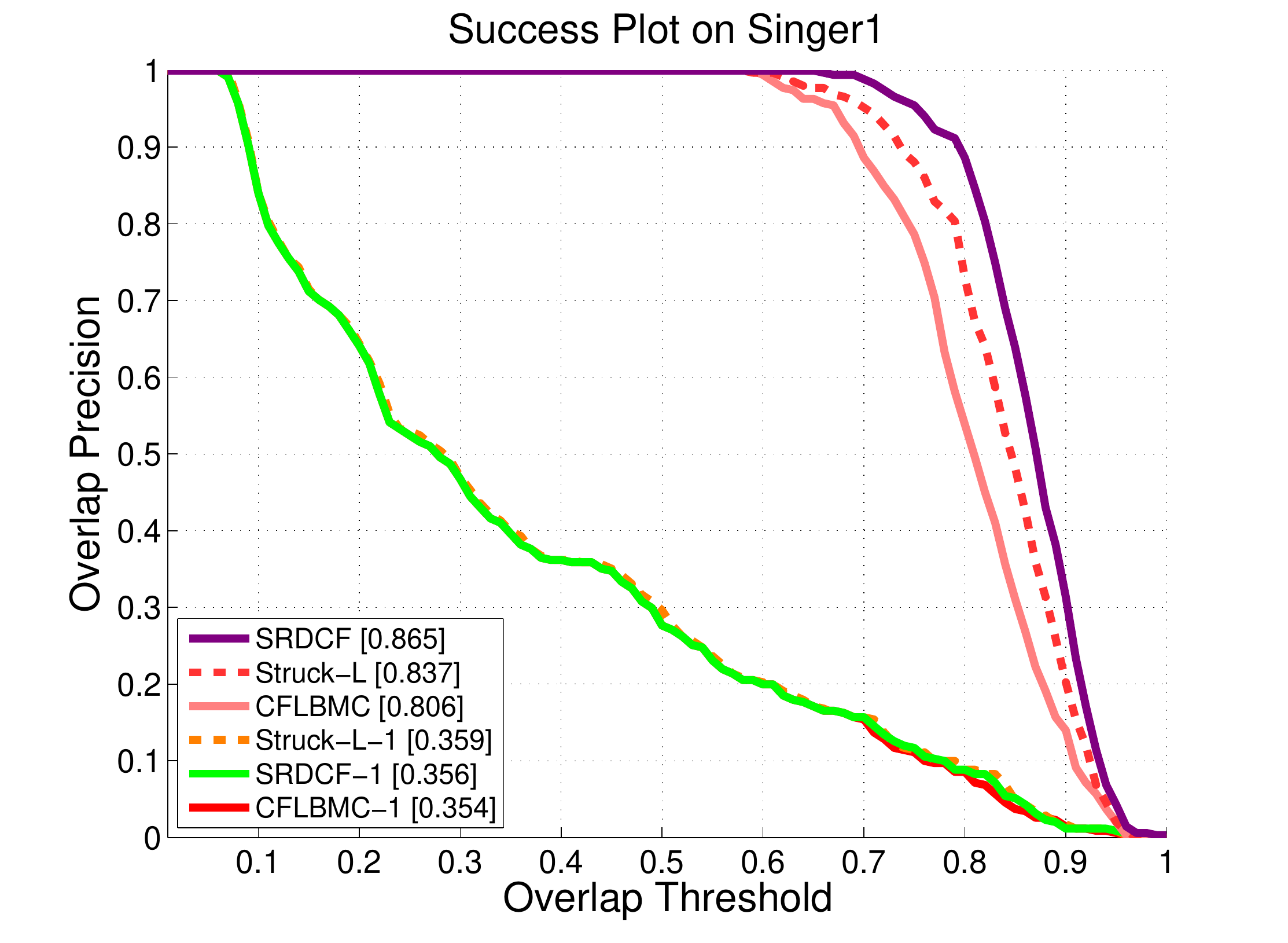}
  \caption{The success plots of six trackers in the three sequences CarScale, Dog1, and Singer1. SRDCF, CFLBMC, and STRUCK-L denote the original SRDCF, CFLBMC with scale estimation, and the modified Struck with line kernel and scale estimation. The AUCs of the trackers are reported in the legends. See text for details.}
  \label{fig:scale}
\end{figure*}

In order to investigate the effect of scale estimation on their relation, we introduce the scale estimation into CFLBMC-1 and Struck-L-1, and denote their new versions as CFLBMC and Struck-L, respectively. To identify the scale levels adopted in SRDCF, seven candidate scales are tested in CFLBMC and Struck-L after they locate the object center. And the optimal scale is the one under which the object has the highest response among all the scale candidates. We compare these trackers in CarScale, Dog1 and Singer1 in which the scale estimation would be the only challenge. Fig.~\ref{fig:scale} shows the results. It is seen in the figure that the performances of SRDCF, CFLBMC and Struck-L are significantly improved in comparison to their non-scale-estimation versions, SRDCF-1, CFLBMC-1, and Struck-L-1. It is noticed that the slight differences between SRDCF-1, CFLBMC-1, and Struck-L-1 are magnified by the scale estimation in their counterparts, SRDCF, CFLBMC and Struck-L. Note that the success plots of CFLBMC are even remarkably inferior to those of the other two trackers. This is because it only employs the strategy of location accuracy up to 4 pixels, while SRDCF and Struck-L employ the strategy up to 1 pixel, and the mathematical characteristic of the overlap ratio curve in small objects and center errors further amplifies their performance differences.

\section{Conclusion}
\label{sec:conclusion}
In this paper, we have mainly investigated two types of visual tracking algorithms, CF trackers (specifically, SRDCF and CFLBMC) and Struck, theoretically proved three relations between them, and experimentally verified these relations. According to the experimental results, we summarized how slight differences of these three trackers qualitatively affect their performances. We have also analyzed the relations of CF trackers, Struck, and the ranking SVM based tracker.

We wish our work would encourage the visual tracking community not only to design new algorithms and compare a variety of trackers experimentally, but also to investigate their relationship theoretically, so as to develop essentially stronger algorithms in both location performance and efficiency.

\bibliographystyle{ieee}
\bibliography{qwangbib}

%

\end{document}